\documentclass[twoside]{article}

\usepackage[linewidth=1pt]{mdframed}

\usepackage{amsthm}
\usepackage{aistats2020}
\usepackage[round]{natbib}

\usepackage{booktabs}
\usepackage{multirow}
\usepackage{graphicx}
\usepackage{caption}
\usepackage{subcaption}
\usepackage{amssymb}
\usepackage[ruled]{algorithm2e}
\usepackage{color}
\usepackage{url}
\usepackage{tikz}
\usepackage{xspace}

\theoremstyle{definition}
\newtheorem{example}{Example}[section]
\newtheorem{definition}{Definition}[section]

\newcommand{\name}[1]{{\textsc{#1}}\xspace}

\newcommand{\AND}{\name{and}}
\newcommand{\OR}{\name{or}}
\newcommand{\NOT}{\name{not}}

\newcommand{\CBA}{\name{cba}}
\newcommand{\CPAR}{\name{cpar}}
\newcommand{\CMAR}{\name{cmar}}
\newcommand{\CN}{\name{cn2}}
\newcommand{\FOIL}{\name{foil}}
\newcommand{\BRS}{\name{brs}}

\newcommand{\LIBRE}{\name{libre}}
\newcommand{\RBFSVM}{\name{rbf-svm}}

\newcommand{\DT}{\name{dt}}
\newcommand{\RIPPERK}{\name{ripper-k}}
\newcommand{\SBRL}{\name{s-brl}}

\newcommand{\RF}{\name{rf}}
\newcommand{\IDS}{\name{ids}}

\newcommand{\MODLEM}{\name{modlem}}

\newcommand{\BRACID}{\name{bracid}}
\newcommand{\CG}{\name{cg}}

\newcommand{\Adult}{\name{Adult}}
\newcommand{\Australian}{\name{Australian}}
\newcommand{\Balance}{\name{Balance}}
\newcommand{\Bank}{\name{Bank}}

\newcommand{\Haberman}{\name{Haberman}}
\newcommand{\Heart}{\name{Heart}}
\newcommand{\Ilpd}{\name{Ilpd}}

\newcommand{\Liver}{\name{Liver}}
\newcommand{\Pima}{\name{Pima}}
\newcommand{\Sap}{\name{Sap}}
\newcommand{\Sapclean}{\name{Sap-Clean}}
\newcommand{\Sapfull}{\name{Sap-Full}}
\newcommand{\Sonar}{\name{Sonar}}

\newcommand{\Tictactoe}{\name{Tictactoe}}
\newcommand{\Transfusion}{\name{Transfusion}}
\newcommand{\Wisconsin}{\name{Wisconsin}}

\usepackage{hyperref}
\usepackage{cleveref}

\begin{document}

%

%

\twocolumn[
	\aistatstitle{
		LIBRE: Learning Interpretable Boolean Rule Ensembles
	}
	\aistatsauthor{
		Graziano Mita \And 
		Paolo Papotti \And 
		Maurizio Filippone \And 
		Pietro Michiardi
	}
	\aistatsaddress{
		EURECOM, 06410 Biot (France)\\
		\{mita, papotti, filippone, michiardi\}@eurecom.fr
	}
]

\begin{abstract}
\label{sec:abstract}
We present a novel method -- \LIBRE -- to learn an interpretable classifier, which materializes as 
a set of Boolean rules. \LIBRE uses an ensemble of bottom-up, weak learners operating on a random subset
of features, which allows for the learning of rules that generalize well on unseen data even in imbalanced
settings. Weak learners are combined with a simple union so that the final ensemble is also
interpretable. Experimental results indicate that \LIBRE efficiently strikes the right balance between prediction 
accuracy, which is competitive with black box methods, and interpretability, which is often superior to
alternative methods from the literature.
\end{abstract}

\section{INTRODUCTION}
\label{sec:intro}
Model interpretability has become an important factor to consider when applying machine
learning in critical application domains. In medicine, law, and predictive maintenance,
to name a few, understanding the output of the model is at least as important as the
output itself. However, a large fraction of models currently in use (e.g. Deep Nets,
SVMs) favor predictive performance at the expenses of interpretability.

To deal with this problem, interpretable models have flourished in the machine
learning literature over the past years. Although defining interpretability
is difficult~\citep{DBLP:journals/corr/Miller17a,DBLP:journals/corr/Doshi-VelezK17}, the
common goal of such methods is to provide an explanation of their output. The form and
properties of the explanation are often application specific.

In this work, we focus on predictive rule learning for challenging applications
where data is unbalanced.
For rules, interpretability translates into simplicity, and it is measured as a function of the
number of rules and their size (average number of atoms): such proxies are easy to
compute, understandable, and allow comparing several rule-based models.
The goal is to learn a set of rules from the training set that (i) effectively
predict a given target, (ii) generalize to unseen data, (iii) and are interpretable, i.e.,
a small number of short rules (e.g., \cref{fig:liver_ruleset}). The first
objective is particularly difficult to meet in presence of imbalanced data.
In this case, most rule-based methods fail at characterizing the minority class. 
Additional data issues that hinder the application of rule-based methods~\citep{Weiss2004} are
data fragmentation (especially in case of \textit{small-disjuncts}~\citep{Holte1989}), overlaps between imbalanced
classes, and presence of rare examples.

\begin{figure}

\footnotesize
\begin{mdframed}
\textbf{IF}  mean\_corpuscular\_volume $\in$ [90, 96)\\
\textbf{OR}  gamma\_glutamyl\_transpeptidase $\in$ [20, max]\\
\textbf{THEN}  liver\_disorder = True\\
\textbf{ELSE}  liver\_disorder = False
\end{mdframed}

\caption{Example of rules learned by \LIBRE for \Liver.}
\label{fig:liver_ruleset}
\end{figure}


Many seminal 
rule learning methods come from
the data mining community: \CBA~\citep{Liu:1998:ICA:3000292.3000305}, \CPAR~\citep{Yin03},
and \CMAR~\citep{Li01}, for example, use mining to identify class association rules
and then choose a subset of them according to a ranking 
to implement the
classifier. In practice, however, these methods output 
a huge number of rules, which
negatively impacts interpretability.

Another family of approaches includes methods like \CN~\citep{Clark1989}, \FOIL~\citep{Quinlan93Foil},
and \RIPPERK~\citep{Cohen95}, whereby \textit{top-down} learners build rules by greedily adding
the condition that best explains the remaining data. 
\textit{Top-down} learners are well suited for noisy data and are known to find general
rules~\citep{Furnkranz:2014}. They work well for the so called \textit{large
disjuncts}, but have difficulties to identify \textit{small-disjuncts} and rare examples, which are
quite common in imbalanced settings. In contrast, \textit{bottom-up} learners like
\MODLEM~\citep{Modlem2001}, start directly from very specific rules (the examples themselves)
and generalize them until a given criteria is met. Such methods are susceptible to noise, and
tend to induce a very high number of specific rules, but are better suited for cases where
only few examples characterize the target class~\citep{Furnkranz:2014}.

Hybrid approaches such as \BRACID~\citep{Napierala2012Bracid} take the best from both worlds:
\textit{maximally-specific} (the examples themselves) and general rules are used
together in a hybrid classification strategy that combines rule learning and instance-based
learning. Thus, they achieve better generalization, also in imbalanced settings, but still
generate many rules, penalizing interpretability. Other approaches to
tackle data-related issues include heuristics to inflate the importance of rules for
minority classes~\citep{Grzymala2000, Nguyen2005, Blaszczynski2010}.

Recent work focus on marrying competitive predictive accuracy with {\em high
interpretability}. A popular approach is to use the output of an association rule discovery
algorithm (like \textsc{FP Growth}) and combine the discovered rules in
a \textit{small} and \textit{compact} subset with high predictive performance. The rule combination
process can be formalized either as an integer optimization problem or solved heuristically, explicitly
encoding interpretability needs in the optimization function.
Such approaches have been successfully applied to rule lists~\citep{Yang2017_sbrl,Chen2018_frl,Angelino2018}
and rule sets~\citep{Lakkaraju2016_ids,Wang2017_brs}. Alternatively, rules can be directly learned
from the data through an integer optimization framework~\citep{Hauser10,Chang2012AnIO,Malioutov2013,Goh:2014,Su16,Dash2018}.

Both rule-mining and integer-optimization based approaches underestimate the complexity and
importance of finding good candidate rules, and become expensive when the input dimensionality
increases, unless some constraints are imposed on the size and support of
the rules. Although such constraints favour interpretability, they have a negative impact on the
predictive performance of the model, as we show empirically in our work. Additionally, these methods 
do not consider class imbalance issues.

The key idea in our work is to exploit the known advantages of bottom-up learners in imbalanced settings, and
improve their generalization and noise-tolerance through an ensembling
technique that does not sacrifice interpretability.
As a result, we produce a rule-based method that is (i) \textit{versatile} and \textit{effective} in dealing with both
balanced and imbalanced data, (ii) \textit{interpretable}, as it produces small and compact
rule sets, and (iii) \textit{scalable} to big datasets.


\noindent \textbf{Contributions.}
(i) We propose \LIBRE, a novel ensemble method that, unlike other ensemble proposals in
the literature \citep{Cohen1999,Friedman2008,Dembczynski2010} is interpretable.
Each weak learner uses a \textit{bottom-up} approach based on monotone Boolean function
synthesis and generates rules with no assumptions on their size and support. Candidate rules
are then combined with a simple union, to obtain a final interpretable rule set. The idea of ensembling
is crucial to improve generalization, while using \textit{bottom-up} weak learners allows
to generate meaningful rules even when the target class has few available samples. (ii) Our base algorithm
for a weak learner, which is designed to generate a small number of compact rules, is inspired by
\cite{Muselli2005_SC}, but it dramatically improves computational efficiency.
(iii) We perform an extensive experimental validation indicating that \LIBRE scales to large
datasets, has competitive predictive performance compared to state-of-the-art approaches (even black-box models),
and produces few and simple rules, often outperforming existing interpretable models.

\section{BACKGROUND AND DEFINITIONS}
\label{sec:background_and_definitions}
Our methodology 
targets binary classification, although it can be
easily extended to multi-class settings.
For the sake of building interpretable models, we focus on Boolean functions for the mapping between
inputs and labels, which are amenable to a simple interpretation. 

Boolean functions can be used as a model for binary classifiers $f(\mathbf{x}) = y$, 
where $\mathbf{x} \in \{0,1\}^d,\, y \in \{0,1\}$. The function $f$ induces a separation 
of $\{0,1\}^d$ in two subsets $\mathcal{F}$ and $\mathcal{T}$, where 
$\mathcal{F} = \{\mathbf{x} \in \{0,1\}^d : f(\mathbf{x}) = 0\}$ and 
$\mathcal{T} = \{\mathbf{x} \in \{0,1\}^d : f(\mathbf{x}) = 1\}$. We call 
such subsets positive and negative subsets, respectively. 
Clearly, $\mathcal{F} \cup \mathcal{T} = \{0,1\}^d$ corresponds to the full truth 
table of the classification problem.
We restrict the input space $\{0,1\}^d$ to be a \textit{partially ordered set} (\textit{poset}): a 
\emph{Boolean lattice} on which we impose a partial ordering relation. 



\begin{definition}\label{def:poset}
    Let $\bigwedge$, $\bigvee$, $\neg$ be the \AND, \OR, and \NOT logic operators respectively. 
	A \textit{Boolean lattice} is a $5$~tuple $(\{0,1\}^d,\bigwedge,\bigvee,0,1)$. 
	The lack of the $\neg$ operator implies that a lattice is \textbf{not} a Boolean algebra.
	Let $\leq$ be \textit{a partial order relation} such that 
	$\mathbf{x} \leq \mathbf{x}' \iff \mathbf{x} \bigwedge \mathbf{x}' = \mathbf{x}'$. 
	Then, $(\{0,1\}^d,\leq)$ is a poset, a set on which a partial order 
	relation has been imposed.
\end{definition}

The theory of Boolean algebra ensures that the class $\mathcal{B}_d$ of Boolean functions 
$f:\{0,1\}^d \rightarrow \{0,1\}$ can be realized in terms of $\bigwedge$, $\bigvee$, and 
$\neg$. However, if $\{0,1\}^d$ is a Boolean lattice, $\neg$ is not allowed and only a subset 
$\mathcal{M}_d$ of $\mathcal{B}_d$ can be realized. The class $\mathcal{M}_d$ coincides with 
the collection of monotone Boolean functions.
The lack of the $\neg$ operator may limit the family of functions we can reconstruct. 
However, by applying a suitable transformation of the input
space, we can enforce the monotonicity constraint \citep{Muselli2005_UA}. 
As a consequence, it is possible to find a function $\tilde{f} \in \mathcal{M}_d$ that 
approximates $f \in \mathcal{B}_d$ arbitrarly well. 

\begin{definition}\label{def:monotone_f}
    Let $(\mathcal{X},\leq)$ and $(\mathcal{Y},\leq)$ be two posets. 
	Then, $f : \mathcal{X} \rightarrow \mathcal{Y}$ is called \textit{monotone} if 
	$\mathbf{x} \leq \mathbf{x}'$ implies $f(\mathbf{x}) \leq f(\mathbf{x}')$.
\end{definition}

\begin{definition}
    Given $\mathbf{x} \in \{0,1\}^d$, let $\mathcal{I}_m$ be the set of the first $m$ positive 
	integers $\{1,\dots,m\}$. $\mathcal{P}(\mathbf{x}) = \{i \in \mathcal{I}_m : \mathbf{x}(i)=1\}$.
    The inverse of $\mathcal{P}$ is denoted as $p(\mathcal{P}(\mathbf{x}, m)) = \mathbf{x}$.
\end{definition}

\begin{definition}
    Let $\tilde{f} \in \mathcal{M}_d$ be a monotone Boolean function, and $\mathcal{A}$ be a partially 
	ordered set. Then, $\tilde{f}$ can be written as: 
	$\tilde{f}(\mathbf{x}) = \bigvee_{\mathbf{a} \in \mathcal{A}}\bigwedge_{j \in \mathcal{P}(\mathbf{a})} {\mathbf{x}(j)}$.
\end{definition}

The monotone Boolean function $\tilde{f}$ is specified in \emph{disjunctive normal form}
(DNF), and is univocally determined by the set $\mathcal{A}$ and its elements. Thus, given $\mathcal{F}$ 
and $\mathcal{T}$, learning $\tilde{f}$ amounts to finding a particular set of lattice elements $\mathcal{A}$
defining the {\bf boundary} separating positive from negative samples. 

\begin{definition}\label{def:boundary}
    Given $\mathbf{a} \in \{0,1\}^d = \mathcal{T} \cup \mathcal{F}$, if $\mathbf{a} \leq \mathbf{x}$ 
	for some $\mathbf{x} \in \mathcal{T}$, and $\nexists \mathbf{y} \in \mathcal{F} : \mathbf{a} \leq \mathbf{y}$, and 
	$\exists \mathbf{y} \in \mathcal{F} : \mathbf{b} \leq \mathbf{y}\,,\forall \mathbf{b} < \mathbf{a}$, 
	then $\mathbf{a}$ is a \textit{boundary point} for $(\mathcal{T},\mathcal{F})$.
    The set $\mathcal{A}$ of boundary points defines the \textit{separation boundary}. If $\mathbf{a}' \nleq \mathbf{a}''$ and 
	$\mathbf{a}'' \nleq \mathbf{a}'\,,\forall \mathbf{a}', \mathbf{a}'' \in \mathcal{A}, \mathbf{a}' \neq \mathbf{a}''$, 
	then the separation boundary is \textit{irredundant}.
\end{definition}

In other words, a boundary point is a lattice element that is smaller than or equal to at least one 
positive element in $\mathcal{T}$, but larger than all negative elements $\mathcal{F}$. 
In practical applications, however, we usually have access to a subset of the whole space, $\mathcal{D}_+ \subseteq \mathcal{T}$
and $\mathcal{D}_- \subseteq \mathcal{F}$. The goal of the algorithms we present next is to approximate the boundary $\mathcal{A}$, given
$\mathcal{D}_+$ and $\mathcal{D}_-$. We show that boundary points, and binary samples in general,
naturally translate into classification rules.
Indeed, let $\mathcal{R}$ be the set of rules corresponding to the discovered boundary. $\mathcal{R}(\cdot)$ represents
a binary classifier: $\mathcal{R}(\textbf{x})= \{1 \text{ if } \exists r \in \mathcal{R} : r(\textbf{x}) = 1 \text{; } 0 \text{ otherwise}\}$.
Then, \textbf{x} is classified as positive if there is at least one rule in $\mathcal{R}$ that is true for it.

\section{BOOLEAN RULE SETS}
\label{sec:algorithms}
We presented a theoretical framework that casts binary classification
as the problem of finding the boundary points for
$\mathcal{D}_+ \subseteq \mathcal{T}$ and  $\mathcal{D}_- \subseteq \mathcal{F}$.
Next, we use such framework to design our interpretable classifier.

First, we describe a base, bottom-up method -- which will be later used as a weak learner -- that
illustrates how to move inside the boolean lattice to find boundary points.
However, the base method does not scale to large datasets, and tends to overfit.
Thus, we present \LIBRE, an ensemble classifier that overcomes such limitations by running on randomly selected
subset of features. \LIBRE is interpretable
because it combines the output of an ensemble of weak learners with a simple union operation.
Finally, we present a procedure to select a subset of the generated points -- the ones with the best predictive
performance -- and reduce the complexity of the boundary.

We assume that the input dataset is a poset and that the function we want to reconstruct
is monotone. This is ensured by applying inverse-one-hot-encoding
on discretized features, and concatenating the resulting binary features, as done in \cite{Muselli2006_SNN}.
Given $z \in \mathcal{I}_m=\{1,...,m\}$, inverse-on-hot encoding produces a binary string 
$\mathbf{b}$ of length $m$, where $b(i)=1$ for $i \neq z$, $b(i)=0$ for $i=z$. More details can be found 
in the supplementary material.

\begin{example}
	Consider a dataset with two continuous features, $f_1$ and  $f_2$, both taking values in the domain 
	$[0,100]$. Suppose that, a discretization algorithm outputs the following discretization ranges for the 
	two features: $[[0,40), [40,100]]$ and $[[0,30), [30,60), [60, 100]]$ respectively. Once all records are 
	discretized, we apply inverse one-hot encoding, as previously defined. For example, 
	$f_1=33.1, f_2=44.7$ is first discretized as $f'_1=1, f'_2=3$, and then binarized as $01\;101$. In other words,
	each feature of a record is encoded with a number of bits equal to its discretized domain, and can have only 
	one bit set to zero.
\end{example}

%
%


\subsection{The Base, Bottom-up Method}\label{sec:naive}
We develop an approximate algorithm that learns the set $\mathcal{A}$ for $(\mathcal{D}_+, \mathcal{D}_-)$.
The algorithm strives to find lattice elements such that both $|\mathcal{A}|$ and 
$|\mathcal{P}(\mathbf{a})|\,, \forall \mathbf{a} \in \mathcal{A}$ are small, translating in a 
small number of sparse boundary points (short rules).


\noindent \textbf{Algorithm Design.} To proceed with the presentation of our algorithm, we need the following definitions:
\begin{definition}\label{def:cover} 
    Given two lattice elements $\mathbf{x}, \mathbf{x}' \in \{0,1\}^d$, we say that $\mathbf{x}'$ 
	\emph{covers} $\mathbf{x}$, if and only if $\mathbf{x}' \leq \mathbf{x}$, 
\end{definition}
\begin{definition}\label{def:flip}
    Given a lattice element $\mathbf{x} \in \{0,1\}^d$, \emph{flipping off} the $k$-th element of 
	$\mathbf{x}$ produces an element $\mathbf{z}$ such that $\mathbf{z}(i) = \mathbf{x}(i)$ for 
	$i \neq k$ and $\mathbf{z}(i) = 0$ for $i = k$.
\end{definition}
\begin{definition}\label{def:conflict}
    Given a positive binary sample $\mathbf{x} \in \mathcal{D}_+$, we say that a flip-off 
	operation produces a \emph{conflict} if the lattice element $\mathbf{z}$ resulting from the flip-off 
	is such that $\exists \mathbf{x}' \in \mathcal{D}_- : \mathbf{z} \leq \mathbf{x}'$. 
\end{definition}

Then, a boundary point 
is a lattice element that covers at least one 
positive sample, and for which a flip-off operation would produce a conflict, as defined above.

\begin{algorithm}
    \caption{FindBoundary}
    \label{alg:fb}
	\scriptsize
    Set $\mathcal{A}=\emptyset$ and $\mathcal{S}=\mathcal{D}_+$\;
    \While{$\mathcal{S} \neq \emptyset$}{
        Choose $\mathbf{x} \in \mathcal{S}$\;
        Set $\mathcal{I} = \mathcal{P}(\mathbf{x})$, $\mathcal{J} = \emptyset$\;
        \texttt{FindBoundaryPoint}($\mathcal{A}, \mathcal{I}, \mathcal{J}$)\;
        Remove from $\mathcal{S}$ the elements covered by $\mathbf{a},\, \forall \mathbf{a} \in \mathcal{A}$\;
    }
\end{algorithm}

\Cref{alg:fb} presents the main steps of our algorithm, where $\mathcal{A}$ is the boundary set 
and $\mathcal{S} = \{ \mathbf{s} \in \mathcal{D}_+ : \nexists \mathbf{a} \in \mathcal{A}, \mathbf{a} \leq \mathbf{s} \}$ 
is the set of elements in $\mathcal{D}_+$ that are not covered by a boundary point in 
$\mathcal{A}$. 
$\mathcal{I}$ is the set of indexes of the components of the 
current positive sample $\mathbf{x}$ that can be flipped-off, and $\mathcal{J}$ is the set of indexes 
that cannot be flipped-off to avoid a conflict with $\mathcal{D}_-$. Until $\mathcal{S}$ 
is not empty, an element $\mathbf{x}$ is picked from $\mathcal{S}$. Then, the procedure 
\texttt{FindBoundaryPoint} is used to generate one or more boundary points by flipping-off the candidate bits of 
$\mathbf{x}$. According to \cref{def:flip}, a boundary point is generated when an additional 
flip-off would lead to a conflict, given \cref{def:conflict}. When the \texttt{FindBoundaryPoint} procedure 
completes its operation, both $\mathcal{A}$ and $\mathcal{S}$ are updated.

\begin{example}
    Let $\mathcal{D}_+ = \{11001\}$ and $\mathcal{D}_- = \{01101, 01101\}$. Take the positive 
	sample 	$11001$, for which $\mathcal{I} = \{1,2,5\}$ and $\mathcal{J} = \emptyset$. Suppose 
	that \texttt{FindBoundaryPoint} flips-off the bits in $\mathcal{I}$ from left to right. Flipping-off 
	the first bit generates $01001 \leq 01101 \in \mathcal{D}_-$. The first bit is moved 
	to $\mathcal{J}$ and kept to 1. Flipping-off the second bit generates $10001 \leq 10101 \in \mathcal{D}_-$. 
	Also the second bit is moved to $\mathcal{J}$. We finally flip-off the last bit and obtain 
	$11000$ that is not in conflict with any 	element in $\mathcal{D}_-$. $11000$ is therefore 
	a boundary point for $(\mathcal{D}_+,\mathcal{D}_-)$.
\end{example}

If we think about binary samples in terms of rules, a positive sample can be seen as a maximally-specific
rule, with equality conditions on the input features (the value that particular feature takes on 
that particular sample). Flipping-off bits is nothing more than generalizing that rule. Our goal is to 
do as many flip-off operations as possible before running into a conflict.

Retrieving the complete set of boundary points requires an exhaustive search, which is expensive, 
restricting its application to small, low-dimensional datasets. It is easy to show that the 
computational complexity of the exhaustive approach is $O(n^2 2^d)$, where $n$ is the number 
of \textit{distinct} training samples, and $d$ is the dimension of the Boolean lattice. 
In this work, we propose an \emph{approximate heuristic} for the \texttt{FindBoundaryPoint} procedure. 

\noindent \textbf{Finding Boundary Points.} The key idea is to find a subset of all possible boundary points, steering their selection through
a measure of their quality. A boundary point is considered to be ``good'' if it contributes to 
decreasing the complexity of the resulting boundary set, which is measured in terms of its cardinality 
$|\mathcal{A}|$ and the total number of positive bits $\sum_{\mathbf{a} \in \mathcal{A}} |\mathcal{P}(\mathbf{a})|$. 
In practice, $|\mathcal{A}|$ can be decreased by choosing boundary points that cover the largest number of elements in $\mathcal{S}$. 
To do this, we iteratively select the best candidate index $i \in \mathcal{I}$ according to a measure of 
potential coverage. Decreasing $\sum_{\mathbf{a} \in \mathcal{A}} |\mathcal{P}(\mathbf{a})|$ implies finding boundary points with low number of $1$s.

Before proceeding, we define a notion of distance between lattice elements:

\begin{definition}\label{def:lower_distance}
	Given $\mathbf{x}, \mathbf{x}' \in \{0,1\}^d$, the \textit{distance} $d_l(\mathbf{x},\mathbf{x}')$ 
	between $\mathbf{x}$ and $\mathbf{x}'$ is defined as: 
	$d_l(\mathbf{x},\mathbf{x}') = \sum_{i=1}^d |\mathbf{x}(i)-\mathbf{x}'(i)|_+$, 
	where $|\cdot|_+$ is equal to $1$ if $(\cdot) \geq 0$, $0$ otherwise.
\end{definition}

\begin{definition}
    In the same way, we can define the distance between a lattice element $\mathbf{x}$ and a set 
	$\mathcal{V}$ as:
	$d_l(\mathbf{x},\mathcal{V}) = \min_{\mathbf{x}' \in \mathcal{V}}{d_l(\mathbf{x},\mathbf{x}')}$.
\end{definition}

Every boundary point $\mathbf{a}$ for $(\mathcal{D}_+,\mathcal{D}_-)$ has distance $d_l(\mathbf{a},\mathcal{D}_-) = 1$; 
in fact, boundary points are all lattice elements for which a flip-off would generate a conflict. 
In the iterative selection process of the best index $i \in \mathcal{I}$ to be flipped-off, indexes 
having high $d_l(p(\mathcal{I} \cup \mathcal{J}), {\mathcal{D}_-}_i^0)$ are preferred, where 
${\mathcal{D}_-}_i^0 = \{\mathbf{x} \in \mathcal{D}_- : \mathbf{x}(i) = 0\}$, because 
they are the ones that contribute most to reduce the number of $1$s of a potential boundary point.

\begin{algorithm}
    \caption{FindBoundaryPoint$(\mathcal{A}, \mathcal{I}, \mathcal{J})$}
    \label{alg:fbp}
	\scriptsize
    For each $i \in \mathcal{I}$ compute $|\mathcal{S}_i^0|, |{\mathcal{D}_+}_i^0|, d_l(p(\mathcal{I} \cup \mathcal{J}), {\mathcal{D}_-}_i^0)$\;
    \While{$\mathcal{I} \neq \emptyset$}{
        Move from $\mathcal{I}$ to $\mathcal{J}$ all $i$ with $d_l(p(\mathcal{I} \cup \mathcal{J}), {\mathcal{D}_-}_i^0) = 1$\;
        \If{$\mathcal{I} = \emptyset$} {
            \text{\textbf{break}}\;
        }
        Choose the best index $i \in \mathcal{I}$\;
        Remove $i$ from $\mathcal{I}$\;
        For each $i \in \mathcal{I}$ update $d_l(p(\mathcal{I} \cup \mathcal{J}), {\mathcal{D}_-}_i^0)$\;
    }
    \If{there is no $\mathbf{a} \in \mathcal{A} : p(\mathcal{J}) \geq \mathbf{a}$}{
        Set $\mathcal{A} = \mathcal{A} \cup p(\mathcal{J})$\;
    }
\end{algorithm}

Algorithm~\ref{alg:fbp} illustrates our approximate procedure, where
$\mathcal{S}_i^0 = \{\mathbf{s} \in \mathcal{S} : \mathbf{s}(i) = 0\}$ and ${\mathcal{D}_+}_i^0 = \{\mathbf{t} \in \mathcal{D}_+ : \mathbf{t}(i) = 0\}$ 
are proxies for the potential coverage of flipping-off a given bit $i$.
The first step of the algorithm computes, for each index $i \in \mathcal{I}$, the terms 
$|\mathcal{S}_i^0|$ and $|{\mathcal{D}_+}_i^0|$ indicating its potential coverage, and $d_l(p(\mathcal{I} \cup \mathcal{J})$. 
Until the set $\mathcal{I}$ is not empty, indexes inducing a unit distance to $\mathcal{D}_-$ are 
moved to $\mathcal{J}$. Then, we choose the best index $i_{best}$ among the remaining indices in 
$\mathcal{I}$, using our \textbf{greedy heuristics}: we can chose to optimize either for the tuple 
$H_1=(|\mathcal{S}_i^0|, |{\mathcal{D}_+}_i^0|, d_l(p(\mathcal{I} \cup \mathcal{J}), {\mathcal{D}_-}_i^0))$ 
or for the tuple $H_2=(d_l(p(\mathcal{I} \cup \mathcal{J}), {\mathcal{D}_-}_i^0), |\mathcal{S}_i^0|, |{\mathcal{D}_+}_i^0|)$.
$H_1$ prioritizes a lower number of boundary points, while $H_2$ tends to generate boundary points with fewer $1$s. 

When $\mathcal{I}$ is empty, $p(\mathcal{J})$ is added to the 
boundary set $\mathcal{A}$ if it does not contain already an element covering $p(\mathcal{J})$. 
Note that, in \cref{alg:fbp}, the distance is computed only once, and updated at each iteration. 
This is because only one bit is selected and removed from $\mathcal{I}$; then, 
$p(\mathcal{I} \cup \mathcal{J})_{new} = p((\mathcal{I} \cup \mathcal{J})_{old} \setminus \{i\})$. 
Formally, we apply \cref{def:lower_distance} exclusively for $i=i_{best}$.

\begin{example}
    Let $\mathcal{D}_+ = \{10101, 01101, 01110\}$ and $\mathcal{D}_- = \{10110, 11010\}$. 
	We describe the procedure for few steps and only for the first positive sample $10101$. 
	Suppose to optimize the tuple 
	$(|\mathcal{S}_i^0|, |{\mathcal{D}_+}_i^0|, d_l(p(\mathcal{I} \cup \mathcal{J})))$. 
	For $10101$ we have $\mathcal{I} = \{1,3,5\}$ and $\mathcal{J} = \emptyset$. At the beginning 
	$\mathcal{S}=\mathcal{D}_+$. $|{\mathcal{D}_+}_1^0| = 2, |{\mathcal{D}_+}_3^0| = 0, |{\mathcal{D}_+}_5^0| = 1$. 
	${\mathcal{D}_-}_1^0 = \emptyset, {\mathcal{D}_-}_3^0 = \{11010\}, {\mathcal{D}_-}_5^0 = \{10110, 11010\}$. 
	Consequently: $d_l(p(\mathcal{I} \cup \mathcal{J}), {\mathcal{D}_-}_1^0) = undefined$, 
	$d_l(p(\mathcal{I} \cup \mathcal{J}), {\mathcal{D}_-}_3^0) = 2$, $d_l(p(\mathcal{I} \cup \mathcal{J}), {\mathcal{D}_-}_5^0) = 1$. 
	Bit 5 is moved to $\mathcal{J}$. Bit 1 has the higher value of $|{\mathcal{D}_+}_i^0|$ and is selected 
	as best candidate to be flipped-off. The distance is recalculated 	and the procedure continues until the set 
	of candidate bits $\mathcal{I}$ is empty.
\end{example}

The algorithmic complexity of \cref{alg:fb}, when it runs \cref{alg:fbp}, is $O(n^2 d^2)$.
This is faster than the exhaustive algorithm, and better than the $O(n^2 d^3)$ complexity of \cite{Muselli2005_SC}.
We also point out that most sequential-covering algorithms repeatedly remove the samples covered by the new rules, 
forcing the induction phase to work in a more partitioned space with less data, especially affecting minority rules, 
which already rely on few samples. The problem is mitigated in our solution: despite $\mathcal{S}$ cannot avoid 
this behavior, our heuristics keep a global and constant view of both $\mathcal{D}_{-}$, 
in the conflict detection, and $\mathcal{D}_{+}$, in the discrimination of the best bits to flip.


\noindent \textbf{From Boundary Set To Rules.} Each element $\mathbf{a}$ of the boundary set
$\mathcal{A}$ can be practically seen as the antecedent
of an \textbf{if-then} rule having as target the positive class. When a binary sample $\mathbf{x}$ is 
presented to $\mathbf{a}$, the rule outputs $1$ only if $\mathbf{x}$ has a $1$ in all positions where 
$\mathbf{a}$ has value $1$, that is if $\mathbf{a} \leq \mathbf{x}$. Then, the antecedent of the rule 
is expressed as a function of the input features in the original domain. 

\begin{example}
	Consider a dataset with two continuous features, $f_1$ and  $f_2$, discretized as follows: 
	$[[0,40), [40,100]]$ and $[[0,30), [30,60), [60, 100]]$ respectively. Let's assume that our algorithm 
	outputs a boundary set $\mathcal{A} = \{01\;100\}$. From the boundary point 
	we obtain a rule as follows: the first two bits referring to feature $f_1$ -- $01$ -- are mapped to 
	``\textbf{if} $f_1 \in [0,40)$'', while the bits referring to $f_2$ -- $100$ -- are mapped to 
	``\textbf{if} $f_2 \in [30,100]$'', where the two consecutive intervals have been combined. The zeros determine
	the ranges in the if conditions. The final rule is therefore ``\textbf{if} $f_1 \in [0,40)$ and $f_2 \in 
	[30,100]$ \textbf{then} label = 1''.
\end{example}

\subsection{The LIBRE Method}\label{sec:ensemble}
The base approach generates boundary points by generalizing input samples, i.e., by flipping-off positive 
bits if no conflict with 
negative samples is encountered. The hypothesis underlying this
procedure is that when no conflicts are found, a boundary point induces a valid rule. However, such rule
might be violated when used with unseen data. 
Stopping the flipping-off procedure as soon as a single conflict is found has two main effects: i) we 
obtain very specific rules, that might be simplified if the approach could tolerate a 
limited number of conflicts; ii) the rules cover no negative samples in the training set and tend to overfit. 

To address these issues, a simple method would be to introduce a measure for the number of conflicts and 
use it as an additional heuristic in the learning process. However, this would dramatically increase the 
complexity of the algorithm.

A more natural way to overcome such challenges is to make the algorithm directly work on (random) subsets 
of features; in this way, the learning process produces more general rules by construction. Randomization 
is a well-known technique to implement ensemble methods that provide superior classification accuracy, as 
demonstrated, for example, in random forests \citep{Ho1998,Breiman2001}. 
By using randomization, we can directly use the methodology described in the previous sections, without 
modifying the search procedure. The new approach -- \LIBRE~-- is an interpretable ensemble of rules that 
operates on a randomized subset of features. 

Formally, let $E$ be the number of classifiers in the ensemble. For each classifier $j \in \{1, \dots, E\}$, 
we randomly sample $k_j$ features of the original space and run \cref{alg:fb} to produce a boundary set 
$\mathcal{A}_j$ for the reduced input space. $A_j$ can be generated in parallel, since weak learners 
are independent from each other. 
At this point, to make the ensemble interpretable, 
we crucially do not apply a voting (or aggregation) mechanism to produce the final class prediction, but 
we do a simple union, such that $\mathcal{A} = \bigcup_{j=1}^{E} \mathcal{A}_j$. 

We note that \LIBRE addresses the problems outlined above, as we show experimentally. By training an ensemble of 
\emph{weak learners} that operate on a small subset of features, we artificially inflate the probability of finding 
negative examples. Each weak learner is constrained to run on less features not only reducing the impact of $d$ 
on the execution time, but also having an immediate effect on the interpretability of the model that is forced 
to generate simpler rules, exactly because it operates on fewer input features. 

Note that there are no guarantees that elements of $\mathcal{A}_j$ will actually be boundary points in the 
full feature space: weak learners have only a partial view of the full input space and might generate rules 
that are not globally true. Thus, it is important to filter out the points that are clearly far from 
the boundary by using the selection procedure described in the next section. 

\subsection{
Producing The Final 
Boundary}\label{sec:simplify}
The model learned by our greedy heuristic materializes as a set $\mathcal{A}$, which might contain 
a large number of elements and, in case of \LIBRE, it might also contain elements that cover many
negative samples. In this section, we explain how to produce a boundary set $\mathcal{A}^*$ with a 
good tradeoff between complexity and predictive performance. This can be cast as a \textit{weighted 
set cover} problem. Since exploring all possible subsets of elements in $\mathcal{A}$ can be 
computationally demanding, we use a standard greedy weighted set cover algorithm.

Each element $\mathbf{a} \in \mathcal{A}$ is assigned a weight that is proportional to the number of positive 
and negative covered samples. The importance of the two contributions is governed by 
a parameter $\alpha$. At each iteration, the element $\mathbf{a}$ with the highest 
weight is selected; if there is more than one, the element with the highest number of zeros is preferred. 
All samples that are covered by the selected element are removed, and the weights are recalculated. 
The process continues until either all samples are a covered or a stopping condition is met.

Before running the selection procedure, with the aim of speeding up execution times, we eventually apply a 
filtering procedure to reduce the size of the initial set to a small number of good candidates: 
as proposed by \cite{Gu2003}, we select the top $K$ rules according to \textit{exclusiveness} and \textit{local support},
that are more sensible than confidence and support for imbalanced settings.




%

\section{EXPERIMENTS}
\label{sec:experiments}
We evaluate \LIBRE in terms of predictive performance, interpretability, and 
scalability, and compare it with other rule-based methods and black-box models.


\begin{table}[t]
    \scriptsize
    \begin{center}
		\begin{tabular}{l|c|c|c}
			\toprule
			\textbf{Dataset} 	& \textbf{\#records} 	& \textbf{\#features}	& \textbf{imbalance\_ratio}\\
			\midrule
			\Adult 				& 48'842				& 14 					& .23 					\\
			\Australian			& 690					& 14					& .44					\\
			\Bank 				& 45'211 				& 17 					& .12 					\\
			\Ilpd 				& 583 					& 10 					& .28 					\\
			\Liver 				& 345 					& 5 					& .51 					\\
			\Pima				& 768					& 8						& .35					\\
			\Transfusion 		& 748 					& 5						& .24					\\
			\Sapclean			& 287'031				& 45					& .01					\\
			\Sapfull			& 1'554'227				& 45					& .01					\\
			\bottomrule
		\end{tabular}
    \end{center}
	\caption{Characteristics of evaluated datasets.}
    \label{tab:datasets}
\end{table}

\begin{table*}[t!]
    \scriptsize
    \begin{center}
		\begin{tabular}{l|c|c|c|c|c|c|c|c|c}
			\toprule
			Dataset & 		\RBFSVM				& \RF 				& \DT 				& \RIPPERK 			& \MODLEM			& \SBRL 			& \BRS 				& \underline{\LIBRE} 	& \underline{\LIBRE 3}	\\
			\midrule		
			\Adult          & .62(.01)			& .68(.01) 			& .68(.01)			& .59(.02) 			& .66(.01)			& .68(.01) 			& .61(.01)			& \textbf{.70(.01)} 	& .62(.01)				\\
			\Australian     & .83(.02)			& \textbf{.86(.02)}	& .84(.02)			& .85(.02) 			& .68(.28)			& .82(.03) 			& .83(.03)			& .84(.03) 				& .84(.03)				\\
			\Bank           & .46(.01)			& .50(.01) 			& .50(.01)			& .44(.04) 			& .50(.03)			& .50(.02) 			& .32(.05)			& \textbf{.55(.01)} 	& .44(.01)				\\
			\Ilpd           & .47(.02)			& .44(.08) 			& .42(.10)			& .20(.11) 			& .48(.08)			& .14(.13) 			& .09(.08)			& \textbf{.54(.06)} 	& .52(.04)				\\
			\Liver			& .58(.08)			& .58(.07) 			& .56(.10)			& .59(.04) 			& .58(.07)			& .54(.03) 			& .61(.05)			& .60(.07) 				& \textbf{.63(.06)}		\\
			\Pima			& .61(.04)			& .63(.04) 			& .60(.01)			& .60(.03) 			& .38(.18)			& .61(.07) 			& .03(.03)			& \textbf{.64(.05)} 	& .\textbf{.64(.05)}	\\
			\Transfusion    & .41(.07)			& .35(.06) 			& .35(.05)			& .42(.10) 			& .42(.08)			& .05(.10) 			& .04(.05)			& \textbf{.49(.12)} 	& \textbf{.49(.12)}		\\
			\Sapclean		& .93(.02)			& .93(.01)			& .85(.03)			& .86(.02)			& .88(.01)			& .90(.01)			& .68(.03)			& \textbf{.95(.02)}		& .72(.03)				\\
			\Sapfull		& -					& -					& -					& -					& -					& .81(.02)			& -					& \textbf{.89(.03)}		& .68(.04)				\\
			\midrule		
			Avg Rank        & 4.7(1.2)			& 3.3(1.6)			& 4.9(2.1)			& 5.3(2.1)			& 4.9(2.2)			& 5.2(2.8)			& 7.2(2.5)			& \textbf{1.4(0.8)}		& 3.6(2.6)				\\
			\bottomrule
		\end{tabular}
    \end{center}
	\caption{F1-score (st. dev. in parenthesis).}
    \label{tab:f1-score}
\end{table*}

\begin{table*}[t]
    \scriptsize
	\begin{center}
        \begin{tabular}{l|c|c|c|c|c|c|c}
			\toprule
            Dataset 		& \DT				& \RIPPERK 			& \MODLEM 				& \SBRL 			& \BRS 				& \underline{\LIBRE} 		& \underline{\LIBRE 3} 		\\
            \midrule																						
            \Adult			& 287.8(6.5)		& 21.4(5.2)			& 4957.8(36.3) 			& 71.4(2.1)			& 10.0(3.3)			& 14.0(2.1)					& \textbf{3.0(0.0)}			\\
            \Australian		& 4.0(0.0)			& 3.8(1.2)			& 86.6(3.2) 			& 5.8(0.7)			& \textbf{1.8(0.4)}	& 2.4(1.4)					& 2.2(0.7)					\\
			\Bank			& 545.4(18.3)		& 9.0(1.8)			& 3722.6(25.5)			& 61.2(5.5)			& 4.8(1.2)			& 15.0(1.1)					& \textbf{2.0(0.6)}			\\
			\Ilpd			& 80.6(30.2)		& \textbf{1.0(0.6)}	& 128.2(7.8)			& 4.8(0.7)			& \textbf{1.0(0.0)}	& 4.4(2.3)					& 2.2(0.4)					\\
			\Liver			& 84.4(15.2)		& 1.4(0.8)			& 98.4(1.6)				& 4.0(0.6)			& \textbf{1.0(0.0)}	& 3.4(1.9)					& 2.8(0.4)					\\
			\Pima			& 84.8(43.1)		& 2.4(2.4)			& 151.8(7.6)			& 8.4(0.5)			& \textbf{1.0(0.0)}	& 1.6(1.0)					& 1.6(1.0)					\\
			\Transfusion	& 100.2(48.4)		& 1.8(0.4) 			& 125.8(6.1) 			& 4.4(0.8)			& \textbf{1.0(0.0)}	& 1.2(0.4)					& 1.2(0.4)					\\
            \Sapclean		& 622.4(51.9)		& 19.3(3.6)			& 3944.5(18.8)			& 47.7(4.4)			& 20.2(3.5)			& 13.0(2.4)					& \textbf{3.0(0.0)}			\\
			\Sapfull		& -					& -					& -						& 56.4(4.6)			& -					& 17.5(5.2)					& \textbf{3.0(0.0)}			\\
			\midrule					
			Avg Rank		& 5.7(0.7)			&3.2(1.0)			& 6.7(0.9)				& 4.9(0.7)			& 1.9(1.2)			& 2.9(0.9)					& \textbf{1.8(0.8)}			\\
			\bottomrule
        \end{tabular}
    \end{center}
	\caption{\#rules (st. dev. in parenthesis).}
    \label{tab:rules}
\end{table*}

\vspace{2mm}
\noindent \textbf{Datasets.} We report the results for seven publicly available datasets from the UCI 
repository and two real industrial IT datasets -- proprietary of \Sap. Results on other UCI datasets are
in the supplementary material. These datasets cover several domains, have different imbalance ratios, 
number of records and features, as summarized in \Cref{tab:datasets}. Some of these datasets have been used 
to evaluate methods for class imbalance \citep{VanHulse2007} and present characteristics that make them difficult 
to learn: overlapping classes, noisy and rare examples. All datasets have, or were transformed to 
have, a binary class. 
The \Sap datasets consist of monitoring data collected across database systems.
They consists of 45 features, hand-crafted by domain experts based on low-level system metrics. 
\Sap runs a predictive maintenance system on this data and notifies customers who confirm or
discard the warnings: we use these as binary labels. \Sapclean is the clean version of \Sapfull, where we removed records
with at least one missing value.

\vspace{2mm}
\noindent \textbf{Comparison With Other Methods.} We compare \LIBRE with two recent works: Scalable Bayesian 
Rule Lists (\SBRL) \citep{Yang2017_sbrl} and Bayesian Rule Sets (\BRS) \citep{Wang2017_brs}. We also report 
the results for a {\sc weka} implementation of \RIPPERK \citep{Cohen95} and \MODLEM \citep{Modlem2001} -- as 
representative of top-down and bottom-up approaches ~-- and {\sc scikit-learn} implementations of Decision 
Tree (\DT) \citep{cart84}, Support Vector Machine with RBF kernel (\RBFSVM) \citep{CortesV95}), 
and random forests (\RF) \citep{Breiman2001}. \RBFSVM and \RF are selected as popular black-box models; \RF 
is also a representative ensemble method. Other relevant methods are not publicly available
(\CG \citep{Dash2018}), do not work properly (\IDS \citep{Lakkaraju2016_ids}), or are only partially
implemented (\BRACID \citep{Napierala2012Bracid}).

\vspace{2mm}
\noindent \textbf{Parameter Tuning.} All results refer to stratified 5-fold cross validation, where the same 
splits are used for all tested methods. The initial set of candidate rules for \SBRL and \BRS is generated 
by running \textsc{FP Growth} with a minimum support of 1 and a maximum mining length of 5. We also optimize 
\BRS and \SBRL's prior hyperparameters by cross validation. For \BRS, we run 2 chains of 500 iterations. For \RIPPERK, 
we change the number of optimization steps between 1 and 5, and activate pruning. For \MODLEM, we try
all available classification strategies and condition measures. For \RBFSVM, 
we optimize $C$ and $\gamma$. For \DT and \RF, we optimize the maximum depth in $\{5, 10, 20, None\}$, we 
tried all possible options for max\_features and use a number of trees in $\{20, 50, 100\}$ for \RF. For \LIBRE, 
we vary the number of weak learners in $E \in \{5, 20, 50\}$. Each weak learner uses up to 5 features. Additionally,
we try the two heuristics $H_1$ and $H_2$ to generate rules and vary $\alpha$ in $\{.5, .7, .9\}$ for weighted 
set cover. Parameters not reported above are all fixed to recommended or default values.

\vspace{2mm}
\noindent \textbf{Data Preprocessing.} Before running \RBFSVM, we apply standardization to the input 
data to get better results. The remaining methods have no benefits from standardization in our experiments. 
For \SBRL and \LIBRE, we apply \textit{ChiMerge} discretization algorithm \cite{Kerber1992} with a discretization 
threshold in $\{6, 4.6, 4\}$; in \BRS, discretization is instead controlled by an internal parameter. 
In both cases, discretization is optimized during training. The remaining algorithms have no explicit need 
for discretization. For the methods requiring binarization, we apply \textit{one-hot encoding}, except for 
\LIBRE that uses \textit{inverse one-hot encoding}.

\vspace{2mm}
\noindent \textbf{Evaluation Metrics.} We use F1-score to compare the predictive performance of the classifiers,
as it is well-suited to evaluate the capability to characterize the target class both in balanced and imbalanced 
settings. For rule-based methods, we use standard metrics from the literature to evaluate the interpretability of 
the rule sets, namely the \emph{number of rules} that implement a model, and the \emph{average number of atoms per 
rule}. For \DT, we extract the rules following the paths from root to leaves: this captures the perception of a user 
who looks at the tree to understand the output of the model. For \SBRL, the number of atoms in a rule is equal to 
the sum of the atoms in the previous rules, highlighting the fact that a user has to go through all the rules up to 
the one that returns the label. For all rule-based methods, we change inequalities ($<, \leq, >, \geq$) to ranges
to have a fair comparison. For example, $f_1 \geq 3$ is converted to $f_1 \in [3, max]$.

\vspace{2mm}
\noindent \textbf{Predictive Performance Evaluation.}
\Cref{tab:f1-score} shows the means and standard deviations of the F1-score for the tested algorithms (best 
results in bold) and the rank of their average performance, where the same splits are used for all tested 
methods. We additionally report the results for \LIBRE when it is constrained to generate at most 3 rules (\LIBRE 3). 

If we look at the average rank, \LIBRE emerges as the best method, beating both \RBFSVM and \RF, 
demonstrating its versatility in both balanced and imbalanced settings. \LIBRE 3 is still better than the other rule-based 
competitors, although being constrained to generate at most 3 rules. 
\DT, \MODLEM, \SBRL and \RIPPERK show very similar performance, even if \MODLEM is usually worse for
balanced settings. \BRS is the worst method in terms of predictive performance. 

Focusing more on the single datasets, we can see that, 
except for \Australian, \LIBRE obtains consistently the highest F1-score. In \Bank, \Ilpd, and \Transfusion the gap 
between \LIBRE and the closest competitor is significant; the gap is even larger in comparison to alternative 
rule-based methods. For the remaining datasets, the differences with the competitors are less pronounced but 
still significant. 
In particular, \Ilpd seems to be very problematic for most of the tested methods: \RIPPERK, 
\BRS and \SBRL do not learn anything useful about the positive class; \MODLEM performs marginally better. From
a deeper analysis, it emerges that \Ilpd is an imbalanced dataset with overlapping classes: 
rules learned by \LIBRE have an error rate close to 50\% on the training set, consequence of the class imbalance. \RIPPERK is
not able to learn these rules, whereas the selection stage of \BRS and \SBRL does not include such rules in the final set even 
when they are in the set of candidate mined rules.

With \Sapclean, \LIBRE 3 performs better than \BRS but limiting
the number of rules to 3 causes a significant drop in F1-score w.r.t. \LIBRE. 
The situation is different for \Sapfull, the original version of the dataset containing also missing values. From \cref{tab:datasets}, \Sapfull is more
than five times bigger than \Sapclean, indicating that missing values are not a negligible problem in real scenarios.
A method that runs without additional preprocessing is thus truly desirable. Only
\LIBRE and \SBRL fit this requirement, while 
\RIPPERK, 
\BRS, and \MODLEM natively manage missing values for categorical features only, but require an additional preprocessing 
for continuous features. Despite the huge number of missing values, results for \LIBRE are comparable to other rule-based methods when executed on \Sapclean.

\vspace{2mm}
\noindent \textbf{Interpretability Evaluation.}
Next, using \cref{tab:rules}, we evaluate interpretability in terms of quantity and simplicity of rules.
In our analysis, we also refer to \cref{tab:f1-score}, to measure the trade-off that exists between interpretability
and predictive performance. We highlight in bold the most interpretable results.

In terms of number of rules, \LIBRE is better than \RIPPERK on average, indicating that it indeed overcomes 
the limitations of bottom-up learners like \MODLEM, that is instead the worst method together with \DT. 
\SBRL is competitive for small datasets, but the number of rules increases considerably for bigger 
datasets like \Adult, \Bank, and \Sap. Overall, \BRS generates compact rule sets, with only one rule for 
half of the tested datasets. However, we should also notice that, except for \Liver, these are the same 
datasets that give F1-score close to zero. \LIBRE 3 outperforms other methods and produces the most compact rule
sets for the three larger datasets, with a small impact on predictive performance.

In terms of average number of atoms, all the tested methods behave similarly (results in the 
supplementary material). Only \SBRL has issues when the number of rules is significant (like in \Adult, 
\Bank, and \Sap datasets): indeed, in rule lists every rule depends on the previous ones, and the number of atoms easily 
explodes. 

%

\noindent \textbf{Scalability Evaluation.}
\Cref{tab:runtime} shows the run time for \LIBRE and three representative rule-based competitors 
on synthetic balanced datasets with 10 features and a varying number of records: from 
10'000 to 1'000'000. For each configuration, we randomly generate the dataset 3 times and report the average run time and 
standard deviation. All methods are tested with their default parameters and run sequentially, for a fair comparison. 
For \LIBRE, the time refers to one weak learner, which is also a good approximation for the computing time of $E$ parallel weak learners. The symbol ``-'' identifies 
\textit{out-of-memory} errors.

\MODLEM and \BRS fail with an out-of-memory error with 500'000 and 1'000'000 records
datasets. They also show much higher run times for smaller datasets w.r.t. \RIPPERK and \LIBRE, that are instead able 
to complete their training in a few minutes also for the large datasets.

Note that each weak learner in \LIBRE works with $\mathcal{D}_+$ and $\mathcal{D}_-$
that consist of \textit{distinct records}: even if the original dataset has millions of entries, the number of 
binary records processed by the algorithm is much lower, especially when the number of input features of each 
weak learner is relatively low. 
We also point out that, for practical applications 
where interpretability is needed, it is more convenient to limit the number of features and train a bigger ensemble with 
more learners to quickly generate 
understandable rules.

\begin{table}[t]
	\scriptsize
	\begin{center}
		\begin{tabular}{l|c|c|c|c}
			\toprule
			\#records	& \RIPPERK	& \MODLEM	& \BRS		& \LIBRE\\
			\midrule
			10'000  	& 1(0)   	& 14(0) 	& 144(1) 	& 5(0)\\
			100'000 	& 7(3)   	& 2457(89)	& 2994(304)	& 44(5)\\
			500'000 	& 39(25) 	& -			& -         & 209(7)\\
			1'000'000	& 101(31)	& - 		& -			& 399(8)\\
			\bottomrule
		\end{tabular}
	\end{center}
	\caption{Runtime in seconds (st. dev. in parenthesis).}
    \label{tab:runtime}
\end{table}

\section{CONCLUSION}
\label{sec:conclusions}
Model interpretability has recently become of primary importance in many applications. In this work, 
we focused on the task of learning a set of rules which specify, using Boolean expressions, the 
classification model. We devised a practical method based on monotone boolean function synthesis to 
learn rules from data. Our approach uses an ensemble of bottom-up learners that generalizes better 
than traditional bottom-up methods, and that works well for both balanced and imbalanced scenarios.
Interpretability needs can be easily encoded in the rule generation and selection procedure that produces short
and compact rule sets.

Our experiments show that \LIBRE strikes the right balance between predictive performance and 
interpretability, often outperforming alternative approaches from the literature.

For future work, we will extend our model considering noisy labels and a Bayesian formulation.

\noindent \textbf{Acknowledgements}
The authors wish to thank SAP Labs France for support.

\bibliography{ms}
\bibliographystyle{abbrvnat}

\clearpage
\appendix

\section{THE BASE METHOD STEP BY STEP}
\label{sec:walkthrough_example}
In this section, we show in detail the main steps of the base algorithm, by using a concrete example.

Consider the scenario of forecasting the failure condition of an IT system from two values 
representing the $CPU$ and main memory ($MEM$) utilization, as depicted in the first two columns of 
\cref{tab:example}. We assume that $CPU$ and $MEM$ are continuous features with values in the domain 
$[0,100]$. The state of the system is described by a binary {\em Label}, where 1 represents a system 
failure. The example reports eight records, of which two are failures.

\begin{table}[ht]
	\centering
	\scriptsize
	\begin{tabular}{c |c | c || c | c || c || c |}
		\cline{2-7} \multicolumn{1}{c|}{}
		   & {\bf CPU}
		   & {\bf MEM}
		   & {\bf  $r_1$}
		   & {\bf $r_2$}
		   & {\bf String}
		   & {\bf Label}
		\\ \cline{2-7} 
		$t_1$ & 95 & 10 & 3  & 1 & 110 01 & 1
		\\ \cline{2-7} 
		$t_2$ &80 & 10  & 1  & 1 & 011 01 & 0 
		\\ \cline{2-7} 
		$t_3$ &81& 85 & 2 & 2 & 101 10 & 1
		\\ \cline{2-7} 
		$t_4$ & 10  & 85 & 1 & 2 & 011 10 & 0
		 \\ \cline{2-7}
		$t_5$ & 10  & 10 & 1 & 1 & 011 01 & 0
		 \\ \cline{2-7}
		$t_6$ & 82  & 10 & 2 & 1 & 101 01 & 0
		 \\ \cline{2-7}
		$t_7$ & 85  & 10 & 2 & 1 & 101 01 & 0
		 \\ \cline{2-7}
		 $t_8$ & 81  & 10 & 2 & 1 & 101 01 & 0
		 \\ \cline{2-7}
	\end{tabular}
	\caption{Original values from $CPU$ and $MEM$, their mappings to discrete ranges ($r_1,  r_2$), binary encoding, and binary label.}
	\label{tab:example}
\end{table}

\subsection{Discretization And Binarization} 

The first operation to do is discretization. Assume the discretization algorithm identifies three 
intervals for $CPU$ and two intervals for $MEM$, as follows. $CPU$: $[0, 81), [81, 95), [95, max)$. $MEM$: 
$[0, 85), [85, max)$. We can now map the original values to integer values over 
the ranges (1, 2, 3) and (1, 2), as shown in columns $r_1$, $r_2$, respectively. The resulting discretized 
records are then mapped to (inverse one-hot encoded) binary strings of five bits, as recorded in the 
{\em String} column. We also define a partial order relation between binary records, such that 
$\mathbf{x} \leq \mathbf{x}' \iff \mathbf{x} \bigwedge \mathbf{x}' = \mathbf{x}'$. 
Moreover, the application of inverse one-hot encoding ensures that the relation between input features and 
labels is monotone, according to \cref{def:monotone_f} in the main paper. We can give you an intuition through a simple example: consider two binary strings $011$ 
and $110$; we see that $011 \nleq 110$ and $110 \nleq 011$, so the relation always holds, independently 
from the label.

\subsection{Learning The Boundary}
Consider the first positive sample $t_1$ with string 110 01. An exhaustive search strategy would explore 
all possible flipping alternatives for the most general conflict-free binary strings. If, for example, we 
flip-off the first bit we obtain 010 01 $<= t_2$: we have therefore a conflict. If, for example, we keep 
the first bit at 1 and flip-off the second bit, we obtain 100 01, which is in conflict with $t_6 - t_8$. 
Finally, if we flip-off the last bit, we obtain 110 00, which has no conflict: this is a candidate boundary 
point. If we repeat the same procedure for $t_3$, after flipping-off the third bit, we obtain another 
boundary point 100 10.

\begin{figure}[ht]
	\centering
	\scriptsize
	\begin{tikzpicture}
		\node (a1) at (-2,1)	{$t_1$: 110 01$^+$};
		\node (b1) at (0,1) 	{$t_3$: 101 10$^+$};
		\node (c1) at (2,1) 	{$t_6-t_8$: 101 01$^-$};
		\node (d1) at (4,1) 	{...};
		
		\node (a2) at (-3,0) {010 01$^-$};
		\node (b2) at (-1.5,0) {100 01$^-$};
		\node (c2) at (0,0) {\textbf{110 00}$^+$};
		\node (d2) at (1.5,0) {001 10$^-$};
		\node (e2) at (3,0) {\textbf{100 10}$^+$};
		\node (f2) at (4.5,0) {101 00$^-$};
		
		\node (a3) at (-1,-1) {000 01$^-$};
		\node (b3) at (2,-1) {100 00$^-$};
		
		\draw (a3) -- (a2) -- (a1) -- (b2) -- (a3) ;
		\draw (b3) -- (b2) -- (c1);
		\draw (a1) -- (c2);
		\draw (b3) -- (b2);   
		\draw (e2) -- (b3) -- (f2) -- (b3) -- (c2);   
		\draw (d2) -- (b1) -- (e2);
		\draw (b1) -- (f2) -- (c1);   
	\end{tikzpicture}
	\caption{Partially ordered set created from the records in \cref{{tab:example}}.}
	\label{fig:poset}
\end{figure}
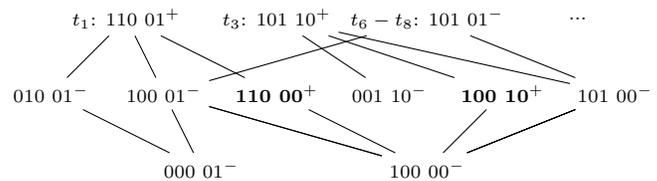

\Cref{fig:poset} shows the partially ordered set corresponding to \cref{tab:example}. 
At the beginning, the nodes at the top are the ones for which we know the label represented with a 
superscript symbol $+$ and $-$ for positive and negative, respectively. They can be seen as 
maximally-specific rules. If we take as target the positive class, we move inside the 
Boolean lattice by flipping-off positive bits, starting from the positive binary samples, and go down 
to find binary elements -- located on the boundary -- that divide positive and negative samples. 
While we navigate the Boolean lattice, nodes are labelled according to the cover test against the 
negative samples. As soon as a conflict is found, we can avoid going down from that node, but there 
is still the possibility to explore that path from another binary sample. This recursive procedure 
corresponds to up-and-down movements in the lattice. However, if at each iteration we are able to 
select the best candidate bit and to avoid conflicts, we only allow steps down in the Boolean lattice. 
We use the 
heuristic described in the main paper to choose the best candidate bit to flip-off.

\subsection{A Practical Example}
Consider again the example in \cref{tab:example}. Since at the beginning $\mathcal{S}=\mathcal{T}$, we 
will only report $|\mathcal{T}_i^0|$. For the first positive record $t_1=$110 01, we have: 
$\mathcal{F}_1^0 = \{01101, 01110\}$, $\mathcal{F}_2^0 = \{10101\}$, $\mathcal{F}_5^0 = \{01110\}$. 
We have therefore: $d_l(t_1, \mathcal{F}_1^0) = 1$, $d_l(t_1, \mathcal{F}_2^0) = 1$, $d_l(t_1, \mathcal{F}_5^0) = 2$. 
We already know that flipping-off either the first or the second bit to 0 would lead to a conflict: 
thus, we directly flip-off the fifth bit to obtain the boundary point 110 00, independently from the 
value of $|\mathcal{T}_5^0|$. Element 110 00 is added in the set of boundary points $\mathcal{A}$. 

For the second positive record $t_3=$ 101 10, we have: 
$\mathcal{F}_1^0 = \{01101, 01110\}$, $\mathcal{F}_3^0 = \emptyset$, $\mathcal{F}_5^0 = \{01110\}$. 
We have therefore: $d_l(t_1, \mathcal{F}_1^0) = 1$, $d_l(t_1, \mathcal{F}_3^0) = undefined$, and 
$d_l(t_1, \mathcal{F}_4^0) = 1$. Although $i=3$ induces a distance from an empty set, since we know 
that flipping-off other indexes generates conflicts, we can immediately label 100 10 as boundary point 
and add it to $\mathcal{A}.$

\subsection{From Boundary Set To Rules}
At the end of the previous phase, we obtain the boundary set $\mathcal{A} = \{11000, 10010\}$. In this 
case, each boundary point covers only one distinct positive sample, therefore the union of the two points 
covers all the set of positive samples and both points are kept after the regularization. Let's suppose 
to follow a positive set cover strategy, without early stopping condition. Then, the boundary set can be 
immediately mapped to the rule set shown in \cref{tab:ruleset_example}.

\begin{figure}[ht]
	\footnotesize
	\begin{mdframed}
        \textbf{IF} CPU $\in$ [95, max)\\
        \textbf{OR} CPU $\in$ [81, max) and MEM $\in$ [85, max)\\
		\textbf{THEN} Label = 1\\
		\textbf{ELSE} Label = 0
	\end{mdframed}
	\caption{Ruleset extracted from the boundary}
    \label{tab:ruleset_example}
\end{figure}

\section{PARALLEL AND DISTRIBUTED IMPLEMENTATION}
\label{sec:parallel_and_distributed}
\LIBRE is amenable to parallel and distributed implementations. Indeed, it processes one positive 
sample at a time. An exhaustive version of the \texttt{FindBoundaryPoint()} procedure is 
embarrassingly parallel and it is easily parallelizable on multi-core architectures: 
it is sufficient to spawn a UNIX process per positive sample, and exploit all available cores. 

Instead, the approximate procedure, requires a slightly more involved approach. Indeed, the approximate 
\texttt{FindBoundaryPoint(.)} procedure processes positive records that have not yet been covered by 
any boundary point. Hence, a global view on the set $\mathcal{S}$ is required. We experimented with two alternatives. 
The first is to place $\mathcal{S}$ in a shared, in-RAM datastore, because UNIX processes -- unlike threads -- 
do not have shared memory access. The second alternative is to simply let each individual process to 
hold their own version of $\mathcal{S}$, thus sacrificing a global view.
Our experiments indicate that the loss in performance due to a local view only is negligible, 
and largely out-weighted by the gain in performance, since the execution time decreases linearly 
with the number of spawned UNIX processes. Moreover, both $\mathcal{D}_+$ and $\mathcal{D}_-$ remain 
consistent throughout the whole induction phase. 

\LIBRE can be easily distributed such that it can run on a cluster of machines, using for example a 
distributed computing framework such as Apache Spark \cite{spark}. This approach, called 
\emph{data parallelism}, splits input data across machines, and let each machine execute, independently, 
a weak learner. The data splitting operation shuffles random subsets of the input features to each 
worker machine. Once each worker finishes to generate the local rule sets, they are merged in the ``driver'' 
machine, which eventually applies the filtering and then executes the rule selection procedure to 
produce the final boundary.

\section{THE IMPACT OF LIBRE'S PARAMETERS}
\label{sec:performance-interpretability_tradeoff}
In this section we investigate how acting on \LIBRE's parameters allows to obtain specific 
performance-interpretability tradeoffs. We will not cover all the possible parameters: in particular, 
we focus on the discretization threshold, $\#estimators$, and $\#features$ per estimator. The effects of 
$\alpha$ and early-stopping in weighted set cover are not reported here since their effects are well known 
from previous studies.
 
When we vary one parameter, all the others are kept fix to isolate its impact. We will also 
give some rules of thumb to choose them.

\subsection{The Effects Of Varying The Discretization Threshold}

The choice of the discretization threshold depends on the specific dataset: a threshold equal to zero 
means no discretization, whereas increasing the threshold is equivalent to increase the tolerance to 
combine consecutive ranges of values with different label distributions. 

In general, a zero threshold gives bad performances and results in a bigger lattice with a consequent 
slower training time; also a too aggressive (high) threshold is not recommended because it would lead to 
a huge loss of information.

The most significant effects occur as soon as we start increasing the threshold: in general, F1-score 
improves (and eventually oscillates) up to a value after which it can eventually decrease. It is clear 
that, if the dataset contains only continuous features and we continue to increase the threshold, original 
and discrete records will coincide at a certain point.

The threshold affects also the number of rules and their size. In general, when there is no 
discretization, two extreme cases are possible: i) We might have as many rules as the number of positive 
examples (if their binary representation does not generate conflicts with the elements in $\mathcal{F}$) 
with $\#atoms = \#features$. It means that the model simply overfitted the training data. ii) We might 
end up with few rules with very high number of atoms (or no rules at all): 	the model tried to generalize 
positive records but it was not able to learn something meaningful because too many conflicts were present 
in the dataset.

From our experiments, the second option is more common (few complex rules). Again, as soon 
as we start increasing the threshold, the model starts to learn: the number of discovered 
rules increases and the number of atoms decreases, since the model is able to filter out useless 
features. After that, changes tend to stabilize: in our experiments, this happens when the 
discretization threshold is roughly between 3 and 6.

%

\subsection{The Effects of Varying $\#estimators$ And $\#features$}

We analyze how $\#estimators$ and $\#features$ affect the predictive performance and interpretability 
of \LIBRE, by keeping fixed the remaining parameters. Results are reported for the \Heart UCI dataset, 
but the considerations we do are quite general.

\noindent \textbf{Parameter Settings.} We fixed a discretization threshold = 6. The search procedure 
optimizes the $H1$ heuristic, without applying any filtering before running weighted set cover, for which 
we set $\alpha=0.7$, without applying any early-stopping condition. We varied $\#estimators\in \{1,5,10,15,20\}$ 
and $\#features \in \{1,2,3,4,5,6,7,8\}$. We performed up to 50 runs for each ($\#estimators, \#features$), 
where features used by each estimator are randomly selected. Please, notice that this is not the 
optimal set of parameters. 

\begin{figure*}[p]
    \centering
    \begin{subfigure}[b]{0.25\textwidth}
        \includegraphics[width=\textwidth]{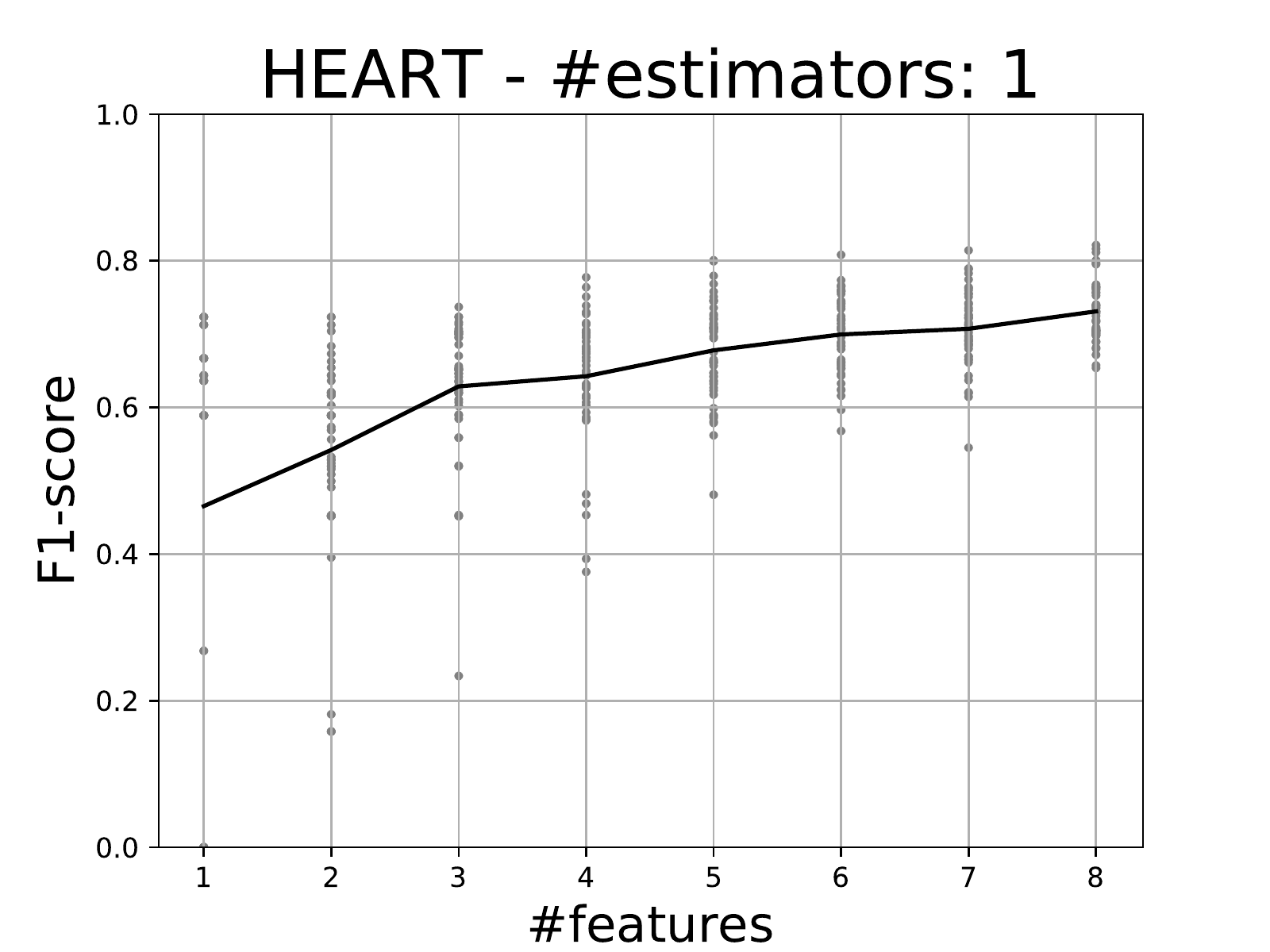}
        \label{fig:F1_n_e1}
    \end{subfigure}
    \begin{subfigure}[b]{0.25\textwidth}
        \includegraphics[width=\textwidth]{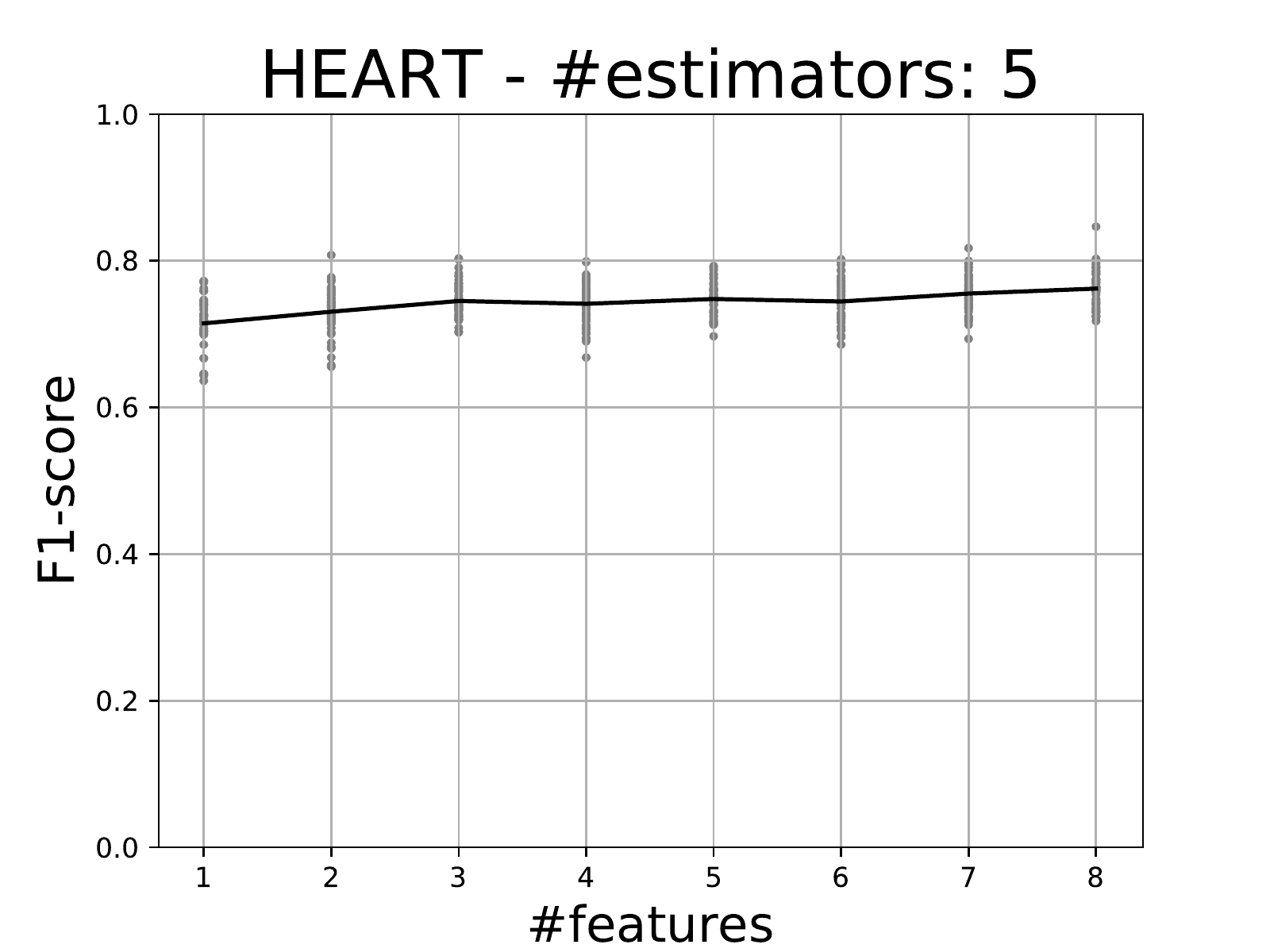}
        \label{fig:F1_n_e5}
    \end{subfigure}
    \begin{subfigure}[b]{0.25\textwidth}
        \includegraphics[width=\textwidth]{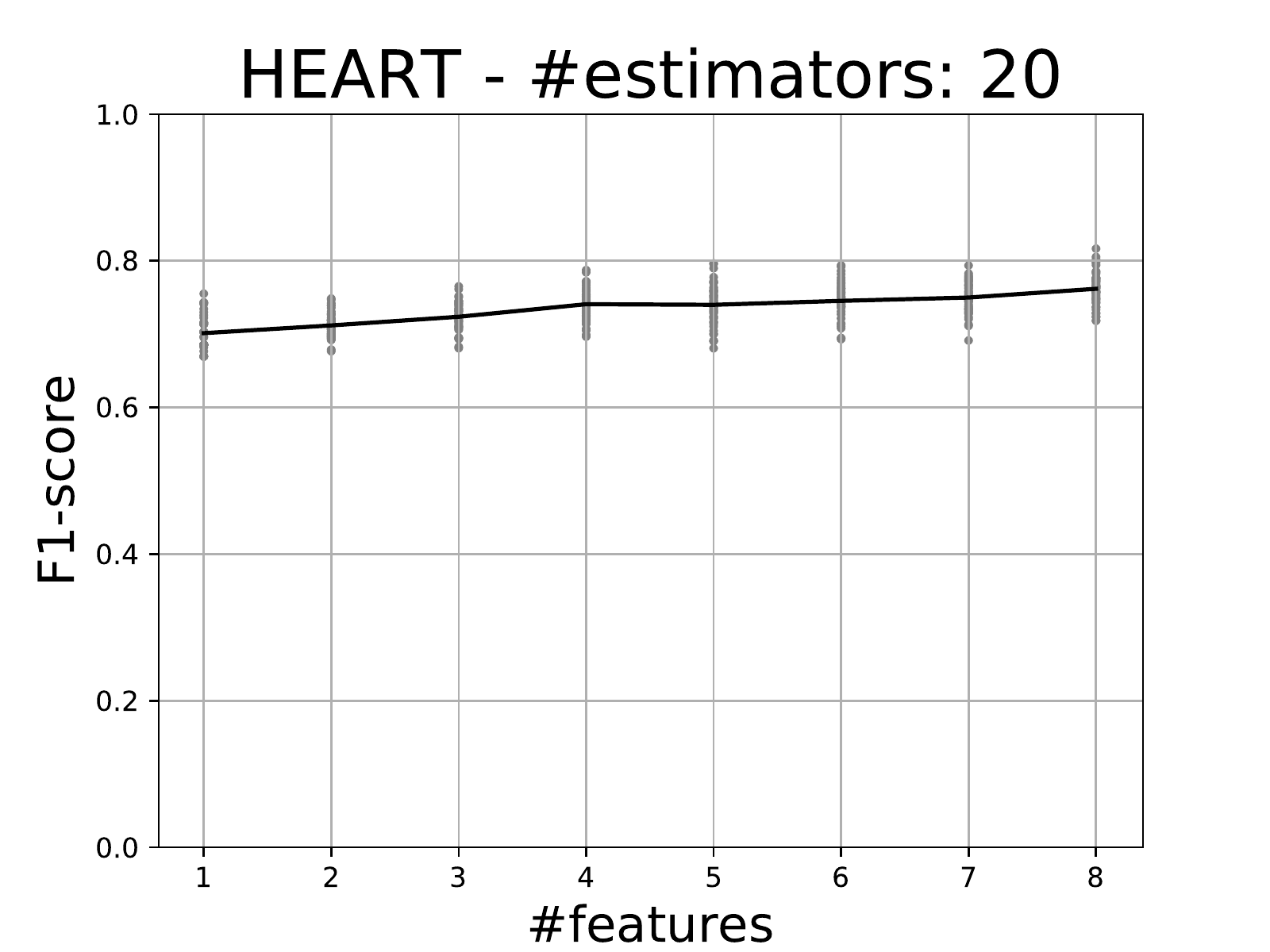}
        \label{fig:F1_n_e20}
    \end{subfigure}
	\vspace*{-7mm}
	\caption{\Heart dataset: F1-score as a function of \#features for two different values of \#estimators.}
	\label{fig:F1_n_e}
	\centering
    \begin{subfigure}[b]{0.25\textwidth}
        \includegraphics[width=\textwidth]{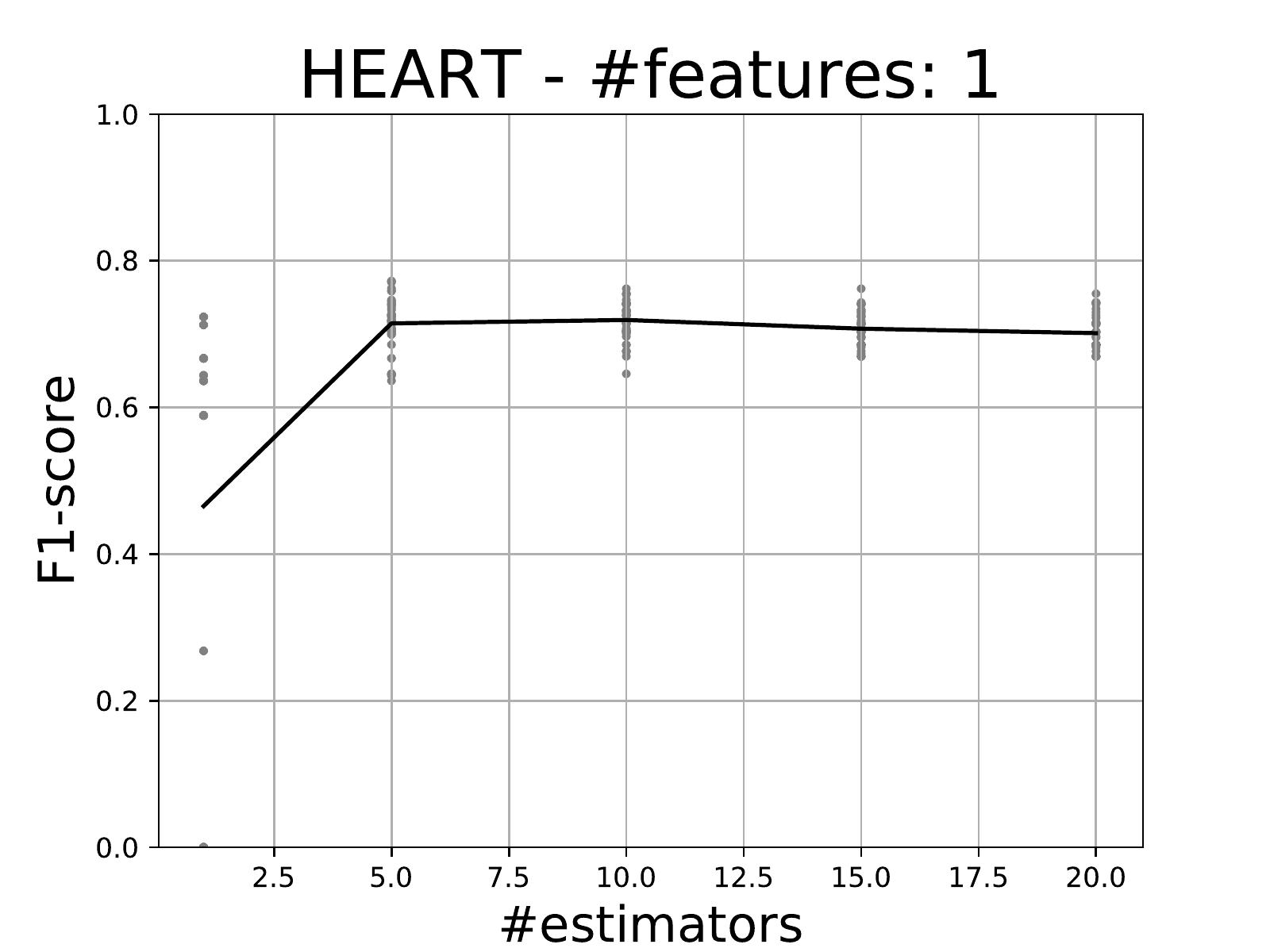}
        \label{fig:F1_n_f1}
    \end{subfigure}
    \begin{subfigure}[b]{0.25\textwidth}
        \includegraphics[width=\textwidth]{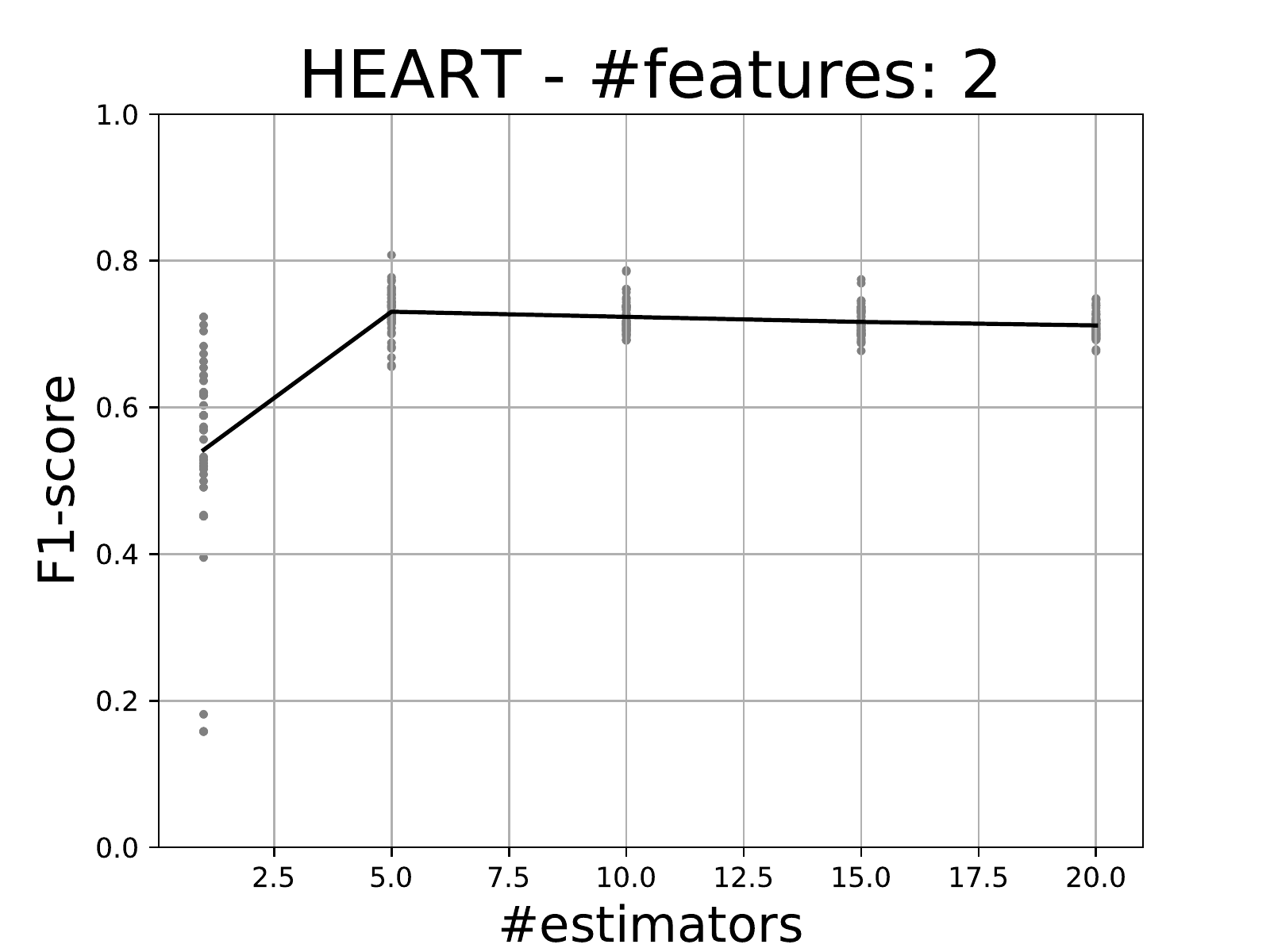}
        \label{fig:F1_n_f2}
    \end{subfigure}
    \begin{subfigure}[b]{0.25\textwidth}
        \includegraphics[width=\textwidth]{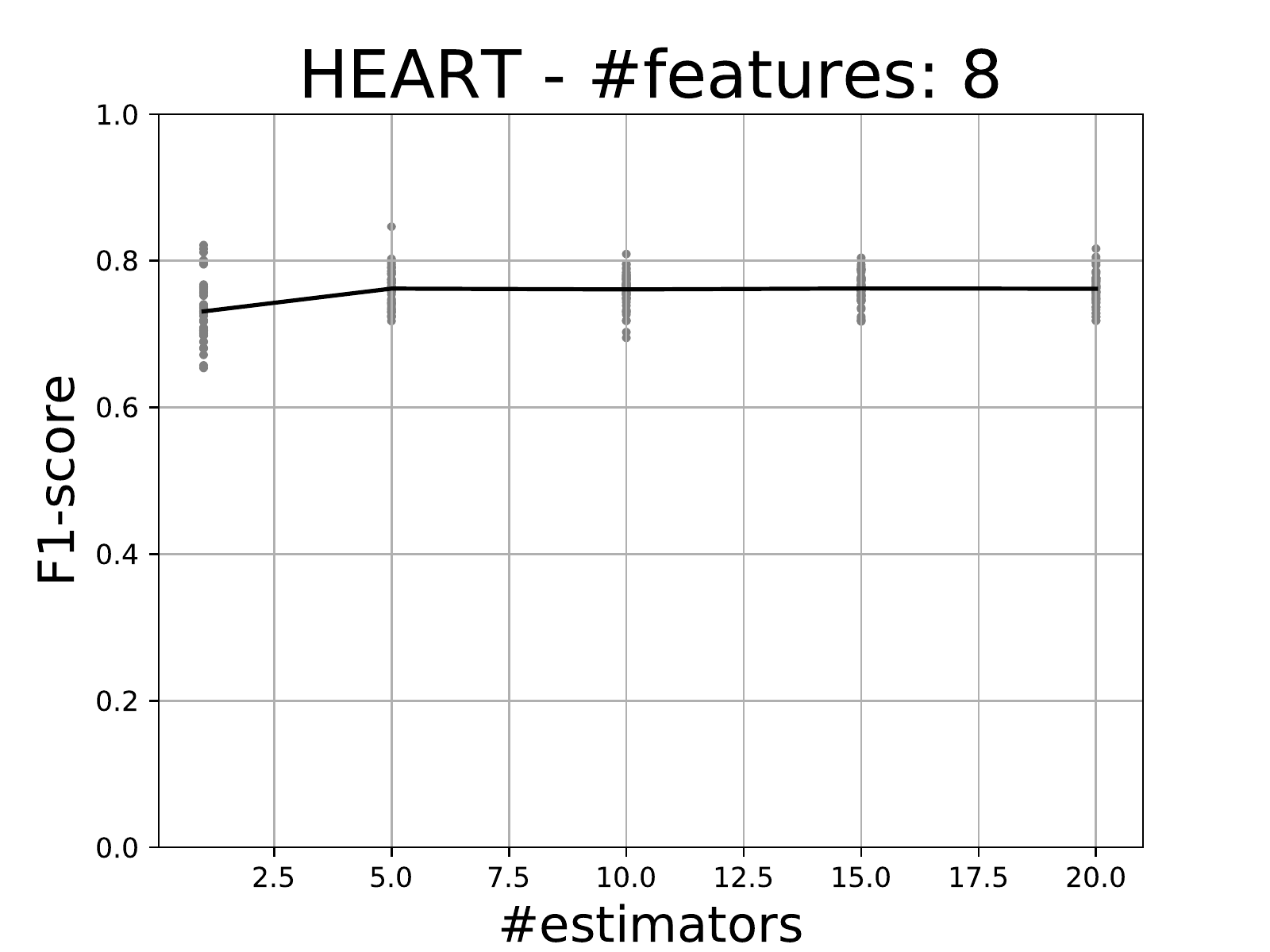}
        \label{fig:F1_n_f8}
    \end{subfigure}
	\vspace*{-7mm}
	\caption{\Heart dataset: F1-score as a function of \#estimators for four different values of \#features.}
    \label{fig:F1_n_f}
    \begin{subfigure}[b]{0.25\textwidth}
        \includegraphics[width=\textwidth]{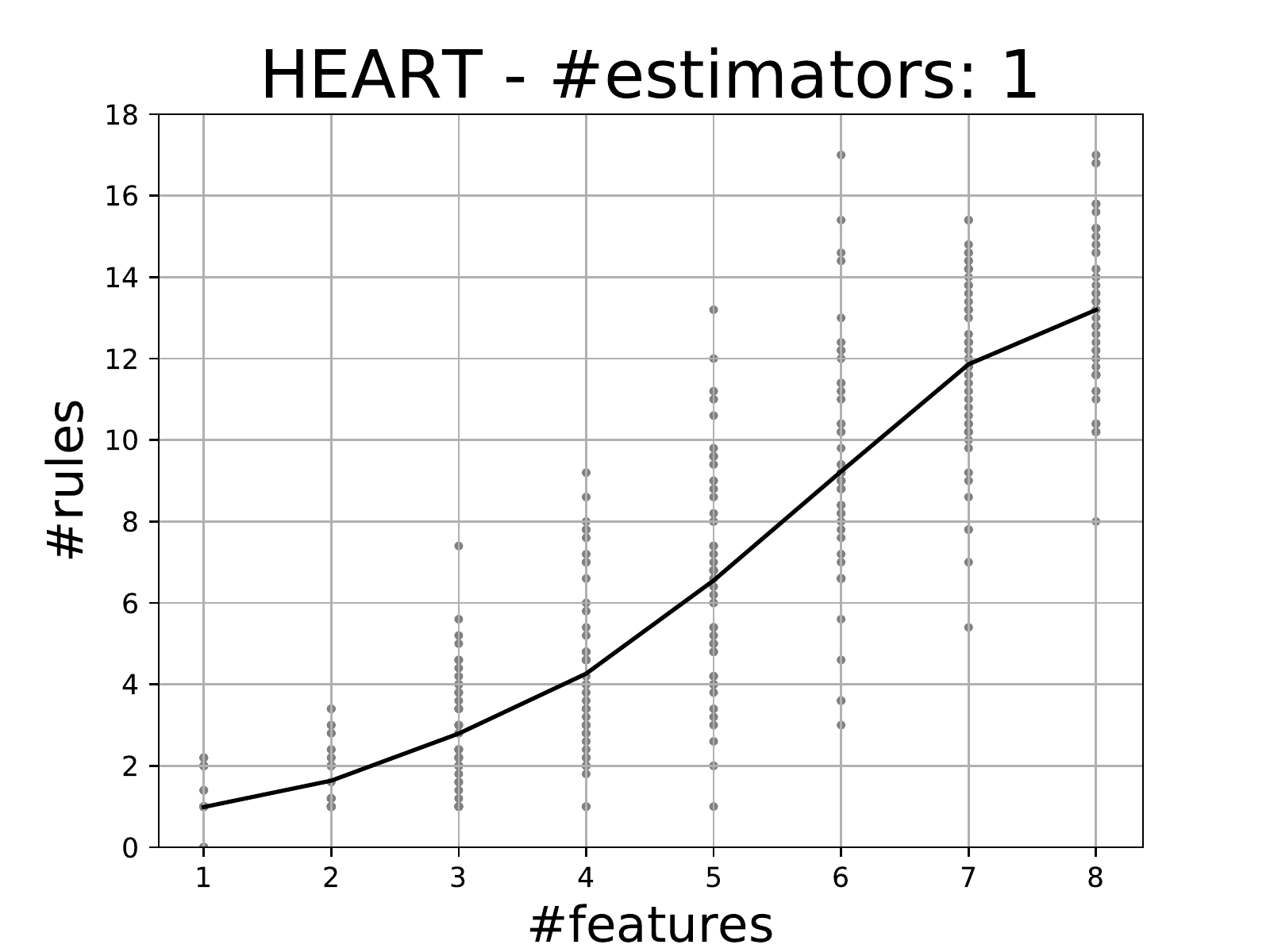}
        \label{fig:rules_n_e1}
    \end{subfigure}
    \begin{subfigure}[b]{0.25\textwidth}
        \includegraphics[width=\textwidth]{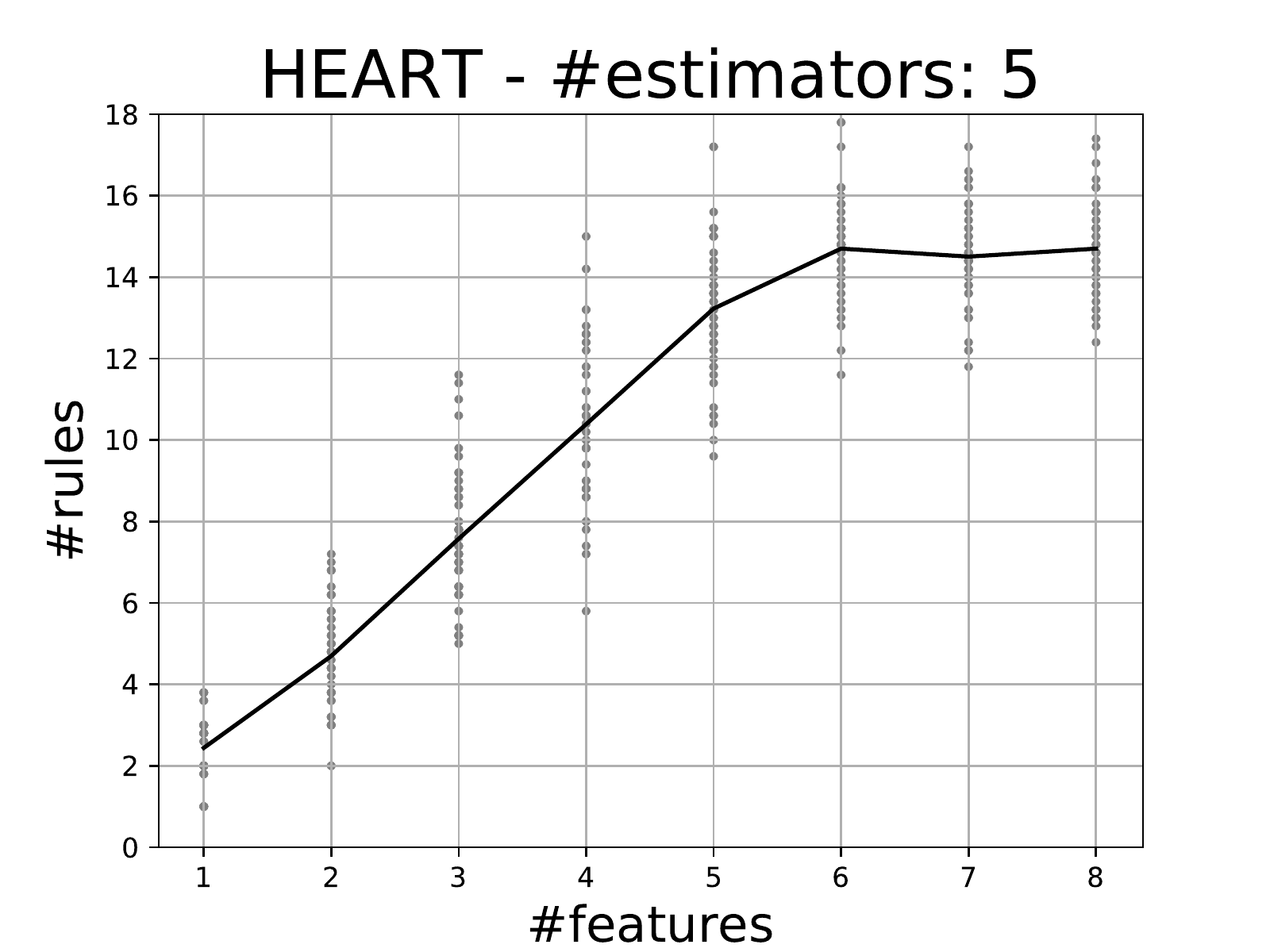}
        \label{fig:rules_n_e5}
    \end{subfigure}
    \begin{subfigure}[b]{0.25\textwidth}
        \includegraphics[width=\textwidth]{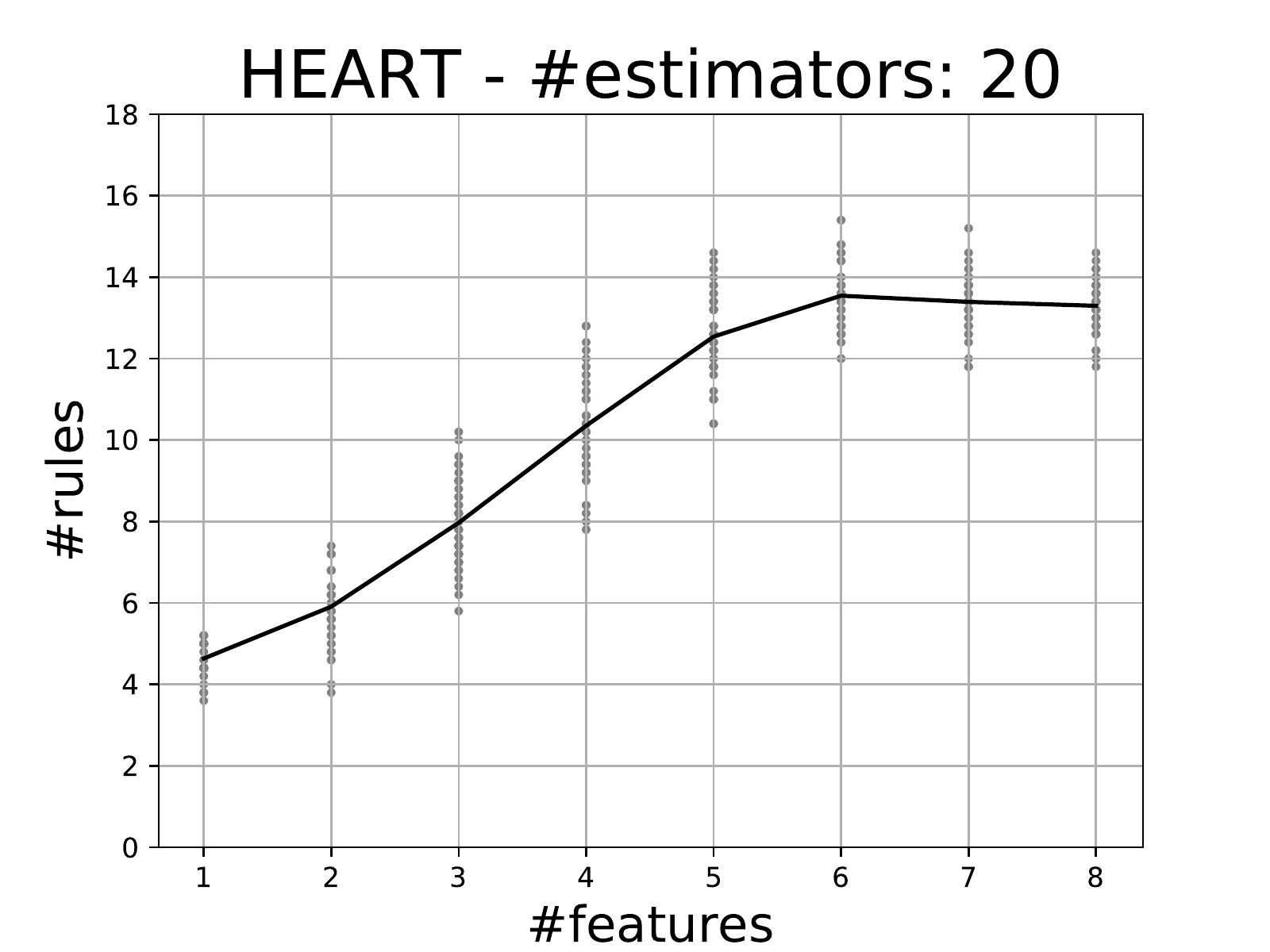}
        \label{fig:rules_n_e20}
    \end{subfigure}
    \vspace*{-7mm}
	\caption{\Heart dataset: \#rules as a function of \#features for four different values of \#estimators.}
    \label{fig:rules_n_e}
    \begin{subfigure}[b]{0.25\textwidth}
        \includegraphics[width=\textwidth]{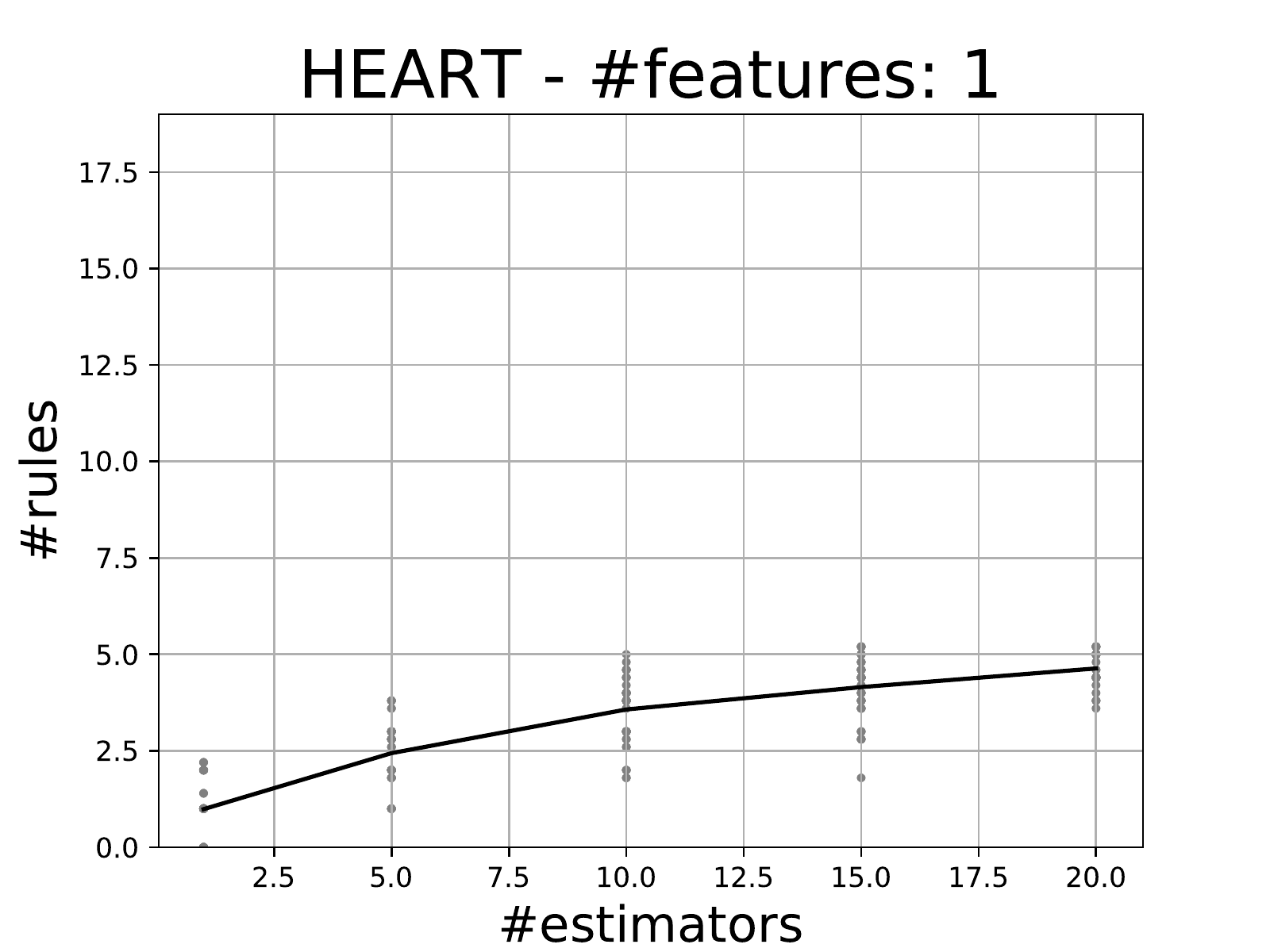}
        \label{fig:rules_n_f1}
    \end{subfigure}
    \begin{subfigure}[b]{0.25\textwidth}
        \includegraphics[width=\textwidth]{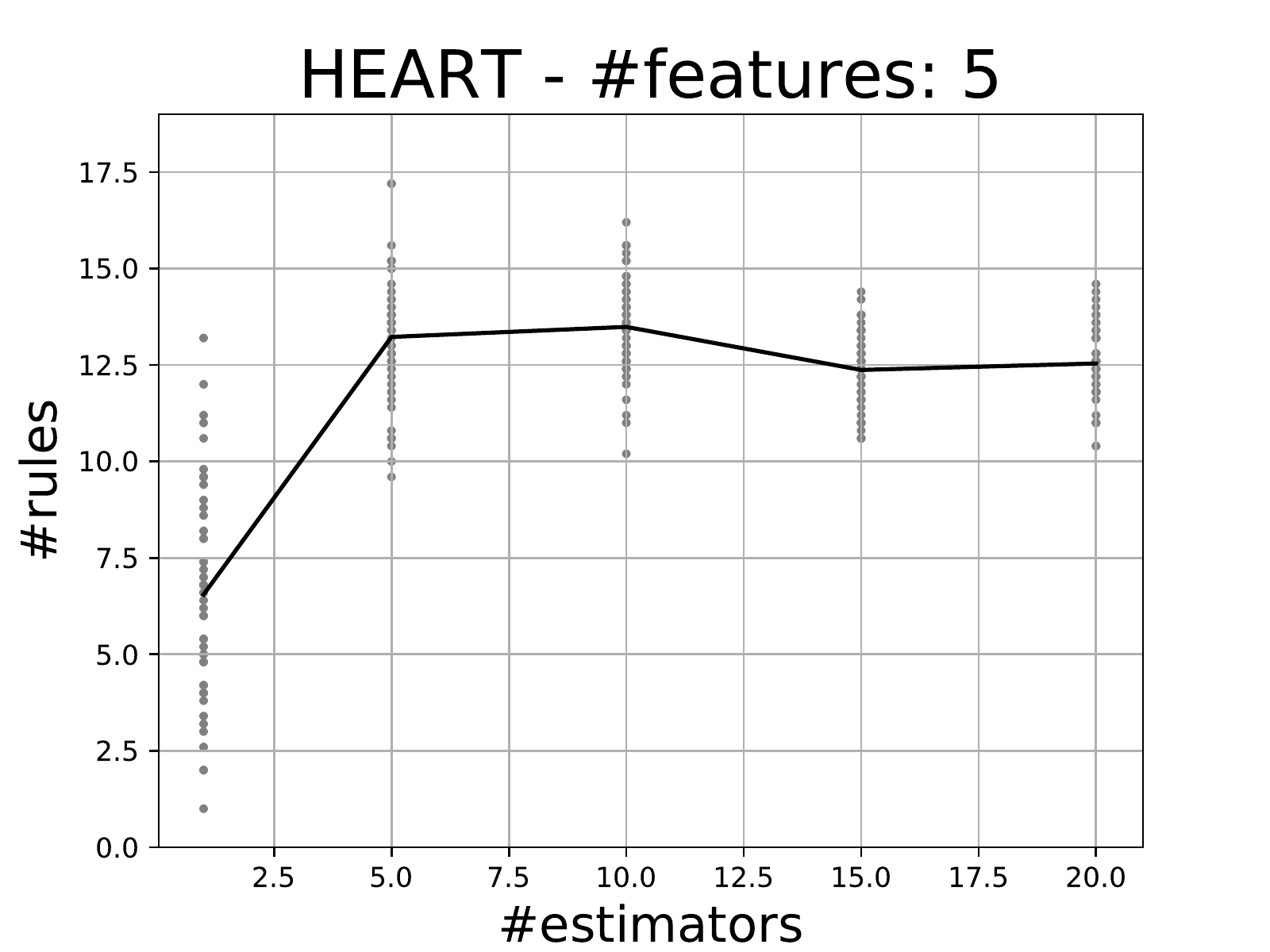}
        \label{fig:rules_n_f5}
    \end{subfigure}
    \begin{subfigure}[b]{0.25\textwidth}
        \includegraphics[width=\textwidth]{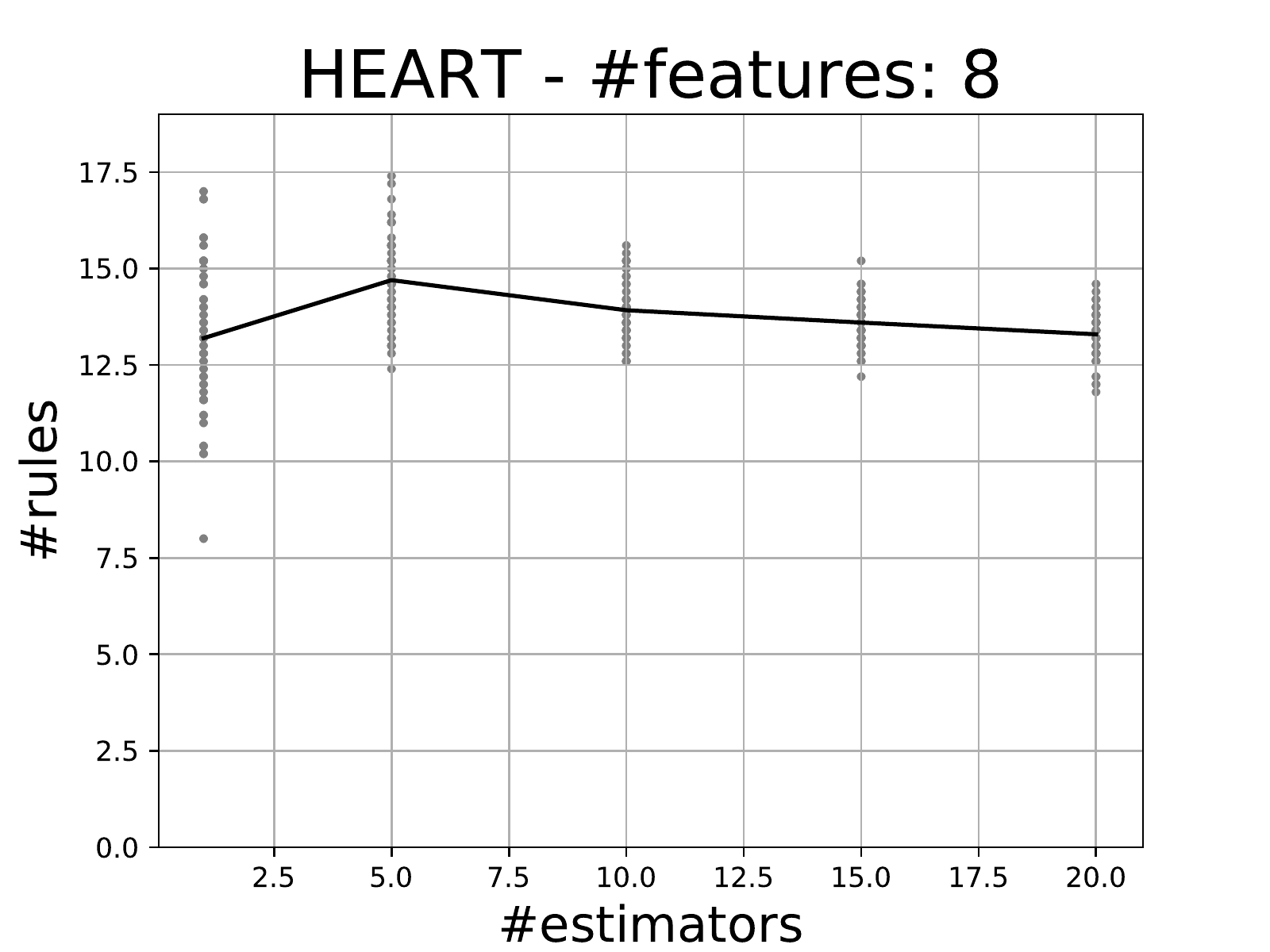}
        \label{fig:rules_n_f8}
    \end{subfigure}
    \vspace*{-7mm}
	\caption{\Heart dataset: \#rules as a function of \#estimators for four different values of \#features.}
    \label{fig:rules_n_f}
    \begin{subfigure}[b]{0.25\textwidth}
        \includegraphics[width=\textwidth]{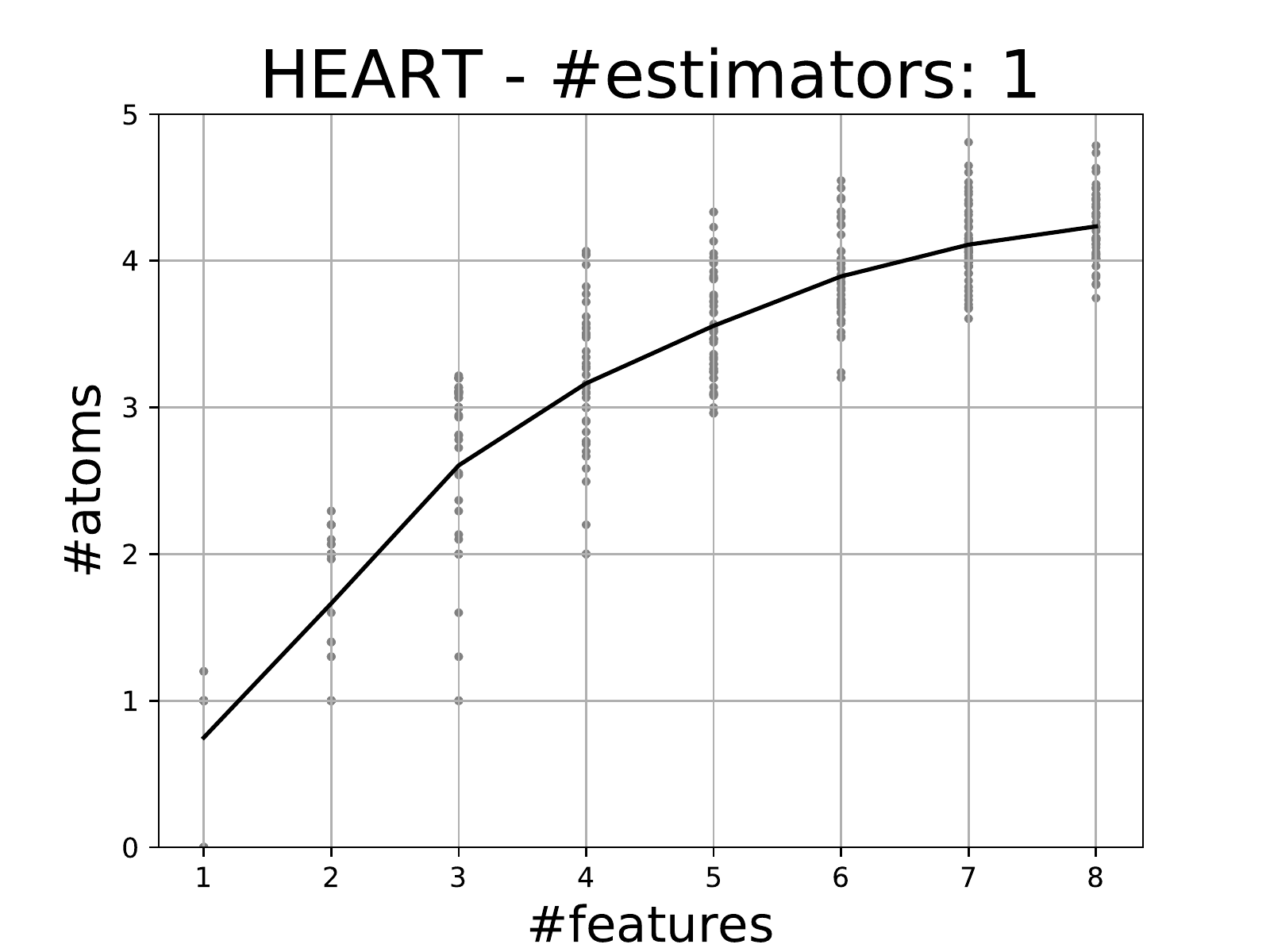}
        \label{fig:n_atoms_n_e1}
    \end{subfigure}
    \begin{subfigure}[b]{0.25\textwidth}
        \includegraphics[width=\textwidth]{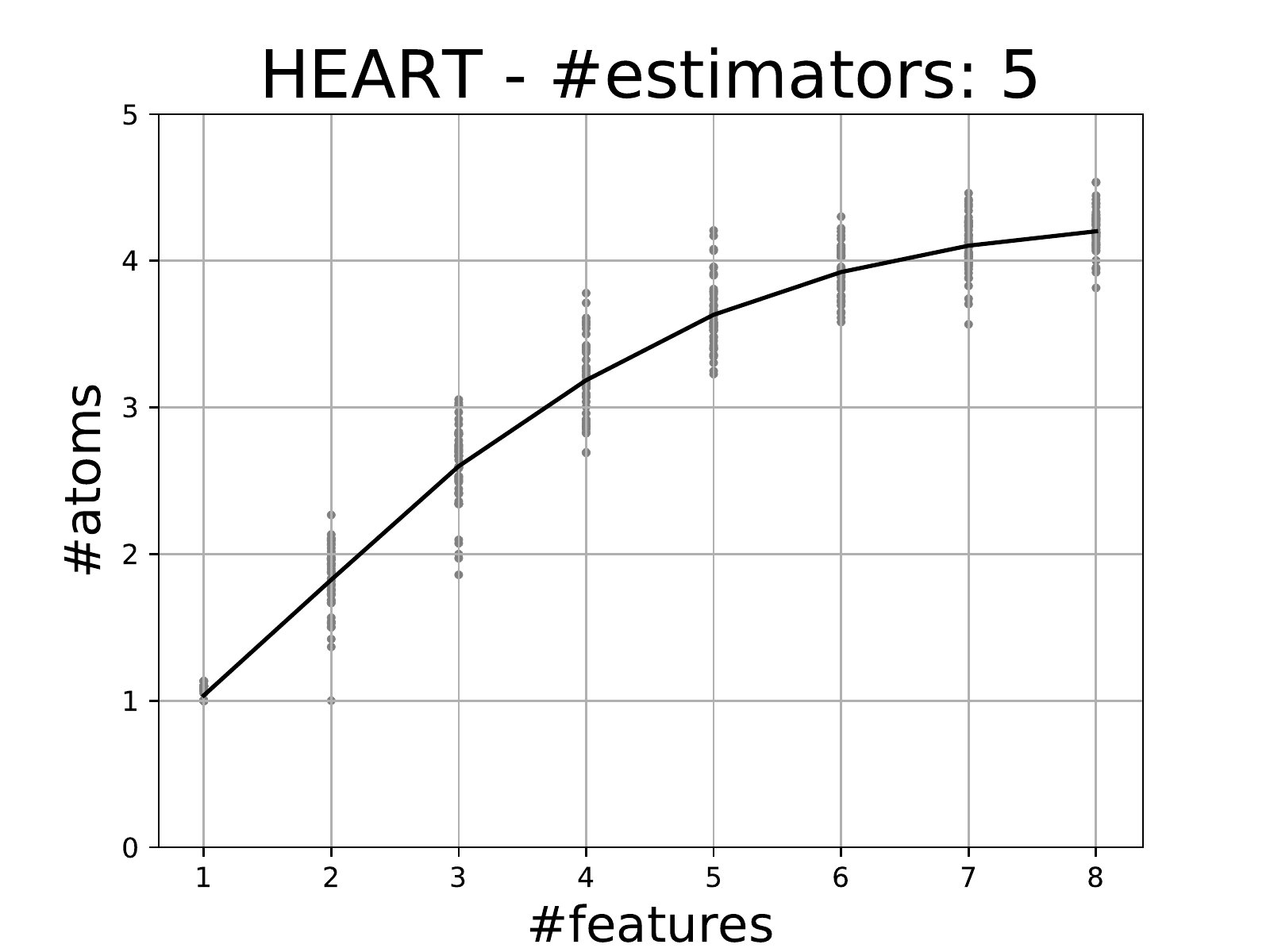}
        \label{fig:n_atoms_n_e5}
    \end{subfigure}
    \begin{subfigure}[b]{0.25\textwidth}
        \includegraphics[width=\textwidth]{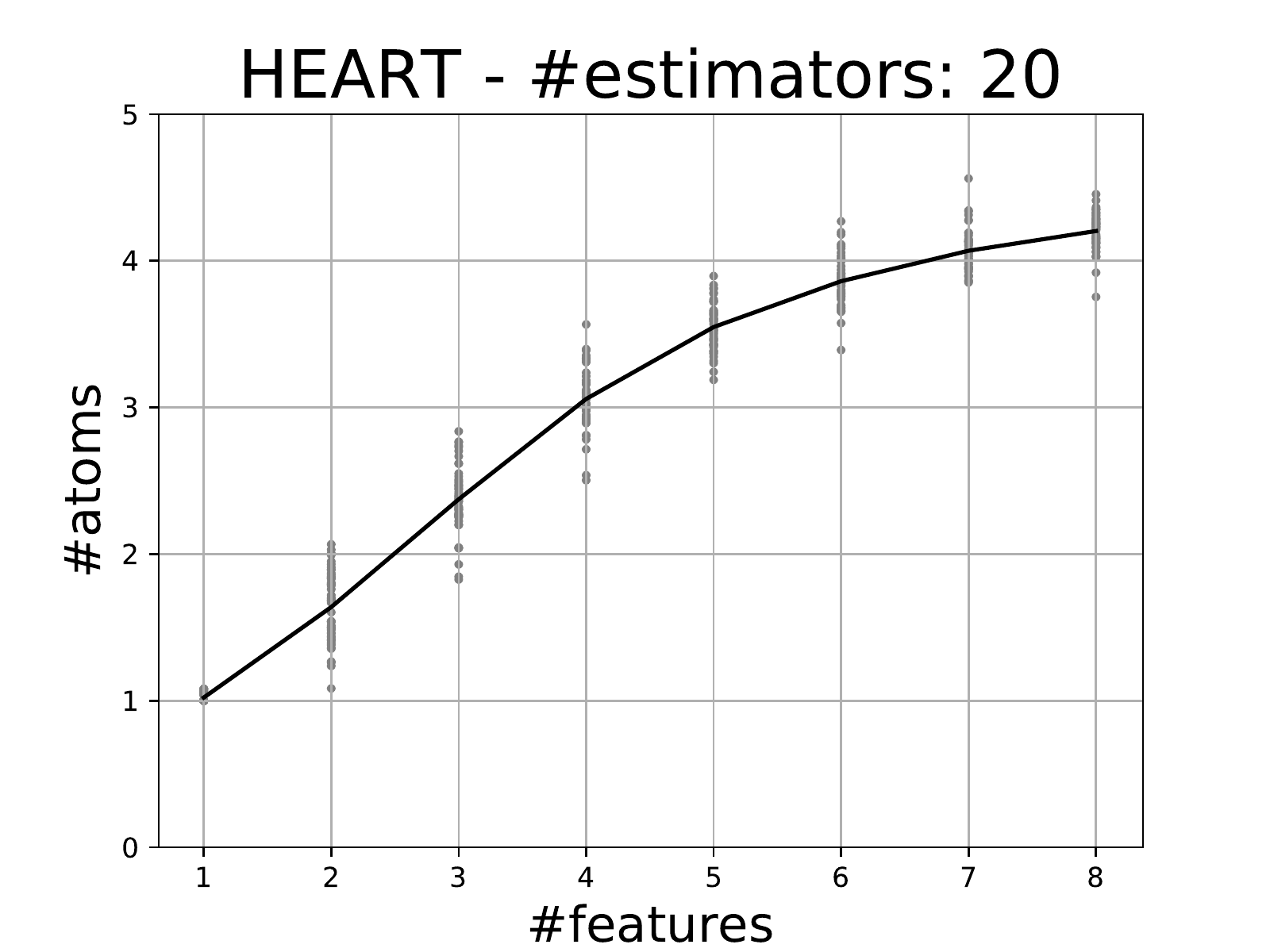}
        \label{fig:n_atoms_n_e20}
    \end{subfigure}
    \vspace*{-7mm}
	\caption{\Heart dataset: \#atoms as a function of \#features for four different values of \#estimators.}
    \label{fig:atoms_n_e}
    \begin{subfigure}[b]{0.25\textwidth}
        \includegraphics[width=\textwidth]{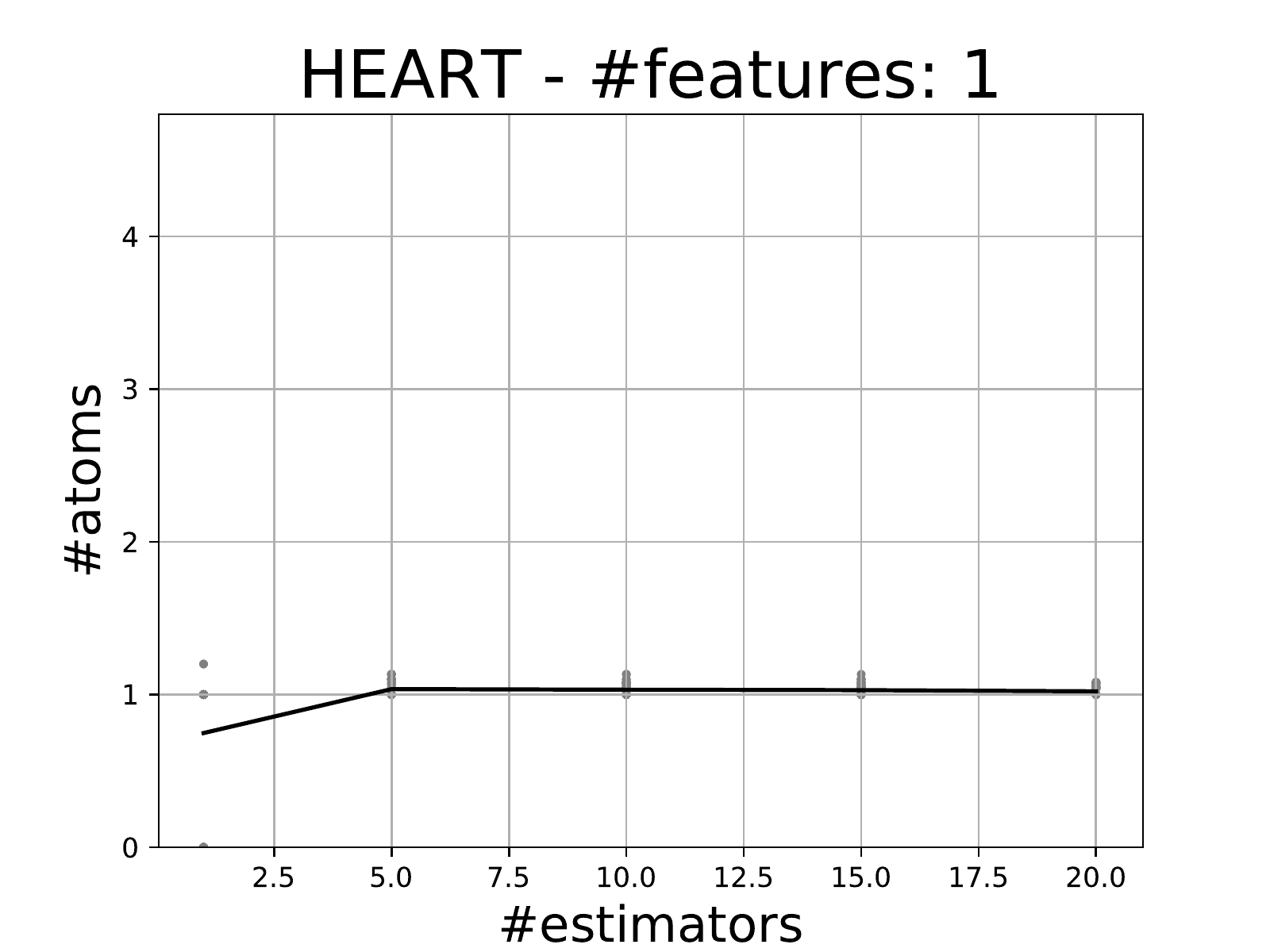}
        \label{fig:n_atoms_n_f1}
    \end{subfigure}
    \begin{subfigure}[b]{0.25\textwidth}
        \includegraphics[width=\textwidth]{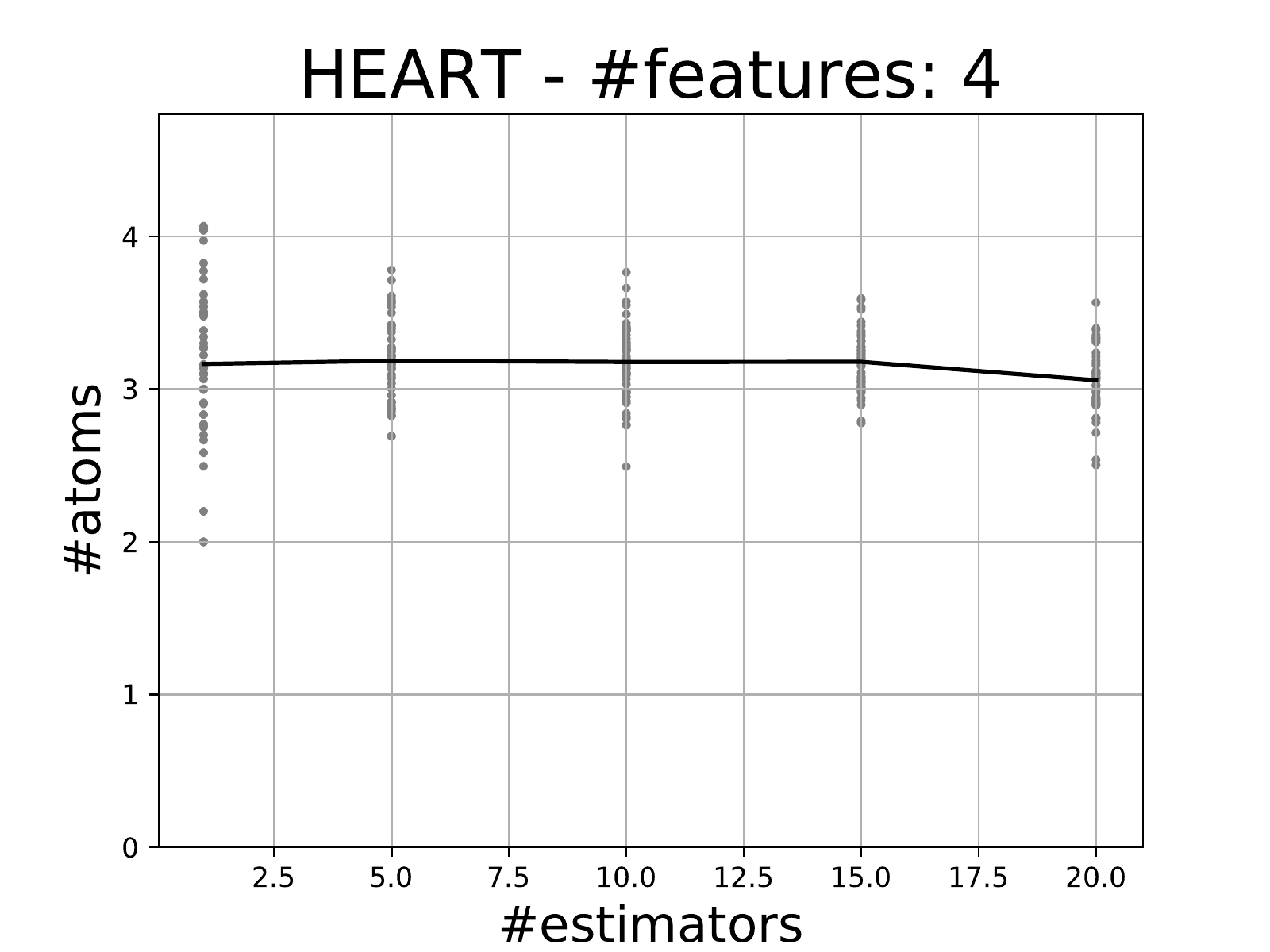}
        \label{fig:n_atoms_n_f4}
    \end{subfigure}
    \begin{subfigure}[b]{0.25\textwidth}
        \includegraphics[width=\textwidth]{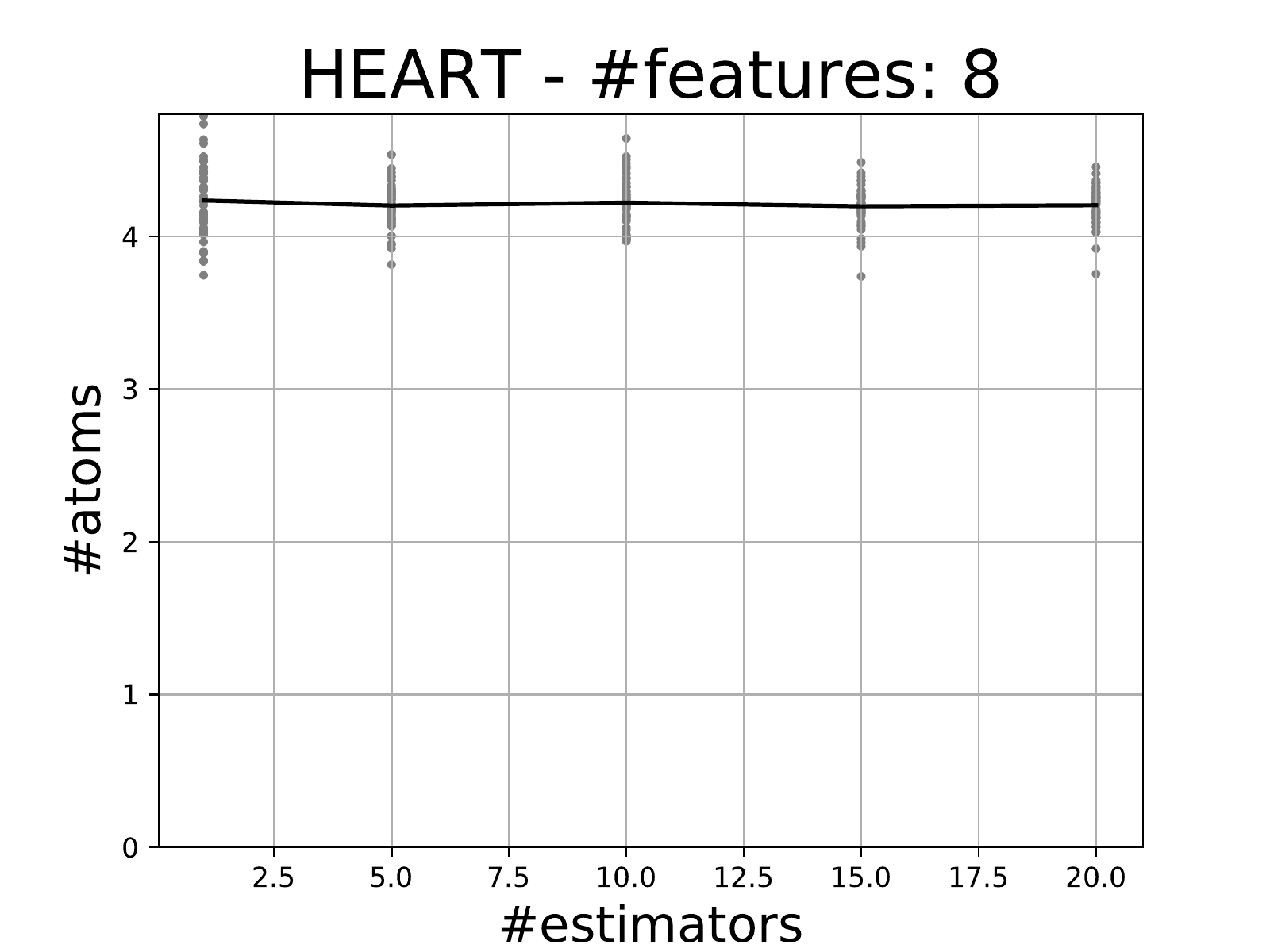}
        \label{fig:n_atoms_n_f8}
    \end{subfigure}
    \vspace*{-7mm}
	\caption{\Heart dataset: \#atoms as a function of \#estimators for four different values of \#features.}
    \label{fig:atoms_n_f}
\end{figure*}

\noindent \textbf{Effects On F1-score.} As shown in \cref{fig:F1_n_e}, if we fix $\#estimators$, when 
$\#estimators$ is low (one estimator), F1-score improves considerably as long as $\#features$ increases. 
When enough $\#estimators$ are used, F1-score stabilizes: we can use less $\#features$ per estimator with 
almost no effect on F1-score.

From \cref{fig:F1_n_f}, we can see that, if we fix $\#features$, F1-score benefits from 
increasing $\#estimators$. When $\#features$ increases, limiting $\#estimators$ to a low value 
does not significantly impact the F1-score.

In other words, for low $\#features$ it is convenient to run more $\#estimators$: each estimator would 
work on different subsets of the input features and the union of rules would be hopefully diverse, 
with a consequent higher F1-score.
For the specific case of \Heart, we do not notice any significant difference in F1-score by passing 
from 5 to 20 estimators. However, it is generally convenient to increase $\#estimators$ in order to 
try as many combinations of features as possible and reduce the variance of results. For datasets 
with many features, this may make the difference.

\noindent \textbf{Effects On $\#rules$.} As shown in \cref{fig:rules_n_e}, if we fix $\#estimators$ 
and increase $\#features$, the average number of rules tends to increase up to a certain value, and 
then stabilizes or get slightly worse.

From \cref{fig:rules_n_f}, we notice that, when $\#features$ is low, the number of rules 
tends to increase as long as we increase the number of estimators. Indeed, the model generates less 
rules when there are not enough discriminant features; increasing the number of estimators, each 
estimator discovers different rules that are combined.
As long as we increase $\#features$ per estimator, the probability that different 
estimators work with similar sets of features increases, together with the probability of 
generating the same rules (or very similar rules): that's why the size of the rule set tends 
to stabilize. In this cases, it might be convenient to run less estimators to save execution time. 

In general, increasing the number of estimators considerably reduces the variance of results.

\noindent \textbf{Effects On $\#atoms$.} As shown in \cref{fig:atoms_n_e}, if we keep $\#estimators$ fixed, 
$\#atoms$ of the rule set increases as long as the number $\#features$ increases.
If we fix $\#features$ (\cref{fig:atoms_n_f}), $\#estimators$ does not seem to affect $\#atoms$ significantly. 

As usual, increasing $\#estimators$ reduces the variance of the results.

\noindent \textbf{Final Remarks.} In conclusion, if we want interpretable rule sets, it is better to 
use few input features per estimator and as many estimators as possible. 

In \cref{sec:performance-interpretability_tradeoff}, we have not used any early stop condition. However, 
it is a good practice to tune this parameter in order to generate rule sets that are more interpretable 
and highly accurate.

\begin{figure*}[ht]
	\centering
    \includegraphics[width=\linewidth]{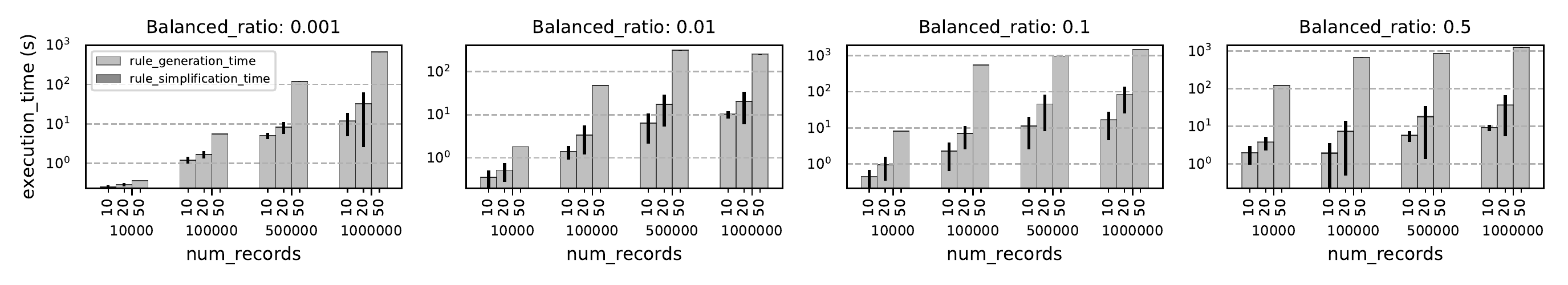}
    \caption{Run time on synthetic data.}
    \label{fig:scalability_plot}
\end{figure*}

\section{SCALABILITY EVALUATION}
\label{sec:scalability}
Here, we extensively test the scalability of \LIBRE. Unlike the main paper, we 
use up to 50 features and investigate also the impact of class imbalance on the execution time.

\noindent \textbf{Synthetic Dataset.} For the scalability evaluation, we synthetically generate 
a dataset with $1'000'000$ records and $50$ continuous features with randomly generated values 
in the domain $[0,100]$. Then, we randomly generate four sets of binary labels with a class 
imbalance ratio of $0.001$, $0.01$, $0.1$, and $0.5$ respectively.

\vspace{2mm}
\noindent \textbf{Settings.} We vary the number of records (10'000, 100'000, 500'000, 1'000'000), features
(10, 20, 50), and class imbalance ratio (0.001, 0.01, 0.1, 0.5): for each dataset configuration, 
\LIBRE runs up to 100 times with different randomly generated subsets of features of size 10, 20, 
and 50; the average execution time in seconds is reported as a sum of two contributions: rule 
generation and simplification times. Times refer to one weak learner only: 
if $N$ weak learners run in parallel, the reported time is still a good estimate. Before executing \LIBRE, we discretize the 
dataset with a discretization threshold equal to 6, that we empirically find out to be a good 
value. The simplification procedure runs on the top 500 rules, if more are generated.

\vspace{2mm}
\noindent \textbf{Results.} As shown in \cref{fig:scalability_plot}, the execution time is dominated by the rule 
generation term. Given a class imbalance ratio, execution time increases as long as we increase 
the number of records and features. The generation time also depends on which features are fed 
into the model for two main reasons: i) ChiMerge encodes bad predictive features with 
bigger domains, increasing the search space; ii) the generation procedure will struggle more to 
generate rules when it runs on features that are not that useful to predict the target class. 
This explains the high variance in the results.

Intuitively, as long as the class imbalance ratio gets close to 0.5, the number of processed records 
increases, together with the execution time. However, we verified experimentally that
this effect is somehow compensated by the higher number of negative records.

As already pointed out in the main paper, we run the rule generation procedure up to 50 
features just for experimental purposes: for practical applications, if interpretability is a need, 
it is more convenient to limit the number of features and train a bigger ensemble with more learners 
in order to generate compact rules in a reasonable time.

\section{FULL EXPERIMENTS}
\label{sec:pred_and_interpr}
In this section we report the full experimental campaign. We use the same methods, training 
procedure, preprocessing, and evaluation measures as the main paper, but we report the results 
for more datasets, as described in \cref{tab:datasets}. We also clarify which class we have 
trained the model on (target class). In case of multi-class classification datasets, records 
not belonging to the target class are considered to be negative.

\Cref{tab:f1-score} reports a comparison between \LIBRE and the selected methods in terms of 
F1-score, whereas \cref{tab:rules} and \cref{tab:atoms} reports the number of rules and average 
number of atoms, respectively. We also compare the rule sets leading to the best F1-scores for 
\RIPPERK, \BRS, and \SBRL with a few configurations for \LIBRE. In \cref{fig:interpretable_rulesets}, 
we report the average number of rules and atoms per rule, as a function of the F1-score: points at 
the bottom-rigth side of each plot are preferable since they correspond to compact and high predictive 
rule sets.

\begin{table*}[p]
    \scriptsize
    \begin{center}
		\begin{tabular}{l|c|c|c|c}
			\toprule
			\textbf{Dataset} 	& \textbf{\#records} 	& \textbf{\#features}	& \textbf{imbalance\_ratio}	& \textbf{target\_class}	\\
			\midrule
			\Adult 				& 48'842				& 14 					& .23 						& $>$50k					\\
			\Australian			& 690					& 14					& .44						& 2							\\
			\Balance			& 625					& 4						& .08						& B							\\
			\Bank 				& 45'211 				& 17 					& .12 						& yes						\\
			\Haberman 			& 306 					& 3 					& .26 						& died						\\
			\Heart				& 270					& 13					& .51						& presence					\\
			\Ilpd 				& 583 					& 10 					& .28 						& liver patient				\\
			\Liver 				& 345 					& 5 					& .51 						& drinks$>$2				\\
			\Pima				& 768					& 8						& .35						& 1							\\
			\Sonar 				& 208 					& 60 					& .53 						& R							\\
			\Tictactoe			& 958					& 9						& .65						& positive					\\
			\Transfusion 		& 748 					& 5						& .24						& yes						\\
			\Wisconsin			& 699					& 9						& .34						& malignant					\\
			\Sapclean			& 287'031				& 45					& .01						& crash\\
			\Sapfull			& 1'554'227				& 45					& .01						& crash\\
			\bottomrule
		\end{tabular}
    \end{center}
	\caption{Characteristics of evaluated datasets.}
    \label{tab:datasets}
\end{table*}

\begin{table*}[p]
    \scriptsize
    \begin{center}
		\begin{tabular}{l|c|c|c|c|c|c|c|c|c}
			\toprule
			Dataset & 		\RBFSVM				& \RF 				& \DT 				& \RIPPERK 			& \MODLEM			& \SBRL 			& \BRS 				& \underline{\LIBRE} 	& \underline{\LIBRE 3}	\\
			\midrule		
			\Adult          & .62(.01)			& .68(.01) 			& .68(.01)			& .59(.02) 			& .66(.01)			& .68(.01) 			& .61(.01)			& \textbf{.70(.01)} 	& .62(.01)				\\
			\Australian     & .83(.02)			& \textbf{.86(.02)}	& .84(.02)			& .85(.02) 			& .68(.28)			& .82(.03) 			& .83(.03)			& .84(.03) 				& .84(.03)				\\
			\Balance		& .03(.07)			& .00(.00) 			& .01(.03)			& .00(.00) 			& \textbf{.16(.04)}	& .00(.00) 			& .00(.00)			& \textbf{.16(.08)}		& .14(.06)				\\
			\Bank           & .46(.01)			& .50(.01) 			& .50(.01)			& .44(.04) 			& .50(.03)			& .50(.02) 			& .32(.05)			& \textbf{.55(.01)} 	& .44(.01)				\\
			\Haberman		& .24(.10)			& .26(.07) 			& .36(.08)			& .38(.07) 			& .40(.07)			& .17(.21) 			& .07(.06)			& \textbf{.41(.04)}		& \textbf{.41(.04)}		\\
			\Heart			& .78(.06)			& \textbf{.79(.07)}	& .71(.01)			& .73(.09) 			& .39(.31)			& .74(.05) 			& .70(.09)			& .77(.06)				& .75(.02)				\\
			\Ilpd           & .47(.02)			& .44(.08) 			& .42(.10)			& .20(.11) 			& .48(.08)			& .14(.13) 			& .09(.08)			& \textbf{.54(.06)} 	& .52(.04)				\\
			\Liver			& .58(.08)			& .58(.07) 			& .56(.10)			& .59(.04) 			& .58(.07)			& .54(.03) 			& .61(.05)			& .60(.07) 				& \textbf{.63(.06)}		\\
			\Pima			& .61(.04)			& .63(.04) 			& .60(.01)			& .60(.03) 			& .38(.18)			& .61(.07) 			& .03(.03)			& \textbf{.64(.05)} 	& .\textbf{.64(.05)}	\\
			\Sonar			& .81(.04)			& \textbf{.83(.05)}	& .75(.05)			& .77(.08) 			& .70(.06)			& .76(.05) 			& .69(.06)			& .79(.03) 				& .76(.04)				\\
			\Tictactoe 		& .99(.01)			& .99(.01) 			& .97(.01)			& .98(.01) 			& .55(.10)			& .99(.01) 			& .99(.01)			& \textbf{1.0(.00)}		& .68(.04)						\\
			\Transfusion    & .41(.07)			& .35(.06) 			& .35(.05)			& .42(.10) 			& .42(.08)			& .05(.10) 			& .04(.05)			& \textbf{.49(.12)} 	& \textbf{.49(.12)}		\\
			\Wisconsin		& \textbf{.95(.02)}	& \textbf{.95(.01)}	& .91(.04)			& .94(.02) 			& \textbf{.95(.01)}	& .94(.02) 			& .88(.03)			& \textbf{.95(.01)}		& .93(.02)				\\
			\Sapclean		& .93(.02)			& .93(.01)			& .85(.03)			& .86(.02)			& .88(.01)			& .90(.01)			& .68(.03)			& \textbf{.95(.02)}		& .72(.03)				\\
			\Sapfull		& -					& -					& -					& -					& -					& .81(.02)			& -					& \textbf{.89(.03)}		& .68(.04)				\\
			\midrule		
			Avg Rank        & 4.0(1.8)			& 3.1(1.9)			& 5.5(1.9)			& 5.3(1.7)			& 5.0(2.8)			& 5.3(2.3)			& 7.3(2.5) 			& \textbf{1.5(0.9)}		& 4.0(2.6)\\
			\bottomrule
		\end{tabular}
    \end{center}
	\caption{F1-score (st. dev. in parenthesis).}
    \label{tab:f1-score}
\end{table*}

\begin{table*}[p]
    \scriptsize
	\begin{center}
        \begin{tabular}{l|c|c|c|c|c|c|c}
			\toprule
            Dataset 		& \DT				& \RIPPERK 			& \MODLEM 				& \SBRL 			& \BRS 				& \underline{\LIBRE} 		& \underline{\LIBRE 3} 		\\
            \midrule																						
            \Adult			& 287.8(6.5)		& 21.4(5.2)			& 4957.8(36.3) 			& 71.4(2.1)			& 10.0(3.3)			& 14.0(2.1)					& \textbf{3.0(0.0)}			\\
            \Australian		& 4.0(0.0)			& 3.8(1.2)			& 86.6(3.2) 			& 5.8(0.7)			& \textbf{1.8(0.4)}	& 2.4(1.4)					& 2.2(0.7)					\\
			\Balance		& 48.0(12.5)		& 0.0(0.0)			& 76.5(4.6)				& 0.0(0.0)			& \textbf{1.0(0.0)}	& 9.0(3.0)					& \textbf{1.0(0.0)}			\\
			\Bank			& 545.4(18.3)		& 9.0(1.8)			& 3722.6(25.5)			& 61.2(5.5)			& 4.8(1.2)			& 15.0(1.1)					& \textbf{2.0(0.6)}			\\
            \Haberman		& 37.4(4.13)		& \textbf{1.0(0.0)}	& 73.6(2.9)				& 5.4(1.9)			& \textbf{1.0(0.0)}	& 1.6(0.8)					& 1.6(0.8)					\\
			\Heart			& 45.6(9.1)			& 2.8(0.7)			& 50.6(2.9)				& 5.8(0.7)			& \textbf{2.4(0.5)}	& 10.6(3.0)					& 2.8(0.4)					\\
			\Ilpd			& 80.6(30.2)		& \textbf{1.0(0.6)}	& 128.2(7.8)			& 4.8(0.7)			& \textbf{1.0(0.0)}	& 4.4(2.3)					& 2.2(0.4)					\\
            \Liver			& 84.4(15.2)		& 1.4(0.8)			& 98.4(1.6)				& 4.0(0.6)			& \textbf{1.0(0.0)}	& 3.4(1.9)					& 2.8(0.4)					\\
			\Pima			& 84.8(43.1)		& 2.4(2.4)			& 151.8(7.6)			& 8.4(0.5)			& \textbf{1.0(0.0)}	& 1.6(1.0)					& 1.6(1.0)					\\
			\Sonar			& 15.0(9.3)			& 3.6(1.4)			& 48.8(1.6)				& 3.2(0.7)			& \textbf{1.0(0.0)}	& 6.6(1.2)					& 1.1(0.2)					\\
			\Tictactoe		& 60.4(5.2)			& 10.6(1.6)			& 25.8(1.6)				& 12.2(1.2)			& 9.0(1.1)			& 9.0(1.1)					& \textbf{3.0(0.0)}			\\
			\Transfusion	& 100.2(48.4)		& 1.8(0.4) 			& 125.8(6.1) 			& 4.4(0.8)			& \textbf{1.0(0.0)}	& 1.2(0.4)					& 1.2(0.4)					\\
            \Wisconsin		& 31.4(5.5)			& 5.0(0.6)			& 29.2(1.9)				& 7.0(1.1)			& 5.0(0.6)			& 4.2(0.7)					& \textbf{3.0(0.0)}			\\
            \Sapclean		& 622.4(51.9)		& 19.3(3.6)			& 3944.5(18.8)			& 47.7(4.4)			& 20.2(3.5)			& 13.0(2.4)					& \textbf{3.0(0.0)}			\\
			\Sapfull		& -					& -					& -						& 56.4(4.6)			& -					& 17.5(5.2)					& \textbf{3.0(0.0)}			\\
			\midrule					
			Avg Rank		& 5.9(0.9)			& 3.3(1.3)			& 6.5(0.9) 				& 4.8(0.8)			& \textbf{1.7(1.1)}	& 3.1(1.1)					& \textbf{1.7(0.8)} \\
			\bottomrule
        \end{tabular}
    \end{center}
	\caption{\#rules (st. dev. in parenthesis).}
    \label{tab:rules}
\end{table*}

\begin{table*}[pt]
    \scriptsize
	\begin{center}
        \begin{tabular}{l|c|c|c|c|c|c|c}
			\toprule
            Dataset 		& \DT				& \RIPPERK 			& \MODLEM 				& \SBRL 			& \BRS 				& \underline{\LIBRE} 		& \underline{\LIBRE 3}\\
            \midrule																						
            \Adult			& 9.3(0.0) 			& 4.4(0.3)			& 4.3(0.1) 				& 87.0(3.2) 		& \textbf{3.3(0.1)}	& 7.8(1.0)					& 6.5(0.7)\\
            \Australian		& \textbf{2.0(0.0)}	& 2.4(0.3)			& 2.3(0.1) 				& 7.1(1.0) 			& 3.5(0.3)			& 4.4(1.8)					& 4.4(1.3)\\
			\Balance		& 4.4(2.9) 			& 0.0(0.0)			& 3.5(0.0) 				& 0.0(0.0)			& 4.0(0.0)			& \textbf{2.1(0.0)}			& \textbf{2.1(0.0)}\\
			\Bank			& 9.5(0.0) 			& 3.0(0.2)			& 3.0(0.0) 				& 89.0(7.8)			& 3.2(0.4)			& 4.7(0.5)					& \textbf{2.0(0.1)}\\
            \Haberman		& 4.6(3.3) 			& \textbf{1.8(0.4)}	& 2.2(0.1) 				& 3.7(1.0)			& 3.2(0.7)			& 2.1(0.3)					& 2.1(0.3)\\
			\Heart			& 6.2(0.3) 			& \textbf{2.1(0.3)}	& 2.3(0.1) 				& 8.1(1.3)			& 3.3(0.2)			& 7.7(0.7)					& 6.1(2.7)\\
			\Ilpd			& 8.5(1.6) 			& \textbf{2.1(1.3)}	& \textbf{2.1(0.0)}		& 4.4(0.4)			& 2.8(0.4)			& 3.3(1.5)					& 3.0(0.6)\\
            \Liver			& 8.7(1.0) 			& \textbf{1.3(0.4)}	& 2.1(0.1) 				& 3.0(0.3)			& 3.4(0.5)			& 2.5(1.0)					& \textbf{1.3(0.4)}\\
			\Pima			& 6.9(2.5) 			& 2.4(0.5)			& \textbf{2.1(0.1)}		& 6.3(0.8)			& 3.6(0.5)			& 2.5(0.7)					& 2.5(0.7)\\
			\Sonar			& 3.8(1.5) 			& 2.1(0.2)			& \textbf{1.4(0.0)}		& 8.2(2.1)			& 4.0(0.0)			& 3.7(1.0)					& 2.2(0.8)\\
			\Tictactoe		& 6.7(0.1) 			& \textbf{2.1(0.2)}	& 3.5(0.0) 				& 21.8(1.6)			& 3.5(0.1)			& 3.8(0.8)					& 3.0(0.0)\\
			\Transfusion	& 6.9(2.5) 			& 2.8(0.4)			& \textbf{2.3(0.0)}		& 3.8(0.6)			& 3.0(0.6)			& 2.8(0.7)					& 2.8(0.7)\\
            \Wisconsin		& 6.1(0.4) 			& \textbf{2.0(0.2)}	& 2.2(0.1)				& 5.9(1.0)			& 3.3(0.3)			& 3.2(1.0)					& 2.8(1.0)\\
            \Sapclean		& 15.2(1.0)			& 3.8(0.3)			& 3.4(0.1)				& 75.4(4.0)			& 3.9(0.4)			& 3.3(0.2)					& \textbf{3.0(0.1)}\\
			\Sapfull		& -					& -					& -						& 85.6(9.7)			& -					& 4.7(0.3)					& \textbf{4.2(0.2)}\\
			\midrule					
			Rank			& 5.8(1.6)			& \textbf{2.3(1.4)}	& \textbf{2.3(1.0)}		& 6.2(1.0) 			& 4.2(1.3)			& 3.7(1.5)					& 2.5(1.7)\\
			\bottomrule
        \end{tabular}
    \end{center}
	\caption{\#atom (st. dev. in parenthesis).}
    \label{tab:atoms}
\end{table*}

\begin{figure*}[p]
    \centering
    \begin{subfigure}[b]{0.4\textwidth}
        \includegraphics[width=\textwidth]{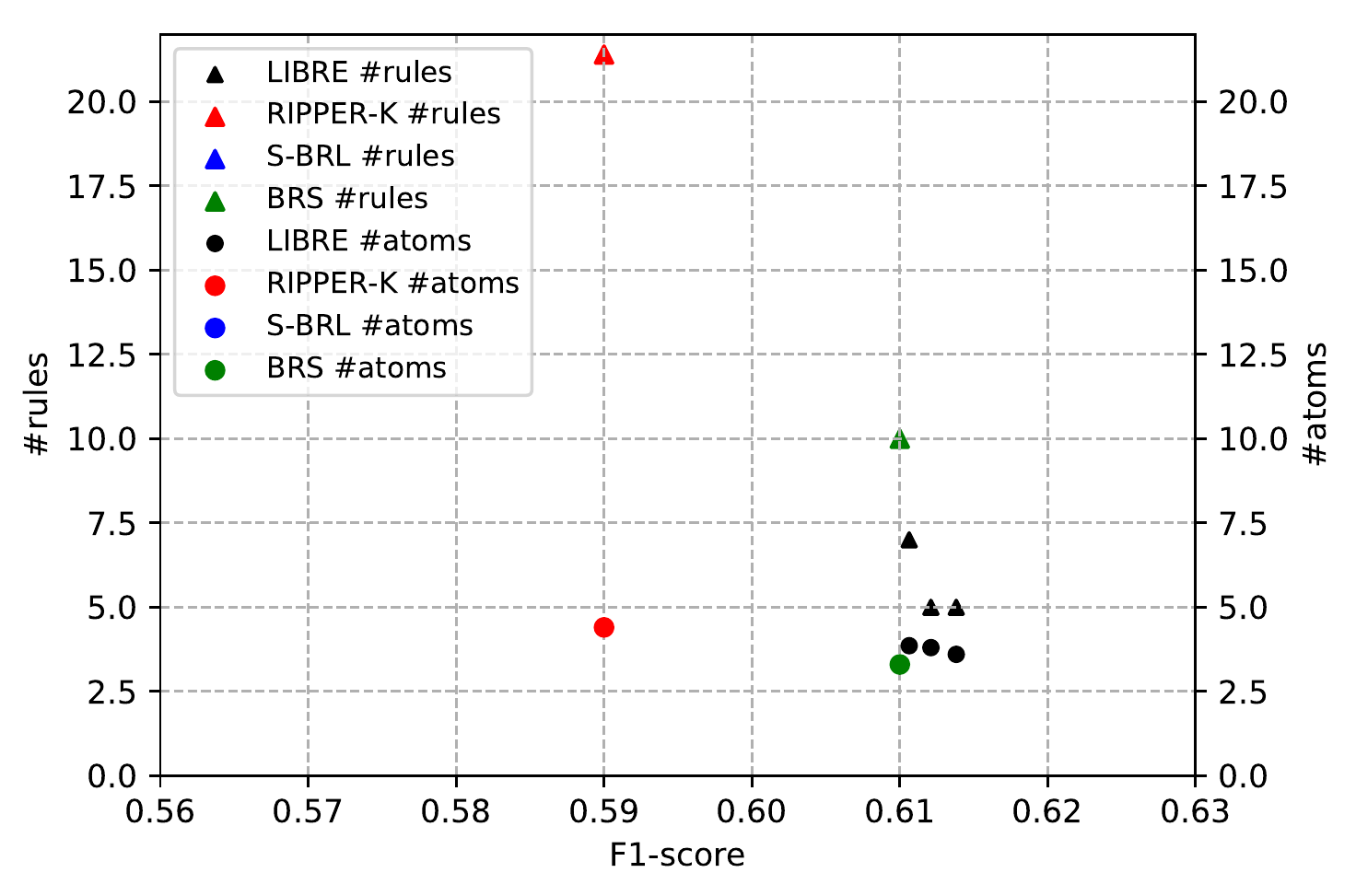}
        \caption{\Adult}
        \label{fig:interpretable_adult}
    \end{subfigure}
    \begin{subfigure}[b]{0.4\textwidth}
        \includegraphics[width=\textwidth]{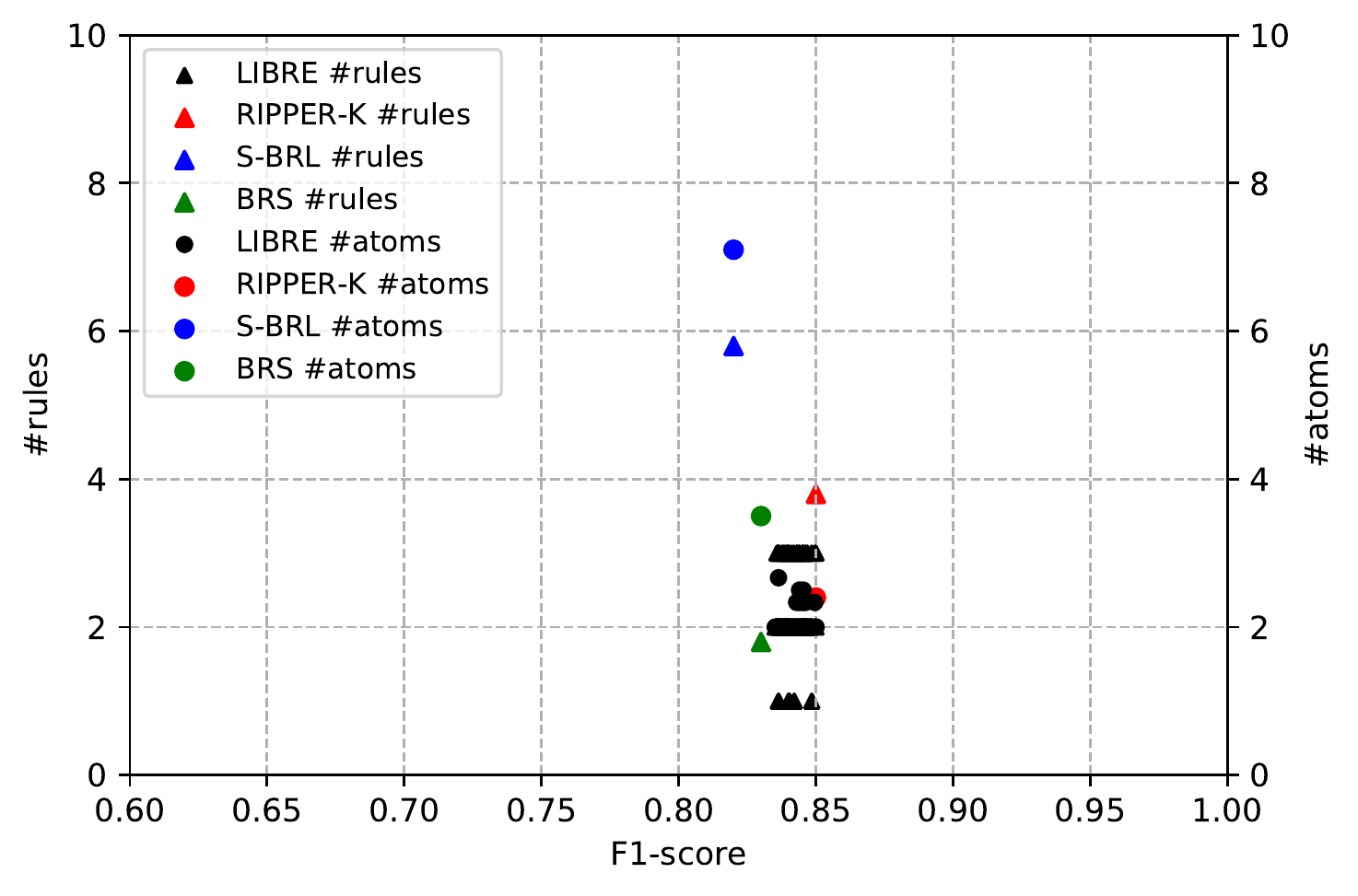}
        \caption{\Australian}
        \label{fig:interpretable_australian}
    \end{subfigure}
	\begin{subfigure}[b]{0.4\textwidth}
        \includegraphics[width=\textwidth]{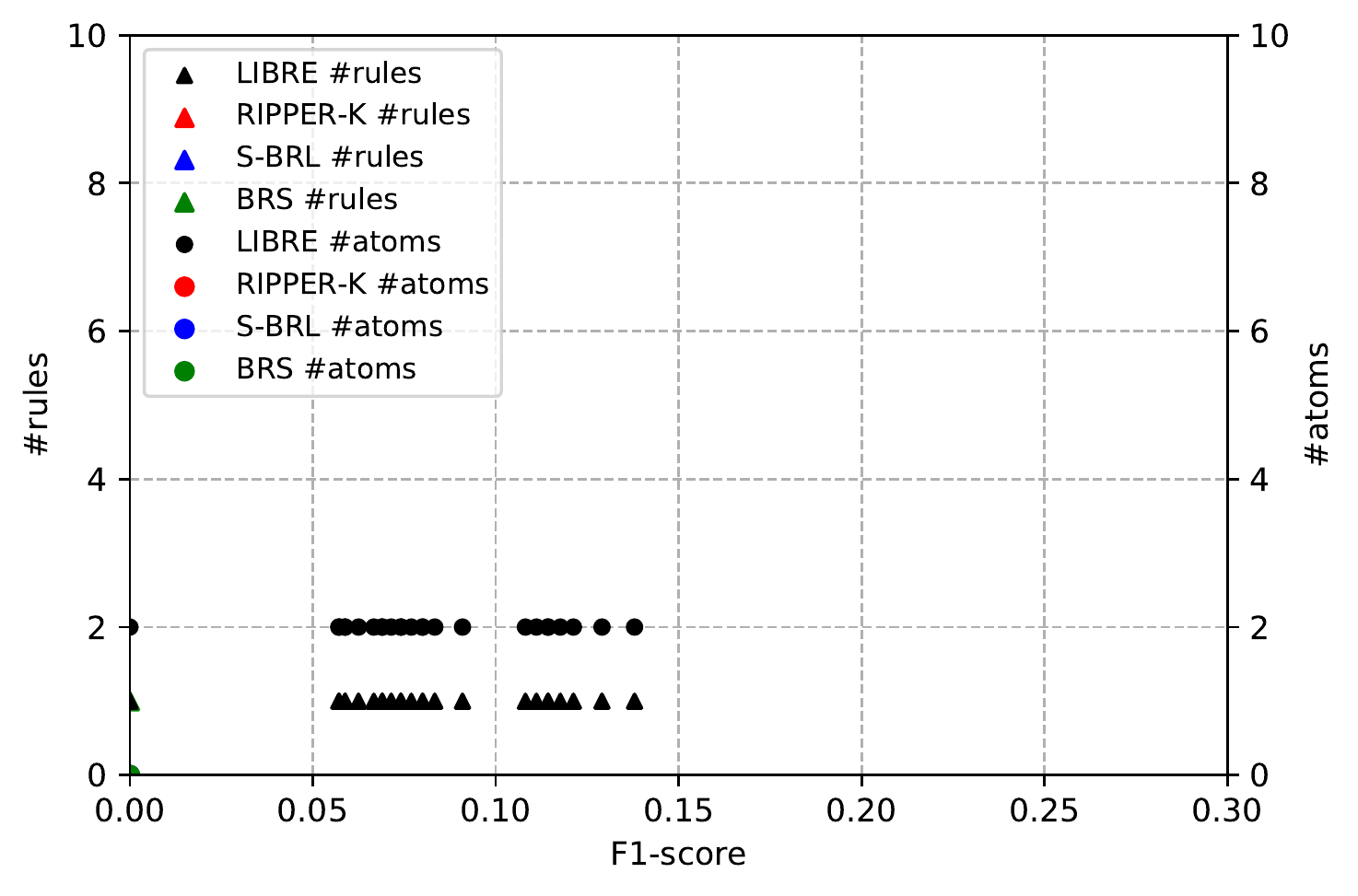}
        \caption{\Balance}
        \label{fig:interpretable_balance}
    \end{subfigure}
    \begin{subfigure}[b]{0.4\textwidth}
        \includegraphics[width=\textwidth]{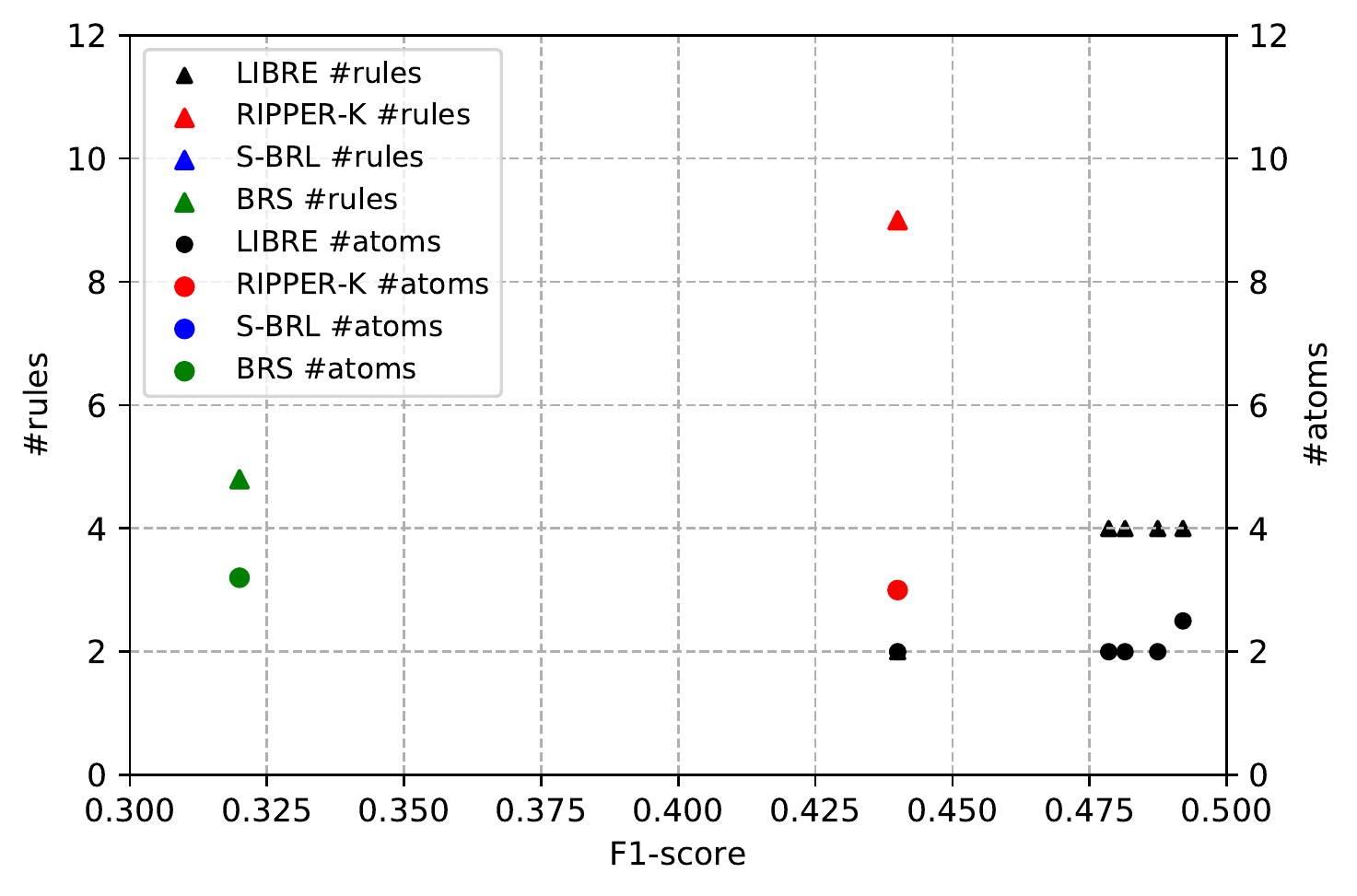}
        \caption{\Bank}
        \label{fig:interpretable_bank}
    \end{subfigure}
	\begin{subfigure}[b]{0.4\textwidth}\ContinuedFloat
        \includegraphics[width=\textwidth]{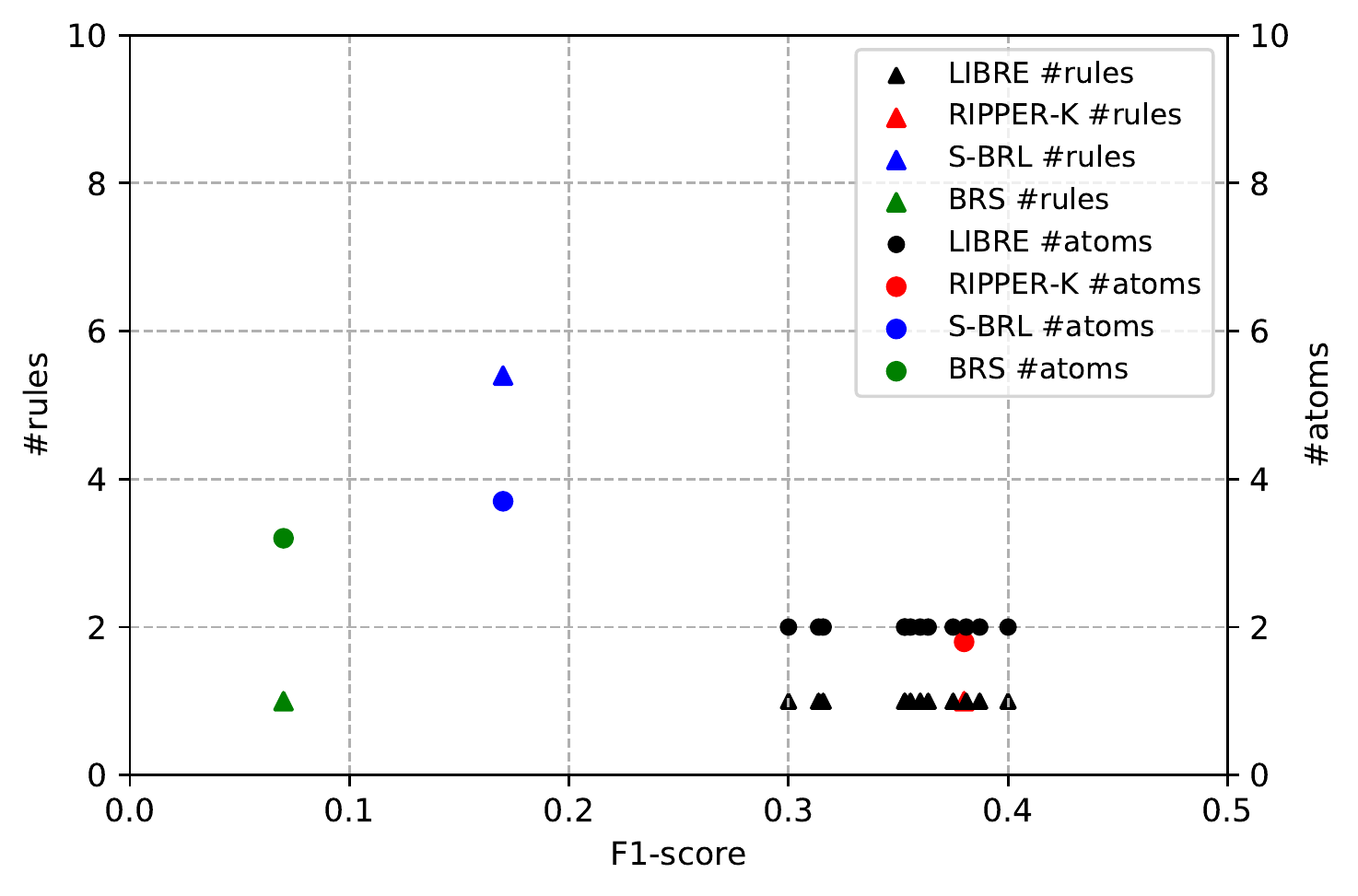}
        \caption{\Haberman}
        \label{fig:interpretable_haberman}
    \end{subfigure}
    \begin{subfigure}[b]{0.4\textwidth}
        \includegraphics[width=\textwidth]{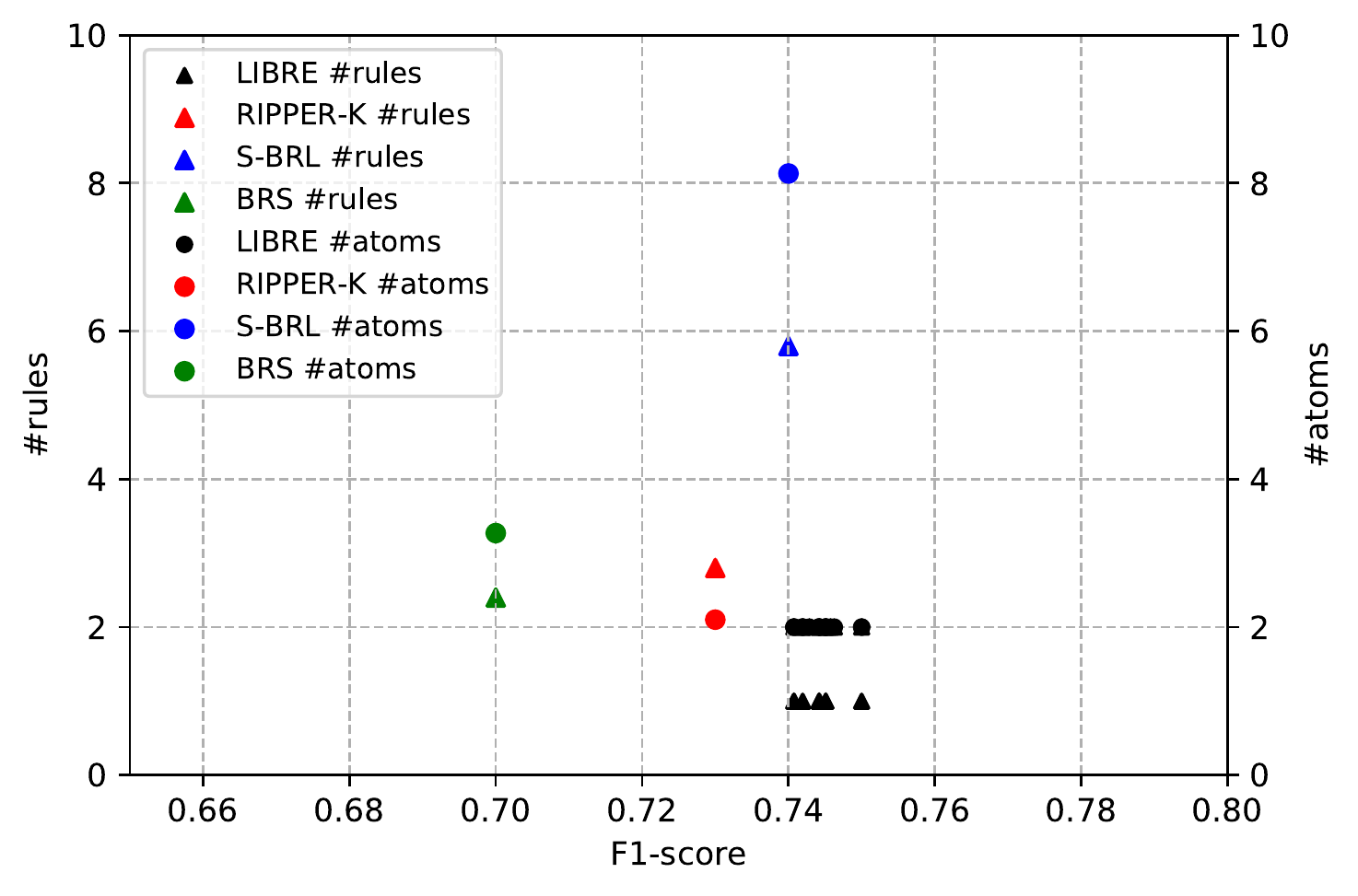}
        \caption{\Heart}
        \label{fig:interpretable_heart}
    \end{subfigure}
	\caption{F1-score vs. \#rules and \#atoms}
	\label{fig:interpretable_rulesets}
\end{figure*}
\begin{figure*}
	\ContinuedFloat
	\centering
    \begin{subfigure}[b]{0.4\textwidth}
        \includegraphics[width=\textwidth]{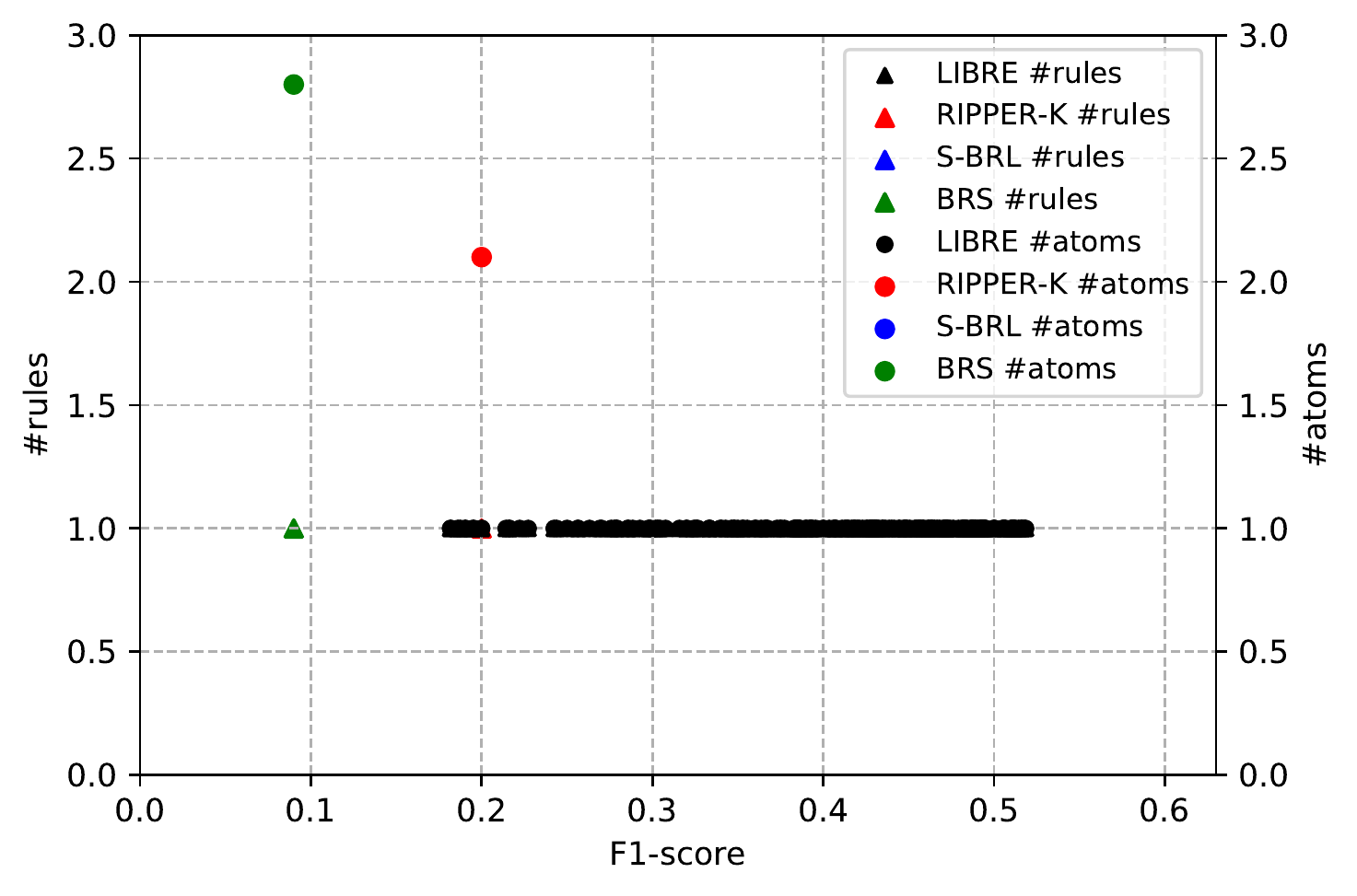}
        \caption{\Ilpd}
        \label{fig:interpretable_ilpd}
    \end{subfigure}
    \begin{subfigure}[b]{0.4\textwidth}
        \includegraphics[width=\textwidth]{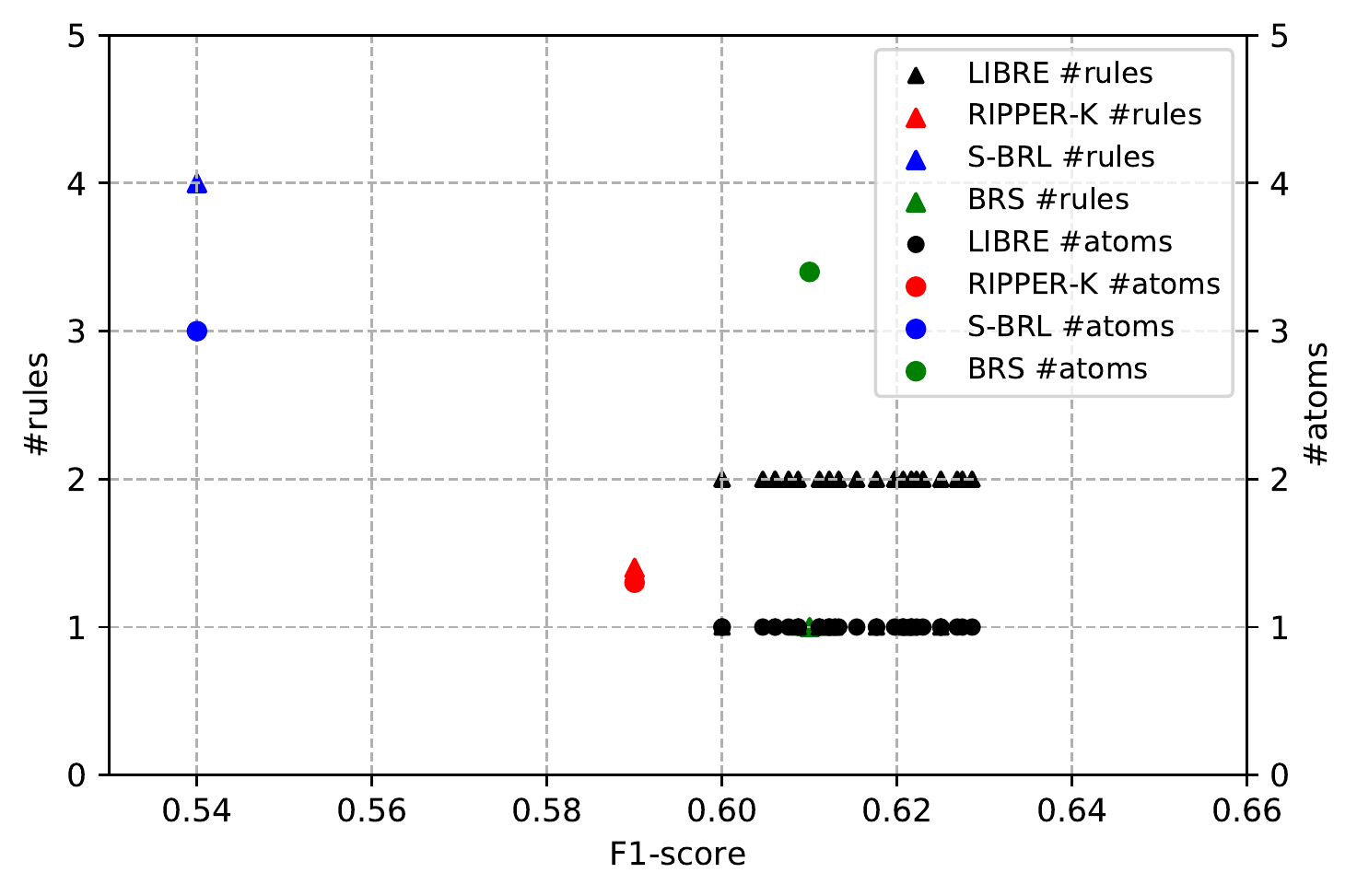}
        \caption{\Liver}
        \label{fig:interpretable_liver}
    \end{subfigure}
    \begin{subfigure}[b]{0.4\textwidth}
        \includegraphics[width=\textwidth]{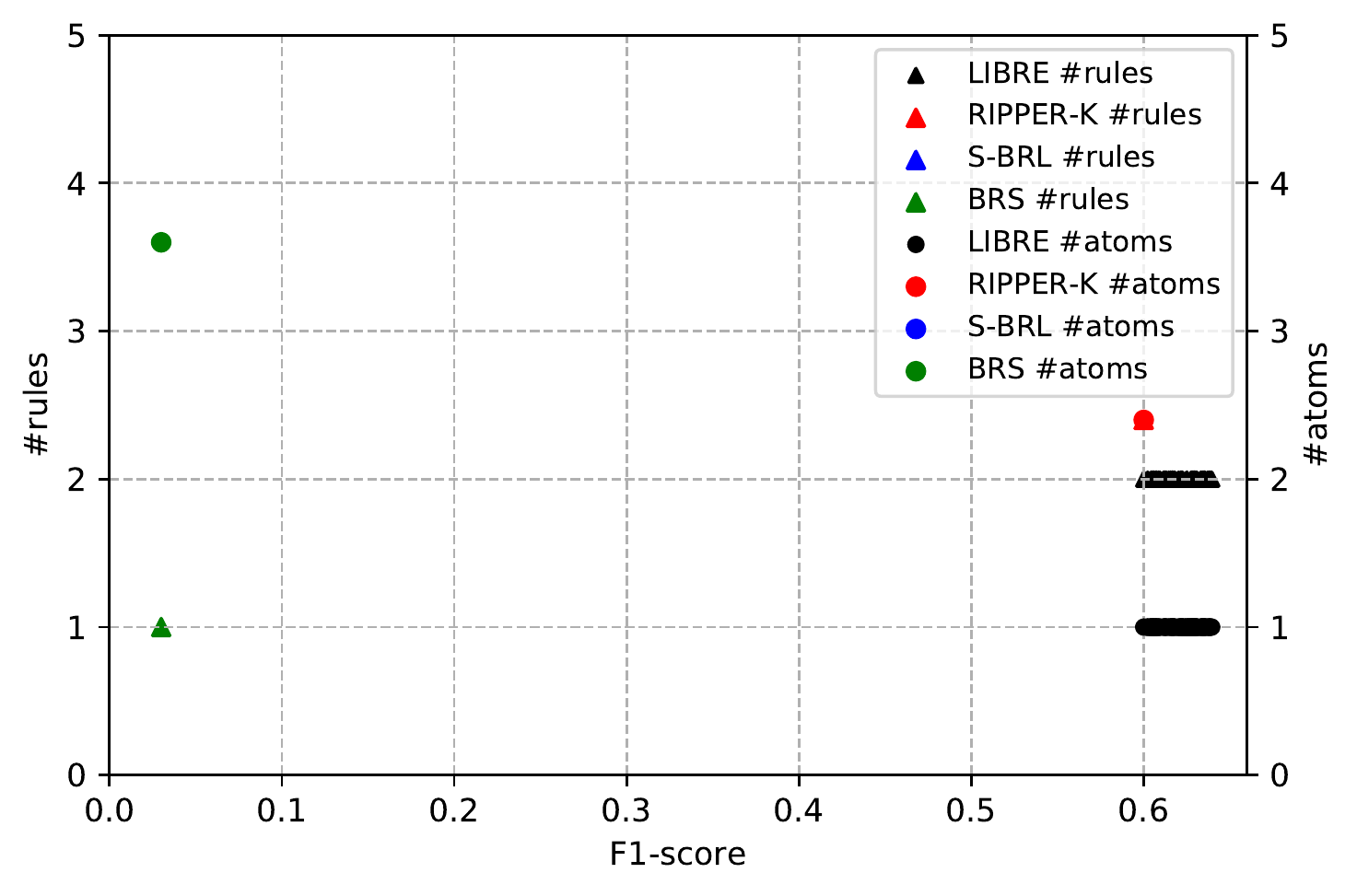}
        \caption{\Pima}
        \label{fig:interpretable_pima}
    \end{subfigure}
    \begin{subfigure}[b]{0.4\textwidth}
        \includegraphics[width=\textwidth]{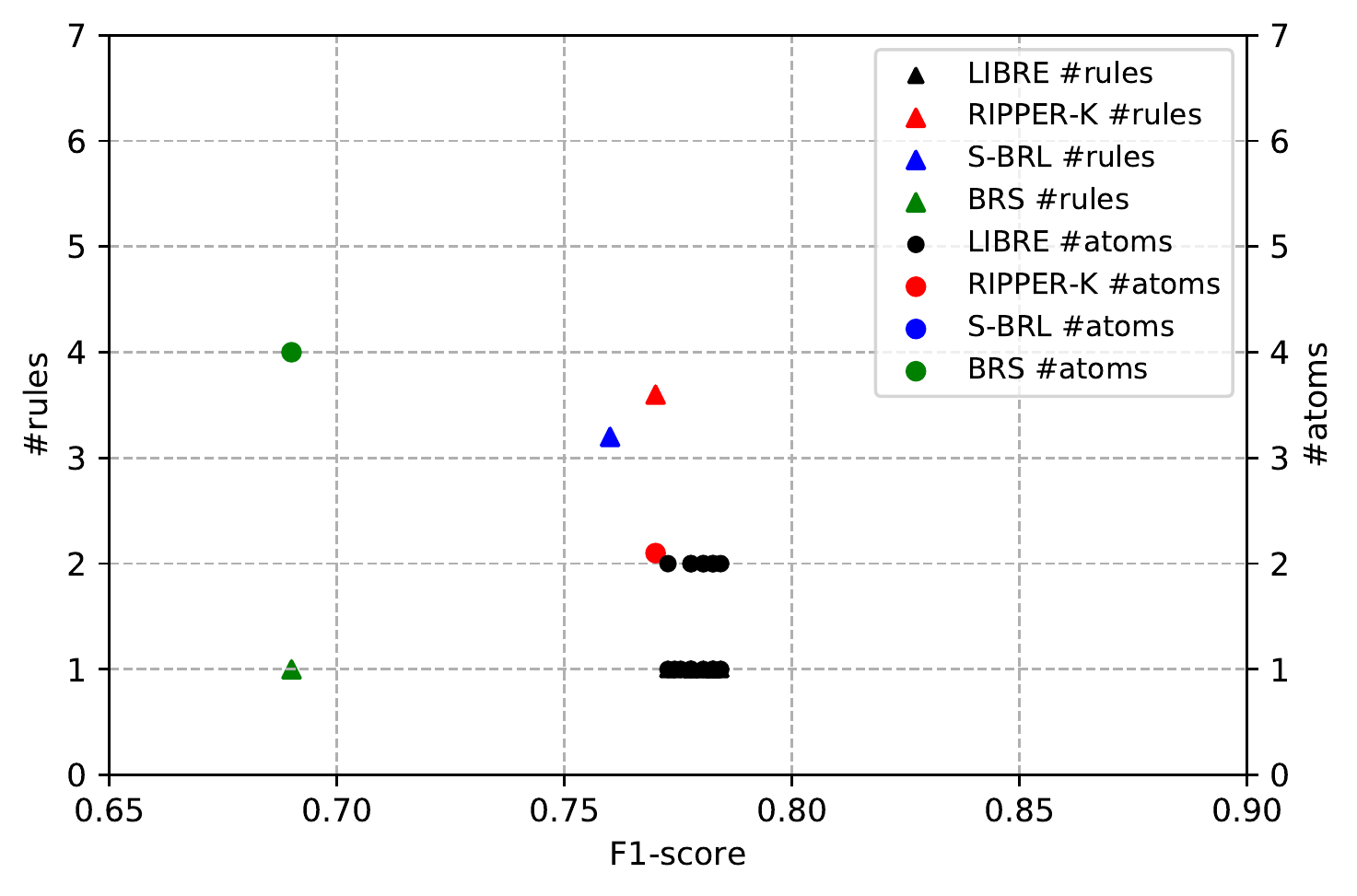}
        \caption{\Sonar}
        \label{fig:interpretable_sonar}
    \end{subfigure}
    \begin{subfigure}[b]{0.4\textwidth}
        \includegraphics[width=\textwidth]{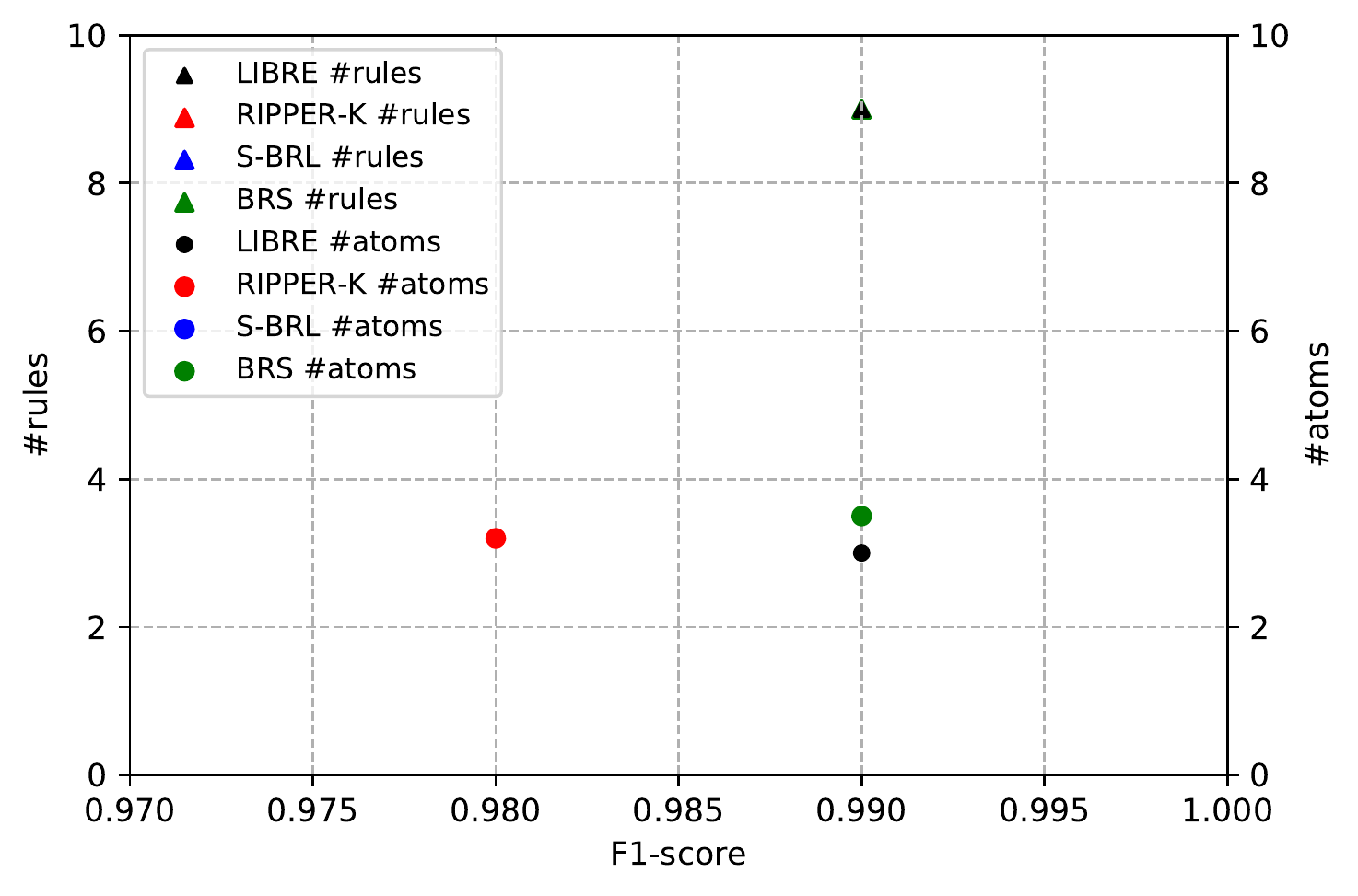}
        \caption{\Tictactoe}
        \label{fig:interpretable_tictactoe}
    \end{subfigure}
    \begin{subfigure}[b]{0.4\textwidth}
        \includegraphics[width=\textwidth]{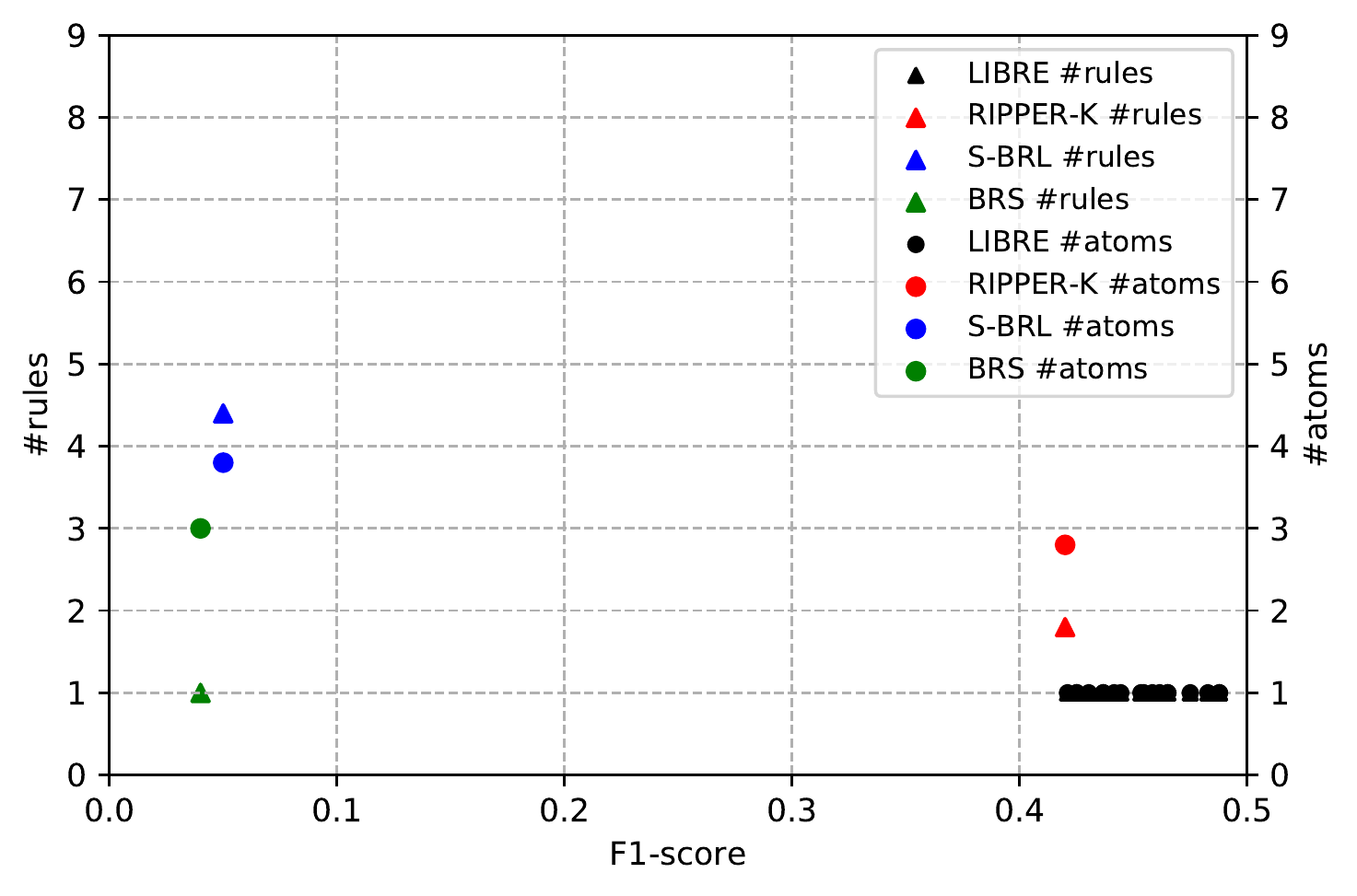}
        \caption{\Transfusion}
        \label{fig:interpretable_transfusion}
    \end{subfigure}
	\begin{subfigure}[b]{0.4\textwidth}
        \includegraphics[width=\textwidth]{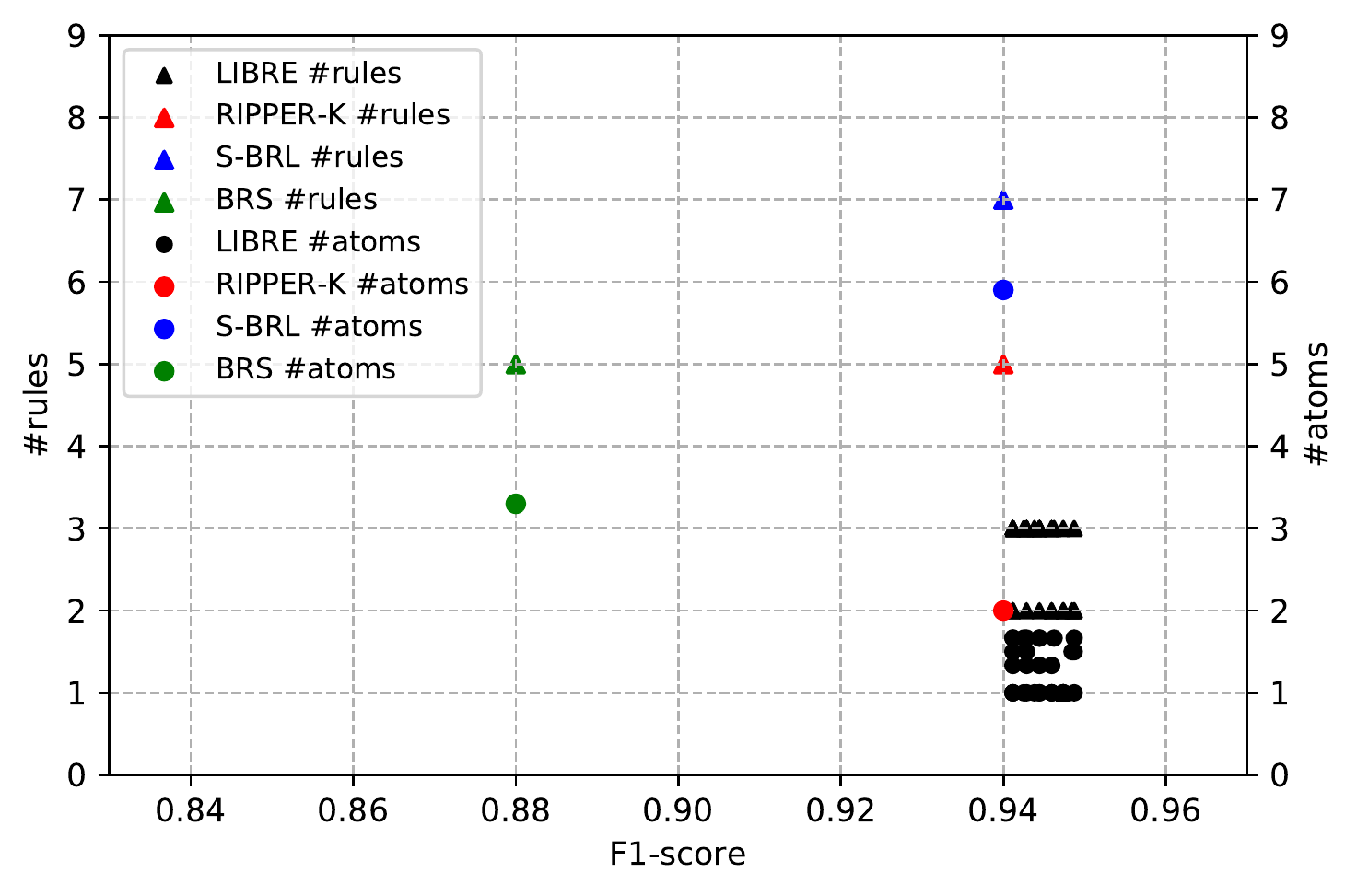}
        \caption{\Wisconsin}
        \label{fig:interpretable_wisconsin}
    \end{subfigure}
	\begin{subfigure}[b]{0.4\textwidth}
        \includegraphics[width=\textwidth]{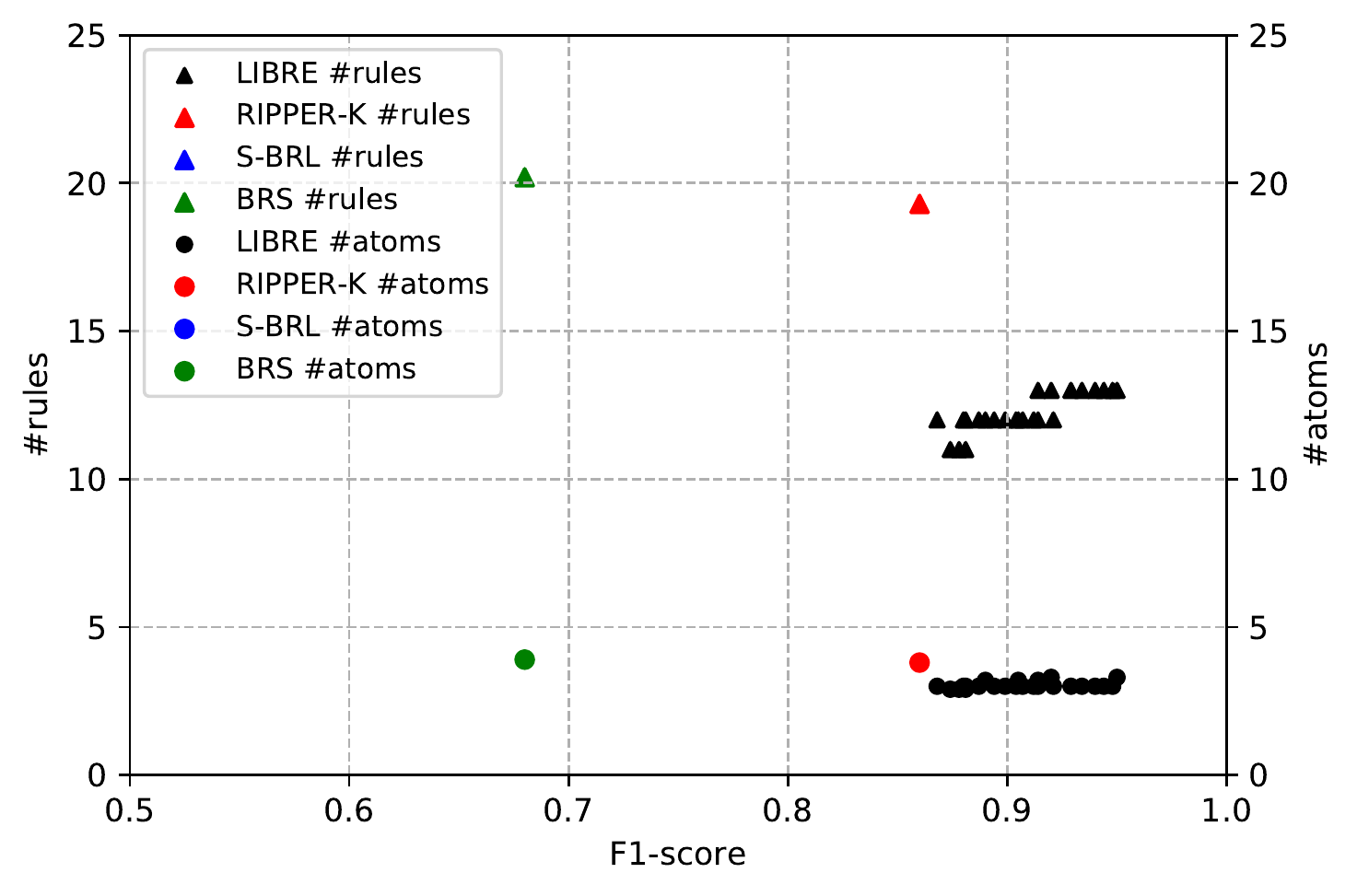}
        \caption{\Sapclean}
        \label{fig:interpretable_sapclean}
    \end{subfigure}
	\caption{F1-score vs. \#rules and \#atoms}
	\label{fig:interpretable_rulesets}
\end{figure*}

\section{MORE EXAMPLES OF RULE SETS LEARNED BY LIBRE}
\label{sec:rule_sets_LIBRE}
In the main paper, we showed an example of rule set learned by LIBRE for \Liver.
In this section, we report additional examples for the medical UCI datasets described in \cref{tab:datasets}, for 
which it might be interesting to understand the relation between input features and the predicted 
diseases. 

Please, notice that different rule sets may be obtained depending on how folds are randomly built during 
cross validation.



\begin{figure}[p]
	\footnotesize
	\begin{mdframed}
		\textbf{IF} (\\
			number\_of\_positive\_axillary\_nodes $\in [2, max]$\\
		)\\
		\textbf{THEN}  died within 5 years\\
		\textbf{ELSE}  survived 5 years or longer
	\end{mdframed}
	\caption{Example of rules learned by \LIBRE for \Haberman.}
	\label{fig:haberman_ruleset}
\end{figure}

\begin{figure}[p]
	\footnotesize
	\begin{mdframed}
		\textbf{IF} (\\
			slope\_of\_the\_peak\_exercise $\in \{flat, downsloping\}$ \textbf{AND}\\
			number\_of\_major\_vessels $\in [1, 3]$\\
		)\\
		\textbf{OR} (\\
			chest\_pain\_type $\in \{asymptomatic\}$ \textbf{AND}\\
			thal $\in \{reversable\_defect\}$\\
		)\\
		\textbf{OR} (\\
			sex $\in \{male\},$ \textbf{AND}\\
			fasting\_blood\_sugar\_$>$120mg/dl $\in \{False\}$ \textbf{AND}\\
			number\_of\_major\_vessels $\in [1, 3]$\\
		)\\
		\textbf{THEN}  class = presence\\
		\textbf{ELSE}  class = absence
	\end{mdframed}
	\caption{Example of rules learned by \LIBRE for \Heart.}
	\label{fig:heart_ruleset}
\end{figure}

\begin{figure}[p]
	\footnotesize
	\begin{mdframed}
		\textbf{IF} (\\
			TB $\in [min, 2)$ \textbf{AND}\\
			sgbp $\in [min, 42)$
		)\\
		\textbf{OR} (\\
			TB $\in [min, 2)$ \textbf{AND}\\
			alkphos $\in [min, 184)$\\
		)\\
		\textbf{OR} (\\
			age $\in [35, 39),[56, 57)$ \textbf{AND}\\
			sgbp $\in [42, 148)$\\
		)\\
		\textbf{THEN}  class = liver patient\\
		\textbf{ELSE}  class = non liver patient
	\end{mdframed}
	\caption{Example of rules learned by \LIBRE for \Ilpd.}
	\label{fig:ilpd_ruleset}
\end{figure}

\begin{figure}[p]
	\footnotesize
	\begin{mdframed}
		\textbf{IF}  (\\
			mean\_corpuscular\_volume $\in [90, 96)$\\
		)\\
		\textbf{OR}	(\\
			gamma\_glutamyl\_transpeptidase $\in [20, max]$\\
		)\\
		\textbf{THEN}  liver\_disorder = True\\
		\textbf{ELSE}  liver\_disorder = False
	\end{mdframed}
	\caption{Example of rules learned by \LIBRE for \Liver.}
	\label{fig:liver_ruleset}
\end{figure}

\begin{figure}[p]
	\footnotesize
	\begin{mdframed}
		\textbf{IF}  (\\
			glucose $\in$ [158, max] \textbf{AND}\\
			blood\_pressure $\in [56, max]$\\
		)\\
		\textbf{OR}  (\\
			glucose $\in$ [110, 158] \textbf{AND}\\
			BMI $\in [30.7, max)$\\
		)\\
		\textbf{OR}  (\\
			pregnancies $\in$ [4, max] \textbf{AND}\\
			diabetes\_predigree\_func $\in [0.529, max]$\\
		)\\
		\textbf{THEN}  diabetes = True\\
		\textbf{ELSE}  diabetes = False
	\end{mdframed}
	\caption{Example of rules learned by \LIBRE for \Pima.}
	\label{fig:pima_ruleset}
\end{figure}

\begin{figure}[p]
	\footnotesize
	\begin{mdframed}
		\textbf{IF}  (\\
			months\_since\_last\_donation $\in$ [0, 8) \textbf{AND}\\
			total\_blood\_donated $\in [1250, max)$\\
		)\\
		\textbf{THEN}  transfusion = Yes\\
		\textbf{ELSE}  transfusion = No
	\end{mdframed}
	\caption{Example of rules learned by \LIBRE for \Transfusion.}
	\label{fig:transfusion_ruleset}
\end{figure}

\begin{figure}[p]
	\footnotesize
	\begin{mdframed}
		\textbf{IF}  (\\
			uniformity\_of\_cell\_shape $\in [5, max]$\\
		)\\
		\textbf{OR}  (\\
			clump\_thickness $\in [2, max]$ \textbf{AND}\\
			bare\_nuclei $\in [8, max)$\\
		)\\
		\textbf{OR}  (\\
			clump\_thickness $\in [7, max]$ \textbf{AND}\\
			marginal\_adhesion $\in [1, 2), [4, max)$\\
		)\\
		\textbf{THEN}  transfusion = Yes\\
		\textbf{ELSE}  transfusion = No
	\end{mdframed}
	\caption{Example of rules learned by \LIBRE for \Wisconsin.}
	\label{fig:wisconsin_ruleset}
\end{figure}

\end{document}